\documentclass{article}

\usepackage{microtype}
\usepackage{graphicx}
\usepackage{subcaption}
\usepackage{booktabs} 

\usepackage[accepted, nohyperref]{icml2023}

\usepackage{amsmath}
\usepackage{amssymb}
\usepackage{mathtools}
\usepackage{amsthm}
\usepackage{placeins}

\usepackage[capitalize,noabbrev]{cleveref}

\theoremstyle{plain}

\theoremstyle{definition}

\theoremstyle{remark}

\usepackage[textsize=tiny]{todonotes}

\icmltitlerunning{Benchmarking BNNs and evaluation metrics regression tasks}

\usepackage{url}            
\usepackage{amsfonts}       
\usepackage{nicefrac}       
\usepackage{xcolor}         
\usepackage{dsfont}

\usepackage{enumerate}
\usepackage{empheq}
\usepackage{tabularx}
\usepackage{comment}
\usepackage{pifont}
\newcommand{\cmark}{\ding{51}}%
\newcommand{\xmark}{\ding{55}}%

\newcommand{\KSD}{\mathrm{KSD}}

\newcommand{\bP}{\mathbb{P}}
\newcommand{\bQ}{\mathbb{Q}}
\newcommand{\bx}{\mathbf{x}}
\newcommand{\by}{\mathbf{y}}

\newcommand{\bX}{\mathbf{X}}
\newcommand{\bY}{\mathbf{Y}}

\newcommand{\btheta}{\mathbf{w}}

\begin{document}

\twocolumn[
\icmltitle{Benchmarking Bayesian neural networks and evaluation metrics for regression tasks}



\icmlsetsymbol{equal}{*}

\begin{icmlauthorlist}
\icmlauthor{Brian Staber}{equal,aut}
\icmlauthor{Sébastien Da Veiga}{sch}
\end{icmlauthorlist}

\icmlaffiliation{aut}{Safran Tech, Digital Sciences \& Technologies, 78114 Magny-Les-Hameaux, France}
\icmlaffiliation{sch}{ENSAI, CREST, F-35000 Rennes, France}

\icmlcorrespondingauthor{Brian Staber}{brian.staber@safrangroup.com}
\icmlcorrespondingauthor{Sébastien Da Veiga}{sebastien.da-veiga@ensai.fr}
\icmlkeywords{Machine Learning, ICML}

\vskip 0.3in
]



\printAffiliationsAndNotice{\icmlEqualContribution} 

\graphicspath{{figures/}}


\begin{abstract}
Due to the growing adoption of deep neural networks in many fields of science and engineering, modeling and estimating their uncertainties has become of primary importance. Despite the growing literature about uncertainty quantification in deep learning, the quality of the uncertainty estimates remains an open question. In this work, we assess for the first time the performance of several approximation methods for Bayesian neural networks on regression tasks by evaluating the quality of the confidence regions with several coverage metrics. The selected algorithms are also compared in terms of predictivity, kernelized Stein discrepancy and maximum mean discrepancy with respect to a reference posterior in both weight and function space. Our findings show that (i) some algorithms have excellent predictive performance but tend to largely over or underestimate uncertainties (ii) it is possible to achieve good accuracy and a given target coverage with finely tuned hyperparameters and (iii) the promising kernel Stein discrepancy cannot be exclusively relied on to assess the posterior approximation. As a by-product of this benchmark, we also compute and visualize the similarity of all algorithms and corresponding hyperparameters: interestingly we identify a few clusters of algorithms with similar behavior in weight space, giving new insights on how they explore the posterior distribution.
\end{abstract}

\section{Introduction}
Due to their recent achievements in the last decade, deep neural networks have been widely adopted in many research fields and industries. However, they still suffer from several shortcomings that prevent their deployement in fields where decisions involve high stakes. These limitations are mainly due to their inability to provide uncertainty estimates which are crucial for many real-world applications. Uncertainty quantification in deep learning has attracted a lot of attention and several approaches have been investigated. Thorough overviews are provided by the recent review papers of \citet{Abdar2019} and \citet{gawlikowski2021survey}. Amongst the available approaches, Bayesian neural networks offer a simple and flexible formulation but raise several challenges. The high-dimensional posterior distribution of the network parameters is intractable, possibly multimodal, and the influence of the prior distribution remains an open question. The posterior distribution is typically approximated using sampling methods, variational inference, or Gaussian approximations. When dealing with complex posterior distributions, gradient-based sampling methods such as Markov Chain Monte Carlo (MCMC) methods are often adopted. In particular, Hamiltonian Monte Carlo (HMC) is usually considered as a gold standard sampling algorithm but is unfortunately extremely computationally demanding \cite{izmailov2021bayesian,cobb2021scaling}. Since the work of \citet{welling2011bayesian}, stochastic gradient MCMC methods have been extensively studied (see, \textit{e.g.}, \citep{ma2015complete} and \citep{nemeth2021}), where the Metropolis-Hastings correction is omitted and a mini-batched stochastic gradient is used. Variational approaches approximate the posterior distribution by minimizing the Kullback-Leibler divergence over a family of tractable distributions \citep{hoffman2013stochastic}. Several scalable methods have been proposed such as the Bayes by Backprop (BBB) method of \citet{graves2011practical} and \citet{blundell2015weight}, the multiplicative normalizing flows proposed by \citet{louizos2017multiplicative}, and the Monte Carlo dropout method \citep{gal2016dropout}. Although ensemble methods are generally used to improve the generalization capabilities of neural networks, they can also be interpreted as a Bayesian approach \cite{lakshminarayanan2017simple,fort2019deep}, where uncertainties are predicted thanks to random initialization and data shuffling. Gaussian approximations have also been investigated by several authors, such as the Laplace approximation \citep{ritter2018scalable,daxberger2021laplace} and the highly scalable stochastic weight averaging Gaussian method \citep{maddox2019simple}. 

Regardless of the approximation method, evaluating the quality of the predictive uncertainties remains an open issue. In contrast to previous works, we propose for the first time to estimate sensible coverage probabilities by taking into account the variability induced by the training dataset. We also compare the selected algorithms with the help of discrepancy measures, namely, the maximum mean discrepancy \citep{gretton2006kernel} and the kernelized Stein discrepancy \citep{liu2016kernelized}.

\section{Related works}
\citet{izmailov2021bayesian} have studied the performance of Bayesian neural networks (BNNs) by relying on exhaustive full-batch Hamiltonian Monte Carlo. The performance is assessed in terms of RMSE for regression problems and accuracy for classification tasks. Most notably, \citet{izmailov2021bayesian} have applied BNNs to practical deep architectures and large datasets thanks to impressive computational ressources. \citet{wenzel20a} have studied the influence of posterior temperature on the performance of stochastic gradient sampling methods. The performance is reported in terms of accuracy for classification tasks only. More closely related to this work, \citet{yao2019quality} has compared several approximation methods, for regression and classification tasks, and reported the prediction interval coverage probability in order to evaluate the quality of the approximated prediction intervals.

The objective of this work is to assess the performance of Bayesian neural networks on simple regression tasks by comparing several metrics. In contrast to previous works, we assess the validity of the confidence intervals with marginal and conditional coverage probabilities instead of predictive interval coverage probability, which are defined in section \ref{sec:evaluation_metrics}. It is worth emphasizing that the aim of this work is not to suggest to use marginal and conditional coverage for practical problems, but to investigate which performance BNNs can achieve. In particular, we investigate if there are correlations between accuracy, coverage metrics and kernel Stein discrepancy, as well as if some algorithms have the same exploration behavior in weight and function spaces. 

\section{Bayesian neural networks} \label{sec:bnns}
Let $\mathcal{D} = \{(\bX_i,\bY_i)\}_{i=1}^{N}$ denote a dataset made of $N$ observations of input $\bX_i \in \mathbb{R}^D$ and output $\bY_i \in \mathbb{R}^M$ pairs. The observations take the form $\bY_i = f(\bX_i) + \boldsymbol{\varepsilon}_i$ where $f : \mathbb{R}^D \rightarrow \mathbb{R}^M$ represents the unknown latent regression function, and $\varepsilon_i \sim \mathcal{N}(0,\sigma^2(\bX_i)\mathbf{I}_M)$ is an additive noise modeling aleatoric uncertainties. The latent function $f$ is approximated by a neural network $\hat{f}(\cdot;\btheta)$ with parameters $\btheta \in \mathbb{R}^d$. The observations $\bY_1, \dots, \bY_N$ are assumed to be i.d.d. and the likelihood function takes the form $p(\bY_i|\bX_i,\btheta) = \mathcal{N}(\bY_i; \hat{f}(\bX_i;\btheta), \sigma^2\mathbf{I}_M)$ for any $i \in \{1,\dots,N\}$. Given a prior distribution $p(\btheta)$ over the network parameters, the posterior distribution $p(\btheta|\mathbf{X},\mathbf{Y})$ can be deduced using Baye's formula: $p(\btheta|\mathbf{X},\mathbf{Y}) \propto p(\bY|\bX,\btheta)p(\btheta)$. The prediction of the model for a new test input $\bx$ is then given by
\begin{align*}
p(\by|\bx,\mathbf{X},\mathbf{Y}) = \int_{\mathbb{R}^d} p(\by|\bx, \btheta) p(\btheta|\mathbf{X},\mathbf{Y})d\btheta.
\end{align*}
Unfortunately, this integral is intractable and we must have recourse to approximate inference. The true posterior distribution is approximated by some $q(\btheta) \approx p(\btheta|\mathbf{X},\mathbf{Y})$ obtained via for instance MCMC methods, variational inference, or Gaussian approximations. In the rest of this section, we briefly describe the approximation methods that we consider in this paper, additional details are given in Appendix \ref{app:implementation}.

\paragraph{Monte Carlo Markov Chain.}
Monte Carlo Markov chain (MCMC) methods generate a Markov chain whose stationary distribution is the target posterior distribution. In this work, we consider the Hamiltonian Monte Carlo (\textbf{HMC}) algorithm \citep{neal2011mcmc} which requires the knowledge of the unnormalized posterior $p(\bY|\bX,\btheta)p(\btheta)$ and the gradient of the potential function $U(\btheta) = - \log(p(\bY|\bX,\btheta)p(\btheta))$, given by
\begin{align*}
\nabla U(\btheta) = - \sum_{i=1}^{N} \nabla U_i(\btheta) - \nabla\log(p(\btheta))\,,
\end{align*}
where $U_i(\btheta) = \log(p(\bY_i|\bX_i,\btheta))$.

\paragraph{Stochastic gradient Monte Carlo Markov Chain.}
In order to alleviate the computational cost of classical MCMC methods, many efforts have been dedicated to the development of stochastic gradient MCMC methods \citep{welling2011bayesian,ahn2012bayesian,ding2014bayesian,chen2014stochastic,ma2015complete,li2016preconditioned,zhang2019cyclical}. Here, the Metropolis-Hastings correction step is omitted and the gradient of the potential function is approximated by a stochastic, mini-batched, gradient:
\begin{align*}
\widehat{\nabla} U(\btheta) = - \frac{N}{|\mathcal{B}|} \sum_{i \in \mathcal{B}} \nabla U_i(\btheta) - \nabla\log(p(\btheta))\,.
\end{align*}
Despite being computationally more efficient than traditional MCMC, stochastic gradient MCMC methods introduce asymptotic bias. In this work, we consider the stochastic gradient Lagenvin dynamics (\textbf{SGLD}) and Hamiltonian Monte Carlo (\textbf{SGHMC}). We also consider variants of SGMCMC methods that include variance reduction techniques \citep{baker2019control,dubey2016variance}, where the stochastic gradient $\widehat{\nabla}U$ is replaced by the following estimation:
\begin{align*}
\widetilde{\nabla}U(\btheta) = \nabla U(\btheta) |_{\btheta=\eta} + \widehat{\nabla} U(\btheta) - \widehat{\nabla} U(\btheta) |_{\btheta=\eta}\,.
\end{align*}
We then consider the variants \textbf{SGLD}-\textbf{CV} and \textbf{SGHMC}-\textbf{CV} where $\mathbf{\eta}$ is set to a MAP estimate $\btheta_{\mathrm{MAP}}$, together with the variants \textbf{SGLD}-\textbf{SVRG} and \textbf{SGHMC}-\textbf{SVRG} where $\mathbf{\eta}$ is first set to a MAP estimate, and then updated to the current value of $\btheta$ every $m$ iterations. Finally, we also evaluate the preconditioned method \textbf{pSGLD} of \citet{li2016preconditioned}, and the cyclical algorithms \textbf{C}-\textbf{SGLD} and \textbf{C}-\textbf{SGHMC} that rely on a cyclical step size \citep{zhang2019cyclical}.

\paragraph{Gaussian approximations.}
We consider two Gaussian approximations: \textbf{SWAG} and the Laplace approximation with a kronecker factored log likelihood Hessian approximation (\textbf{LA-KFAC}), which both require a pre-trained neural network with parameters $\btheta_{\mathrm{MAP}}$. The \textbf{SWAG} approximation is constructed by collecting values of the parameters along a SGD trajectory with a possibly high step size. The \textbf{SWAG} approximation reads as $p(\btheta|\bX, \bY) \approx \mathcal{N}(\btheta|\btheta_{\mathrm{SWAG}}, \Sigma_{\mathrm{SWAG}})$ where $\btheta_{\mathrm{SWAG}}$ is given by the running mean of the SGD iterates, and $\Sigma_{\mathrm{SWAG}}$ is a diagonal plus low rank approximation. The \textbf{LA-KFAC} approximation is given by $p(\btheta|\bX, \bY) \approx \mathcal{N}(\btheta|\btheta_{\mathrm{MAP}},\Sigma_{\mathrm{KFAC}})$ where $\Sigma_{\mathrm{KFAC}}$ denotes the Kronecker-factored approximate
curvature approximation (see, \textit{e.g.}, \citet{daxberger2021laplace}).

\paragraph{Variational methods.}
The Monte Carlo dropout method (\textbf{MC}-\textbf{Dropout}) proposed by \citet{gal2016dropout} is considered herein. The neural network is augmented with dropout layers which are activated during both training and inference stages. The dropout mechanism is here applied to the output features of each layer.

\paragraph{Deep ensembles.}
We also investigate the deep ensembles method of \citet{lakshminarayanan2017simple,fort2019deep}, which consists in training several neural networks independently with random initializations, and gathering their predictions to obtain a mean prediction and uncertainties.

\section{Evaluation metrics} \label{sec:evaluation_metrics}
\paragraph{Validity of the confidence intervals.}
We assess the validity of the confidence intervals produced by an approximation method with coverage probabilities. There are several related notions of coverage probabilities such as the prediction interval coverage probability, marginal coverage probability, and the conditional coverage probability \citep{lin2021locally}. Given a target confidence level $1-\alpha$, the confidence interval $\hat{C}_{\alpha}^{\mathcal{D}}$ is said to have a valid \textit{prediction interval coverage probability} (PICP) if
\begin{align*}
\bP\{\bY^{\star}\in \hat{C}_\alpha^{\mathcal{D}}(\bX^{\star}) | \mathcal{D} \} \geq 1 - \alpha,
\end{align*}
where the probability is taken only over a test data $(\bX^{\star}, \bY^{\star})$. This coverage probability does not take into account the variability induced by the training dataset $\mathcal{D}$. A more suited property would be the \textit{marginal coverage probability} (MCP):
\begin{align*}
\bP\{\bY^{\star} \in \hat{C}_\alpha^{\mathcal{D}}(\bX^{\star})\} \geq 1 - \alpha,
\end{align*}
where the probability is taken over to both the training data $\mathcal{D}$ and the test data $(\bX^{\star}, \bY^{\star})$. An even stronger property can be obtained by conditioning on the test input data $\bX^{\star}$, leading to the \textit{conditional coverage probability} (CCP):
\begin{align*}
\bP\{\bY^{\star} \in \hat{C}_\alpha^{\mathcal{D}}(\bX^{\star}) | \bX^{\star} = \bx \} \geq 1 - \alpha,\,
\end{align*}
for almost all $\bx \in \mathcal{X}$, where the probability is taken over the training dataset $\mathcal{D}$.

\paragraph{Distance to the HMC reference (weight and function space).}
We compute the maximum mean discrepancy (MMD) between each approximation and the approximation obtained via exhaustive \textbf{HMC}. Let $k : \mathbb{R}^d\times\mathbb{R}^d \rightarrow \mathbb{R}$ be a kernel function and let $\mathcal{H}(k)$ be the reproducing kernel Hilbert space with kernel $k$. The MMD between two probability measures $\bQ$ and $\bQ^{\prime}$ is defined as \citep{gretton2006kernel}:
\begin{align*}
\mathrm{MMD}(\bQ,\bQ^{\prime}) = \| \mu_{\bQ} - \mu_{\bQ^{\prime}} \|_{\mathcal{H}(k)}\,, \quad \mu_{\bQ} = \int k(\cdot,\btheta)\,d\bQ\,,
\end{align*}
where $\mu_{\bQ}$ is called the kernel mean embedding of $\bQ$ in $\mathcal{H}(k)$. Here $k$ is chosen as the distance-based kernel function $k(\btheta,\btheta^{\prime}) = \|\btheta\|_2 + \|\btheta^{\prime}\|_2 - \|\btheta - \btheta^{\prime}\|_2$ proposed by \citet{sejdinovic2013equivalence}. In this setting, $\mathrm{MMD}(\bQ,\bQ^{\prime}) = 0$ implies that $\bQ = \bQ^{\prime}$. For a given training dataset $\mathcal{D}$ and a set of hyperparameters, we compute the MMD between each approximation $\hat{\bQ}$ in weight space obtained by \textbf{SGCMCMC}, \textbf{SWAG}, \textbf{LA-KFAC}, \textbf{MC-Dropout}, or \textbf{Deep ensembles}, and the approximation $\bQ_{\mathrm{HMC}}$ obtained via exhaustive \textbf{HMC} that is considered as the reference. In addition, given a sample $\{ \btheta_i \}_{i=1}^{m}$ from an approximation $\hat{\bQ}$, we can deduce the predictions of the neural network $\bx \mapsto f(\bx;\btheta)$ for each algorithm and compute the MMD distance to the HMC reference in function space.


\paragraph{Distance to the target posterior (weight space).}
We also assess the performance of an approximation method by measuring a distance to the target posterior distribution. We rely on the kernelized Stein discrepancy which, in contrast to other distances, only requires the knowledge of the gradient of the log-posterior distribution. The squared KSD between the target posterior measure $\bP$ and any other probability measure $\bQ$ is defined as
\begin{align*}
\mathrm{KSD}^2(\bP,\bQ) = \mathbb{E}_{\btheta \sim \bQ}\mathbb{E}_{\btheta^{\prime}\sim \bQ} k_p(\btheta,\btheta^{\prime})\,,
\end{align*}
where $k_p$ denotes the Stein kernel given by
\begin{align*}
& k_p(\btheta,\btheta^{\prime}) = \langle \nabla_{\btheta}, \nabla_{\btheta^{\prime}} k(\btheta,\btheta^{\prime}) \rangle + \langle s_p(\btheta), \nabla_{\btheta^{\prime}}k(\btheta,\btheta^{\prime}) \rangle \\
& + \langle s_p(\btheta^{\prime}), \nabla_{\btheta} k(\btheta,\btheta^{\prime}) \rangle + \langle s_p(\btheta), s_p(\btheta^{\prime}) \rangle k(\btheta,\btheta^{\prime})\,.
\end{align*}
Here, $s_p$ denotes the score function of the target posterior distribution, namely, $s_p(\btheta) = \nabla_{\btheta} \log(p(\bY|\bX,\btheta)p(\btheta))$. The Stein kernel also depends on an additional kernel function, $k$, which has to be carefully chosen. Based on the theoretical results of \citet{gorham2017measuring}, the kernel $k$ is chosen as the inverse multi-quadratic (IMQ) kernel, which is defined as $k(\btheta,\btheta^{\prime}) = (1 + \|\btheta-\btheta^{\prime}\|_{\Gamma})^{-1/2}$ with $\Gamma = \ell^2 I$. In this setting, the KSD defines a distance between two probability measures such that $\KSD(\bP,\bQ) = 0$ implies that $\bP = \bQ$. The lengthscale $\ell$ of the IMQ kernel is chosen as the median of the pairwise distances, estimated with a subsample of the entire collection of samples generated by all the approximation methods.

\paragraph{Similarities between the algorithms (weight and function space).}
We also use the MMD in order to establish possible similarities between the algorithms. More precisely, the MMD is computed between each pair $(\hat{\bQ},\hat{\bQ}^{\prime})$ of approximations obtained with the considered approximation methods and set of hyperparameters. The resulting matrix of pairwise MMD distances is visualized thanks to multidimensional scaling \citep{torgerson1952multidimensional} and analyzed to highlight any similarities between the approximation methods. We also proceed the same in function space.

\section{Experimental setup}
The performances of the selected algorithms are studied for several regression tasks. In this section, we summarize the considered datasets and neural network architectures. For each algorithm, we also study the influence of some hyperparameters, which are summarized in the sequel of this section. Additional details about the experiment setup have been reported in the Supplementary Material.

\paragraph{Regression tasks.}
We consider $4$ synthetic regression problems where the output is one-dimensional but the input may be one or multi-dimensional. For each synthetic regression problem, we generate $N_{\mathcal{D}} = 500$ independent datasets $\mathcal{D}_1, \dots, \mathcal{D}_{N_{\mathcal{D}}}$ used for MCP and CCP computations. We also consider an additional test dataset $\mathcal{D}^{\star} = \{(\bX_i^{\star}, Y_i^{\star})\}_{i=1}^{N^{\star}}$ for accuracy, PICP, MCP and CCP. For some problems it may have out-of-distribution (OOD) samples. Each element $(\bX_i, Y_i)$ of a training dataset $\mathcal{D}_j$ is such that $Y_i = f(\bX_i) + \varepsilon_i$, where $\epsilon_i \sim \mathcal{N}(0,\sigma^2)$, $i = 1,\dots,N$. The underlying latent regression function $f$ and the variance of the noise $\sigma$ are both known. The four regression problems are constructed with analytical functions, which are summarized in Table \ref{tab:datasets} below. More details about the regression problems and their generative process can be found in Appendix \ref{app:datasets_regression_problems}.

\begin{table}[h]
\setlength{\tabcolsep}{1.1pt}
\centering
\caption{Description of the synthetic datasets where AF stands for analytical functions.}
\label{tab:datasets}
\begin{tabular}{|c|c|c|c|c|c|c|}
\hline
Task & Latent function & $\sigma$ & $D$ & $N$ & $N^{\star}$ & OOD \\
\hline
AF\#1 & $\cos(2x) + \sin(x)$ & $0.2$ & $1$ & $100$ & $200$ & \xmark \\
AF\#2 & $0.1x^2$ & $0.25$ & $1$ & $100$ & $200$ & \cmark \\
AF\#3 & $-(1+x)\sin(1.2x)$ & $0.25$ & $1$ & $82$ & $200$ & \cmark \\
AF\#4 & MLP$(\cdot; \btheta)$, $\btheta \sim \mathcal{N}(0,\mathbf{I})$ & $0.02$ & $2$ & $120$ & $120$ & \cmark \\
\hline
\end{tabular}
\end{table}

\paragraph{Neural network architectures.} A feed-forward neural network with ReLU activations is used as a surrogate model for every regression problem. The number of parameters ranges from $2,651$ to $20,501$ (see Appendix \ref{app:experiments} for more details). We use a centered normalized Gaussian prior distribution for the weights in all the experiments. Note that for each training data $\mathcal{D}_j$, $j = 1, \dots, N_{\mathcal{D}}$, a MAP estimate is computed by training the neural network with an Adam optimizer and an exponentially decaying learning rate. These $N_{\mathcal{D}}$ MAP estimates are subsequently used in \textbf{LA-KFAC}, \textbf{SWAG}, \textbf{SGMCMC-CV}, and \textbf{SGMCMC-SVRG}.
\paragraph{Hyperparameters.}
Besides the $\textbf{LA-KFAC}$ approximation, all the algorithms depend on one or several hyperparameters. For all the remaining methods (\textbf{SGMCMC}, \textbf{Deep ensembles}, \textbf{MC}-\textbf{Dropout}, and \textbf{SWAG}), we study the influence of the step size $\epsilon > 0$ (\textit{i.e.}, the learning rate) by considering $10$ equally log-spaced values $\epsilon_1 < \dots < \epsilon_{10}$. The ranges are carefully chosen for each type of algorithm and are reported in Appendix \ref{app:implementation}. We also investigate the influence of two additional hyperparameters. Five dropout rates in the \textbf{MC}-\textbf{Dropout} method are considered ($0.1, 0.2, \dots, 0.5$) and three cycle lengths in the \textbf{C}-\textbf{SGLD} and \textbf{C}-\textbf{SGHMC} methods are investigated ($10$, $100$, and $1000$). Other hyperparameters are held fixed, see Appendix \ref{app:hyperparameters} for more details.

\paragraph{HMC reference.}
For a given training dataset $\mathcal{D}$, a reference sample is generated by Hamiltonian Monte Carlo and subsequently used to evaluate the performance of the selected algorithms (see section \ref{sec:evaluation_metrics}). We run 3 HMC chains of 200 iterations, discard the first 100 iterations as burn-in, and perform $10,000$ leapfrogs steps. The step size is selected such that the Metropolis-Hastings accept rates are at least above $80\%$.

\section{Regression without OOD testing} \label{sec:without_ood_testing}
We first present the results for the first experiment (AF\#1 in Table \ref{tab:datasets}). Here, the training and test datasets are sampled from the same distribution so that any test input lies within the range of the input training data points. 

\begin{figure}[h]
\centering
\includegraphics[width=0.42\textwidth]{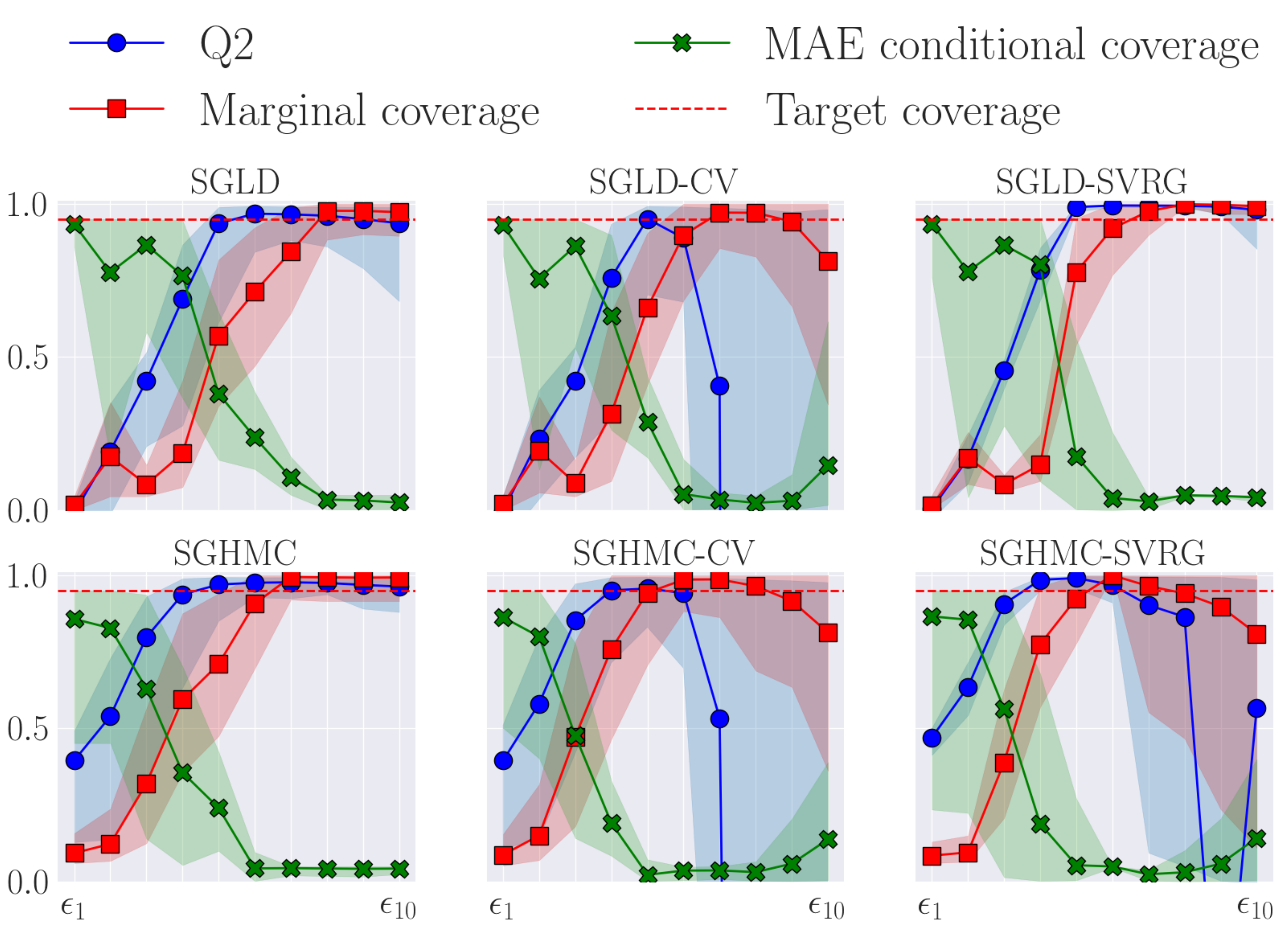}
\includegraphics[width=0.42\textwidth]{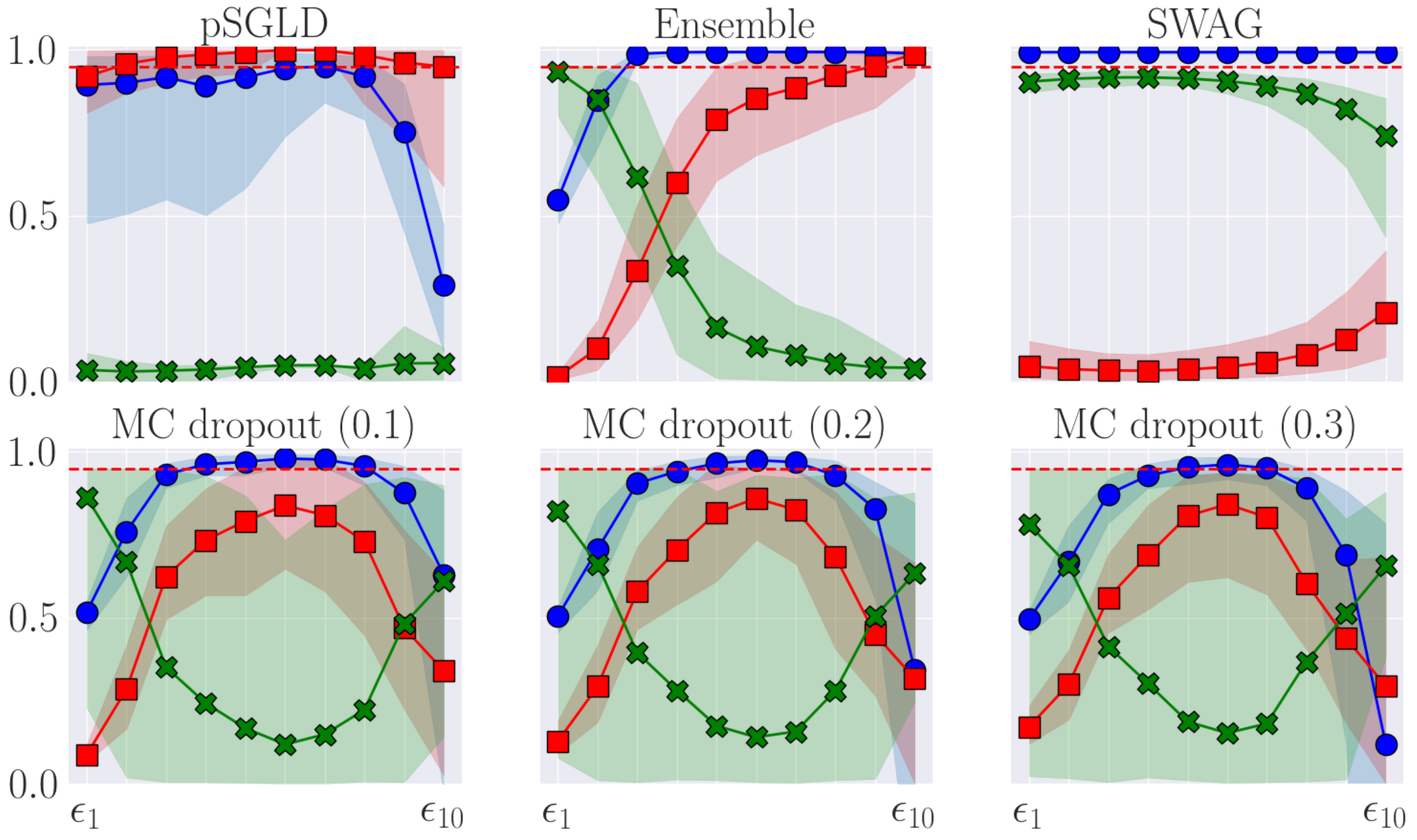}
\caption{Problem AF\#1. Coverage metrics and $Q^2$ coefficient with respect to the step size $\epsilon$. The target coverage is set to $0.95$. Results obtained for the cyclical \textbf{SGMCMC} variants are reported in the Appendix.}
\label{fig:q2_coverage_af1_sgmcmc.pdf}
\end{figure}

\paragraph{Coverage probabilities.}
Figure \ref{fig:q2_coverage_af1_sgmcmc.pdf} gathers the graphs of the marginal coverage probabilities and the mean absolute error in conditional coverage with respect to the step size $\epsilon$ for a target coverage of $0.95$. In addition, the coefficient of determination on the testing set, denoted by $Q^2=1-\sum_{i=1}^n \left(Y_i^{\star}-\hat{f}(\bX_i^{\star};\btheta)\right)^2/\textrm{Var}\; Y^{\star}$, is also reported. It can be observed that 
\begin{itemize}
\item[(a)] The \textbf{SGLD} and \textbf{SGHMC} methods yield similar performances.
\item[(b)] The performance of the variants with control variates (\textbf{SGLD-CV} and \textbf{SGHMC-CV}) drops significantly for high step sizes. As pointed out by \citep{nemeth2021}, if the state of the Markov chain gets far from the control variate (here a MAP estimate), then the variance of the stochastic gradient can increase instead of being reduced, hence explaining the observed behavior. 
\item[(c)] Updating the control variates seems to address this issue as illustrated by the performances of \textbf{SGLD-SVRG}, but \textbf{SGHMC-SVRG} is still affected by high step sizes. Given that \textbf{SGHMC} mixes better than \textbf{SGLD}, we believe that this behavior can be corrected by increasing the frequency at which the control variates are updated in \textbf{SGHMC-SVRG}. 
\item[(d)] The preconditioned SGLD method easily reaches the target coverage probability but has lower coefficients of determination than the other methods.
\item[(e)] \textbf{Deep ensembles} and \textbf{SWAG} both yield high coefficient of determinations but only \textbf{deep ensembles} is able to reach the target coverage probability. 
\item[(f)] \textbf{MC}-\textbf{dropout} has the lowest performances in this example. In particular, the high variability of the absolute error in conditional coverage suggests that the confidence intervals are not smooth with respect to the input space. 
\item[(g)] Finally, for the \textbf{LA-KFAC} method we obtain $Q^2=0.99 \pm 0.004$, $\mathrm{MCP}=0.99 \pm 0.01$, and $\mathrm{MAE}=0.048 \pm 0.003$. As a result, the mean prediction of \textbf{LA-KFAC} is very accurate but the high coverage probability suggests that it overestimates the uncertainties.
\end{itemize}

\begin{figure}[h!]
\centering
\includegraphics[width=0.48\textwidth]{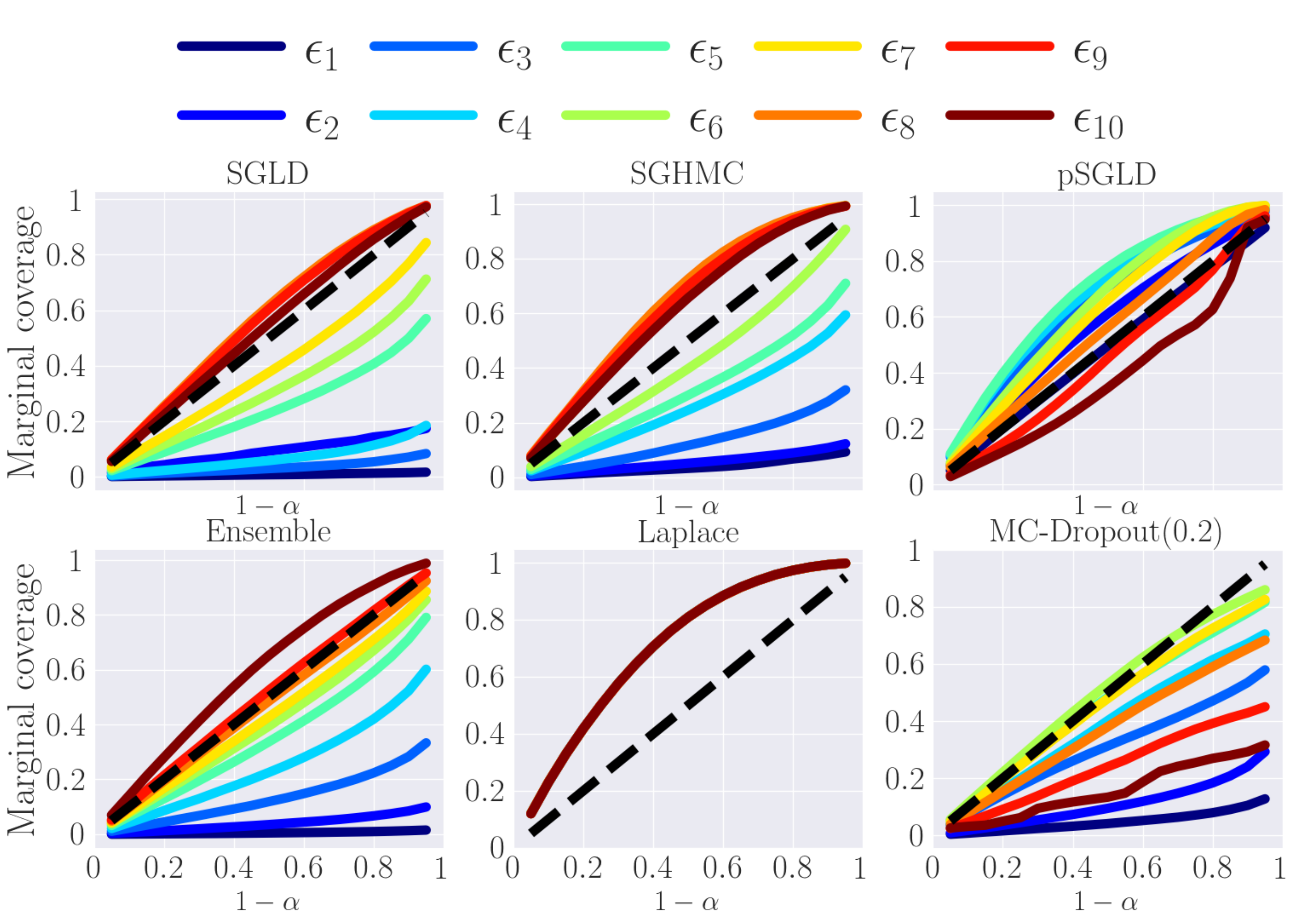}
\caption{Problem AF\#1. Graphs of the marginal coverage probability with respect to the target level and for $10$ step sizes.}
\label{fig:coverage_cherry_pick}
\end{figure}

We also investigate the marginal coverage probability with respect to various target levels $(1-\alpha)$, $\alpha \in ]0,1[$, which are shown in Figure \ref{fig:coverage_cherry_pick} for $6$ selected algorithms (see also Appendix \ref{app:experiments}) . Here, we see that \textbf{pSGLD} and \textbf{LA-KFAC} easily overestimate the target level regardless of the step size. Finally, we report the best marginal coverage probabilities (closest to $0.95$) and the best $Q^2$ coefficients (closest to $1$) for each algorithm in Table \ref{tab:best_mcp_q2}. For each best coverage (resp. $Q^2$), we also report the associated $Q^2$ (resp. coverage). Interestingly, we observe that the best coverages lying in $0.95 \pm 0.1$ do not especially correspond to the best regression coefficients $Q^2$. Most notably, \textbf{Deep ensembles} is able to achieve the best performances in order of both coverage and prediction accuracy.

\begin{table*}[h]
    \setlength{\tabcolsep}{1pt}
    \renewcommand{\arraystretch}{1}
    \centering
    \caption{Best marginal coverage probabilities (MCP) and best $Q^2$ coefficients obtained with each algorithm.}
\begin{tabular}{|l|cc|cc||l|cc|cc|}
   \hline\hline 
    Method & Best MCP & $Q^2$ & Best $Q^2$ & MCP & Method & Best MCP & $Q^2$ & Best $Q^2$ & MCP \\ 
   \hline 
   SGLD & 0.97 & 0.93 & 0.96 & 0.71 & CSGHMC(10) & 0.97 & 0.97 & 0.97 & 0.89 \\
   SGLD-CV & \textbf{0.94} & $-10^3$ & 0.95 & 0.66 & CSGHMC(100) & \textbf{0.94} & 0.97 & 0.97 & 0.94 \\
   SGLD-SVRG & 0.97 & 0.99 & \textbf{0.99} & 0.91 & CSGHMC(1000) & \textbf{0.94} & 0.96 & 0.97 & 0.93 \\
   SGHMC & 0.99 & 0.96 & 0.97 & 0.99 & pSGLD & \textbf{0.94} & 0.29 & 0.95 & 0.99 \\
   SGHMC-CV & \textbf{0.94} & 0.95 & 0.95 & 0.94 & Deep ensemble & \textbf{0.95} & 0.99 & \textbf{0.99} & 0.88 \\
   SGHMC-SVRG & \textbf{0.94} & 0.86 & \textbf{0.99} & 0.92 & SWAG & 0.20 & 0.99 & \textbf{0.99} & 0.03 \\
   CSGLD(10) & \textbf{0.95} & 0.96 & 0.96 & 0.66 & MC Drop.(0.1) & 0.83 & 0.98 & \textbf{0.98} & 0.83 \\
   CSGLD(100) & \textbf{0.94} & 0.96 & 0.96 & 0.76 & MC Drop.(0.2) & 0.85 & 0.97 & 0.97 & 0.85 \\
   CSGLD(1000) & 0.92 & 0.95 & 0.96 & 0.75 & MC Drop.(0.3) & 0.84 & 0.96 & 0.96 & 0.84 \\
   \hline\hline
\end{tabular}
    \label{tab:best_mcp_q2}
\end{table*}

\begin{figure}[h!]
\centering
\includegraphics[width=0.38\textwidth]{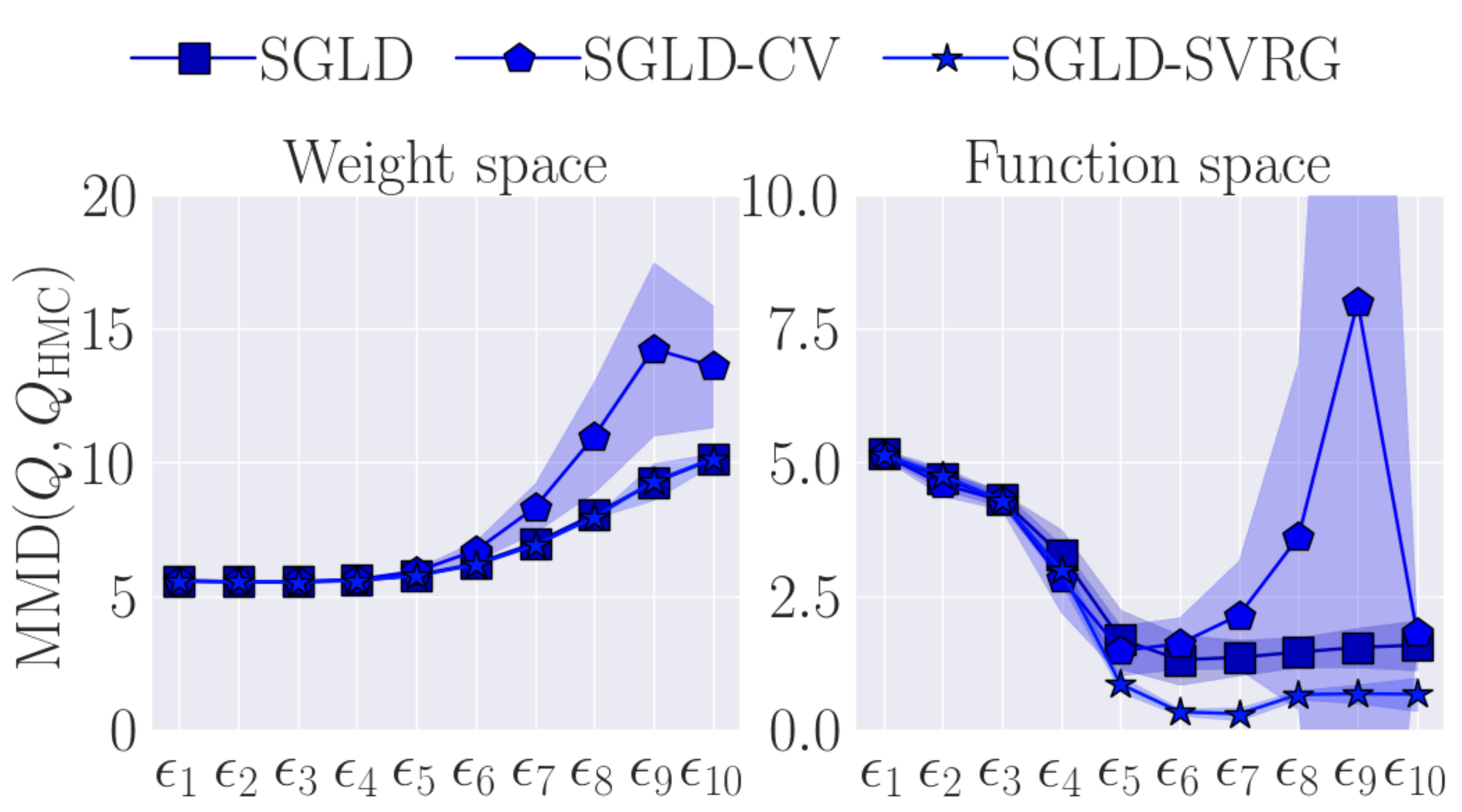}
\includegraphics[width=0.38\textwidth]{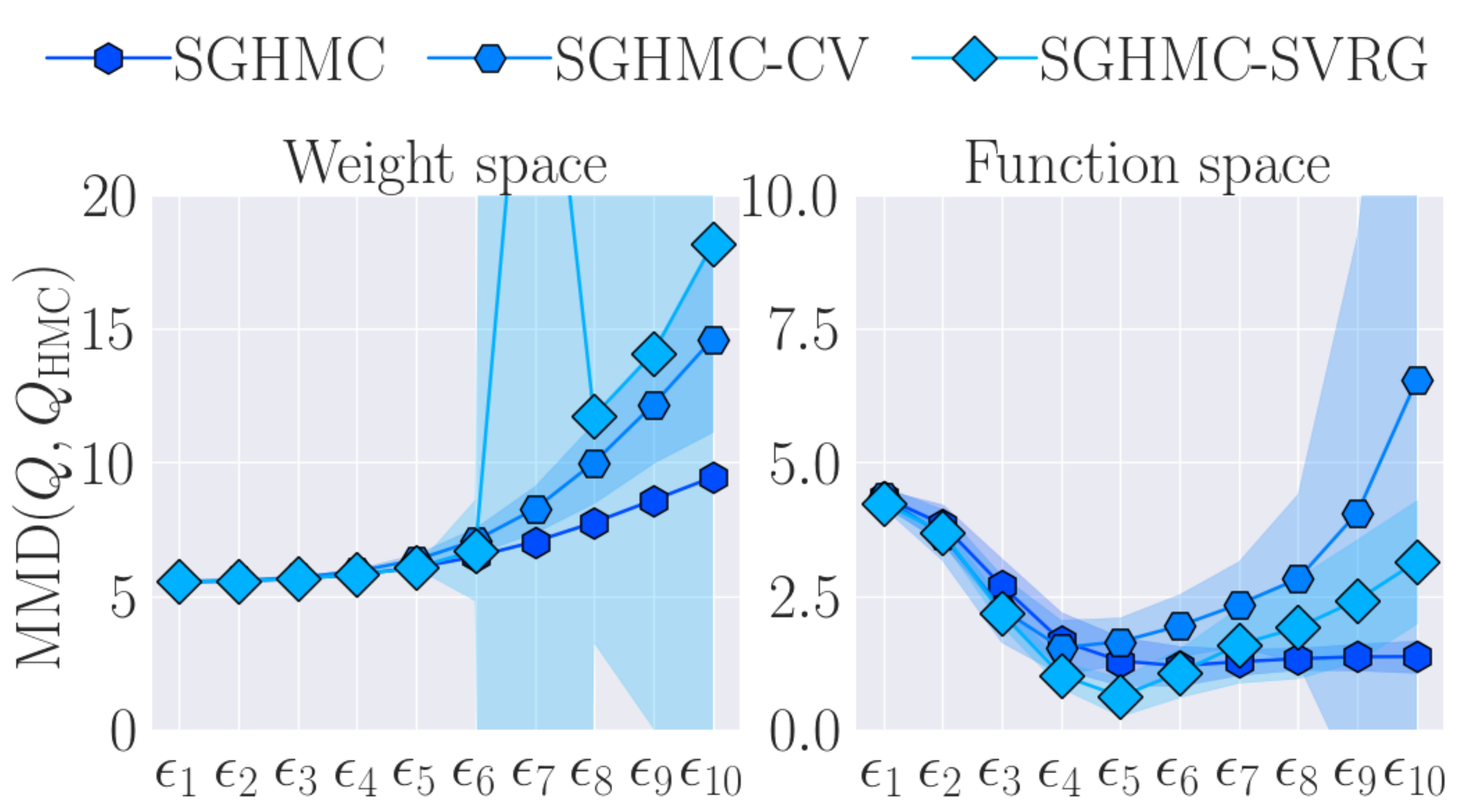}
\includegraphics[width=0.38\textwidth]{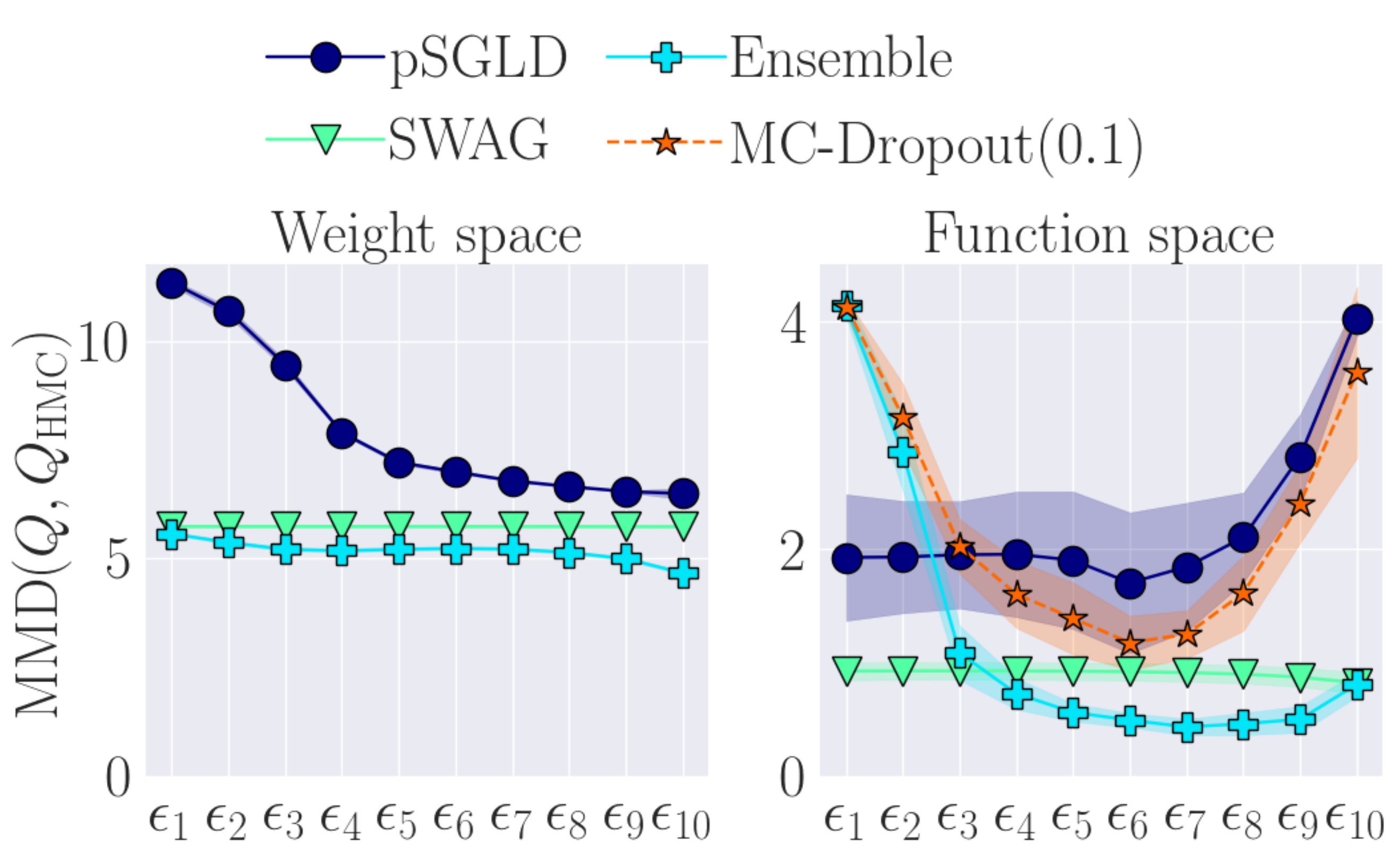}
\caption{Problem AF\#1. Maximum mean discrepancies $\mathrm{MMD}(\bQ,\bQ_{\mathrm{HMC}})$ between the approximated posterior distributions $\bQ$ and the reference HMC sample, in weight and function spaces.}
\label{fig:mmd_weight_function_spaces}
\end{figure}

\paragraph{MMD distance to the HMC reference.}
The distances to the HMC reference sample are shown in Figure \ref{fig:mmd_weight_function_spaces}. In weight space, the MMD distance increases with the step size for most algorithms. \textbf{Deep ensembles} is the only method for which the MMD distance in weight space decreases below $5$. In contrast, the MMD distances in function space may decrease towards $0$. In particular, the \textbf{SGLD-SVRG}, \textbf{deep ensembles}, and \textbf{SWAG} methods yield the lowest MMD distances in function space. With the \textbf{LA-KFAC} method, we obtained MMD distances of $5.76 \pm 0.13$ in weight space, and of $1.27 \pm 0.21$ in function space. Overall and perhaps surprisingly, the behavior with respect to the step size is not the same in weight and function space.



\begin{figure}[h]
\centering
\includegraphics[width=0.42\textwidth]{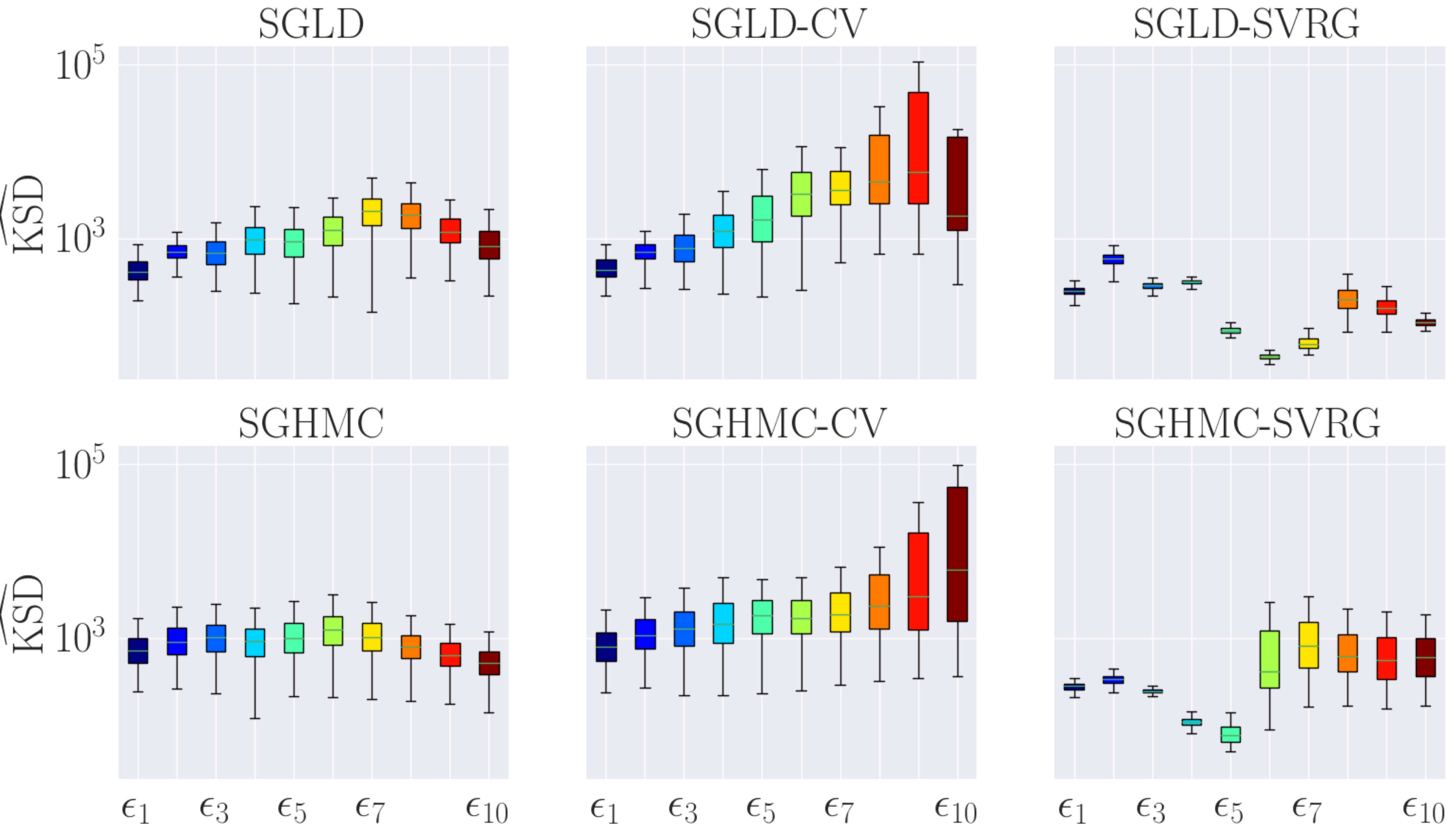}
\includegraphics[width=0.42\textwidth]{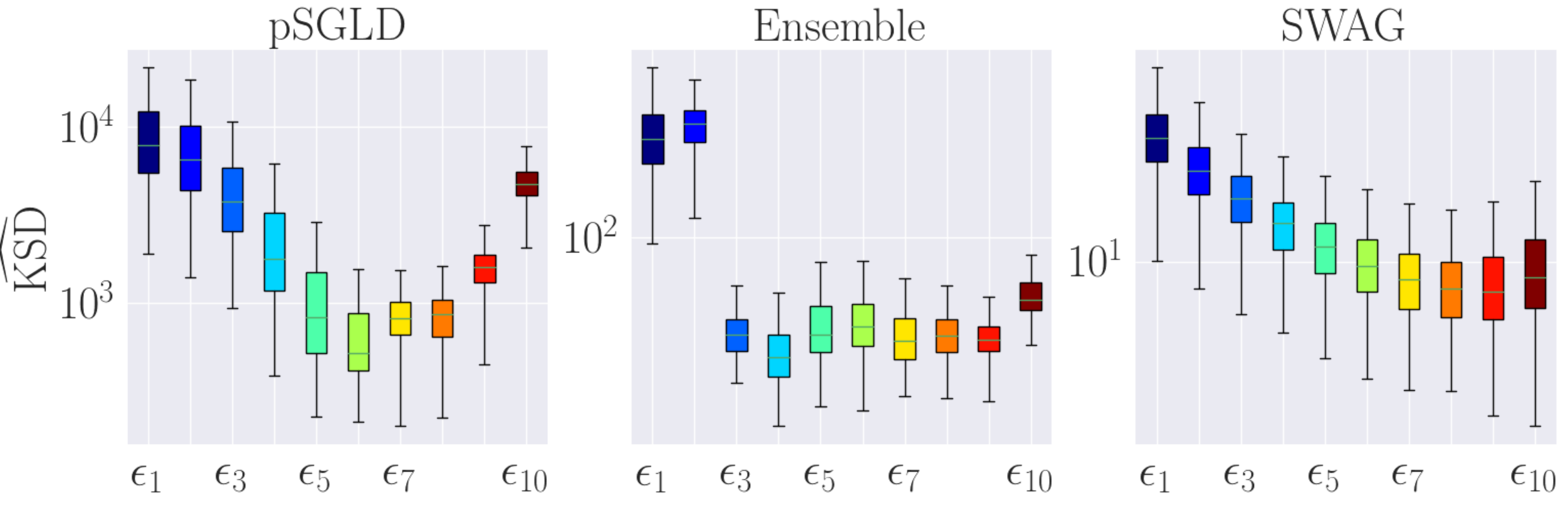}
\caption{Problem AF\#1. Kernelized Stein discrepancies $\mathrm{KSD}(\bP,\bQ)$ between the target posterior measure $\bP$ and the approximated posteriors $\bQ$. KSD values for the cyclical \textbf{SGMCMC} methods are reported in the Appendix.}
\label{fig:ksd_af1}
\end{figure}

\paragraph{KSD distance to the target posterior.}
The kernelized Stein discrepancies between the target posterior $\bP$ and the approximations $\bQ$ are shown in Figure \ref{fig:ksd_af1}. Here, the lowest discrepancies are obtained with \textbf{deep ensembles} and \textbf{SWAG}, while \textbf{SGMCMC} methods yield higher KSDs. Surprisingly, the \textbf{LA-KFAC} method yields the highest KSDs, with values between $10^5$ and $10^7$. If for some algorithms the conclusions from the MMD and the KSD coincide, this is not the case in general: we discuss below some potential explanations related to KSD limitations.

\begin{figure}[h!]
\centering
\includegraphics[width=0.43\textwidth]{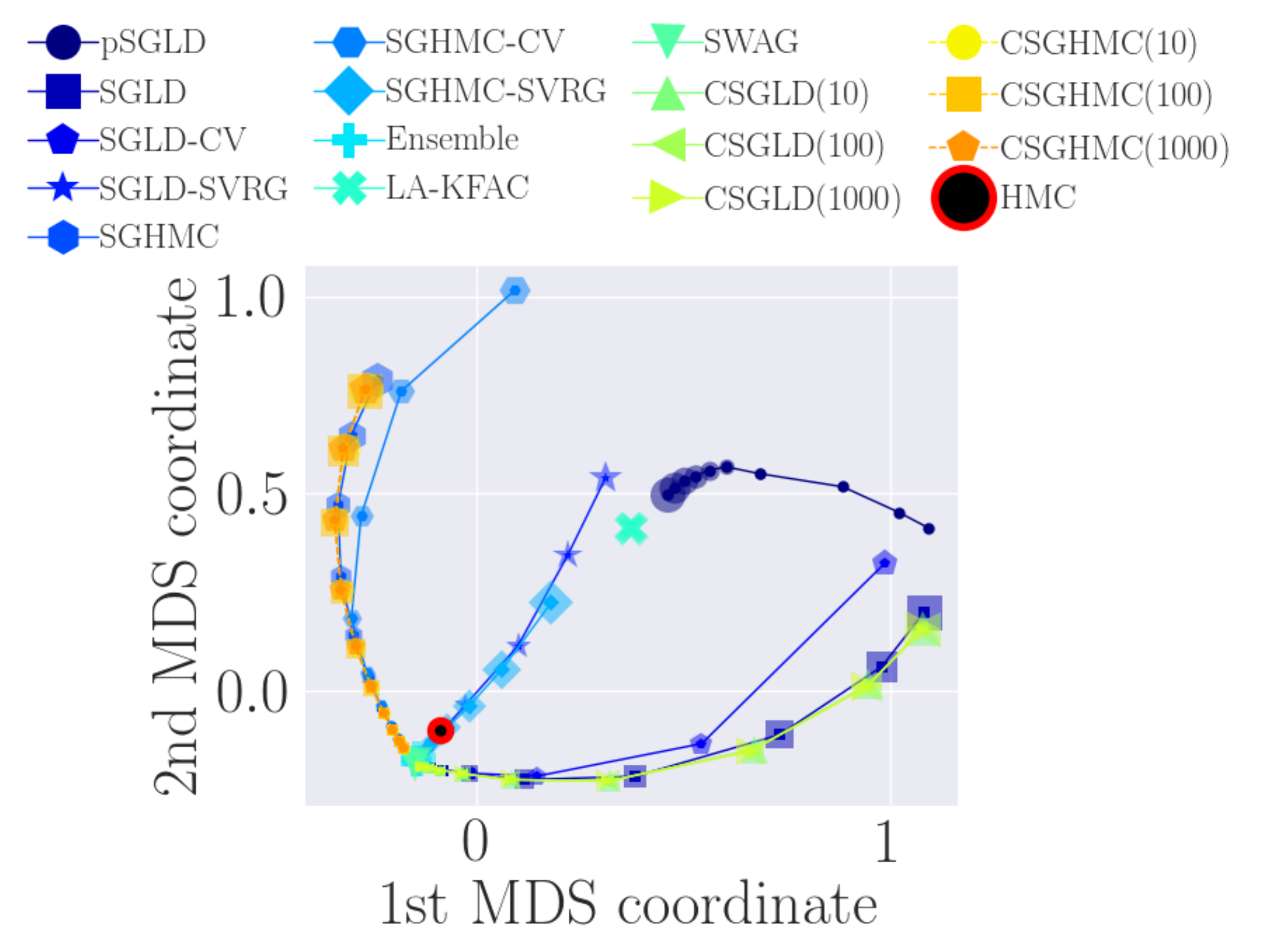}
\caption{Dataset AF\#1. Similarities between the algorithms as measured by the MMD in weight space and represented in a two-dimensional space build with multidimensional scaling. The markers sizes are proportional to the value of the step size $\epsilon$.} 
\label{fig:similarities_af1}
\end{figure}

\paragraph{Similarities between the algorithms.}
The similarities between the algorithms is depicted in Figure \ref{fig:similarities_af1}. At least four groups of approximations can be distinguished in weight space (left panel in Figure \ref{fig:similarities_af1}): \textbf{SGLD}, \textbf{SGLD-CV} and \textbf{CSGLD} methods, \textbf{SGHMC}, \textbf{SGHMC-CV} and \textbf{CSGHMC}, \textbf{pSGLD} and \textbf{LA-KFAC}, and the remaining methods form the last group. Amongst the \textbf{SGLD} and \textbf{SGHMC} methods, it is clearly observed that \textbf{CV} variants yield distinct approximations. Surprisingly, \textbf{SGLD-SVRG} and \textbf{SGHMC-SVRG} generate similar approximations to \textbf{deep ensembles}, in weight space. The similarities in function space do not exihibit any particular structure and have been reported in Appendix \ref{app:similarities_afi} (see Figure \ref{app:similarities_afi}).

\paragraph{Discussion.} Based on the results presented in this section, a few observations can be made. First, we see that low KSD values do not especially correspond to good coverage probabilities and $Q^2$ coefficients. This can be seen in the case of \textbf{SWAG}, which achieves the lowest KSD values but poor coverage probabilities. Amongst the \textbf{SGMCMC} methods, it can also be seen that the lowest KSD values are often obtained for the lowest step sizes. As a result, the KSD seems often unable to identify approximation methods that generate valid confidence intervals. Several studies have shown that the KSD suffers from strong pathologies in simple experiments \citep{wenliang2020blindness, korba2021kernel}. In particular, the KSD is known to be insensitive to weight proportions in multimodal distributions, which may explain our observations. 

A similar behavior is observed with the MMD distance to the HMC reference in weight space. For instance, this distance increases with the step size for \textbf{SGLD} but the best predictive performances are obtained for the highest step sizes. In contrast, the MMD distance to the HMC reference in function space is correlated to the predictive performance. A low MMD distance in function space typically corresponds to the best coverage probabilities and $Q^2$ coefficients obtained by a given approximation method.

Finally, it can be observed that in this experiment, \textbf{deep ensembles}, \textbf{SGLD-SVRG}, and \textbf{SGHMC-SVRG} provide similar approximations as shown by the similarity measures in Figure \ref{fig:similarities_af1}, the MMD distances in function space, and the performances in terms of coverage probabilities and mean predictions. While the \textbf{LA-KFAC} method yields the highest KSD values and higher MMD distances than other methods, it offers a high precision on the test dataset, and a high coverage probability way above the target level.

\section{Regression with OOD testing} \label{ref:with_ood_testing}
In this section, we present the results obtained for the second regression problem (AF\#2 in Table \ref{tab:datasets}). Here, the distribution of test dataset differs from the distribution of the training dataset, which is more challenging in terms of coverage probabilities and predictive performances. We find that the behavior of the KSD with respect to the step size and approximation method is similar to the one observed in the previous experiment (see Figure \ref{fig:ksd_af1}). Furthermore, we also observe that the similarities between the approximation methods is close to the one observed in Figure \ref{fig:similarities_af1}. All the associated figures are reported in Appendix \ref{app:experiments} for brevity.

\begin{figure}[h]
\centering
\includegraphics[width=0.42\textwidth]{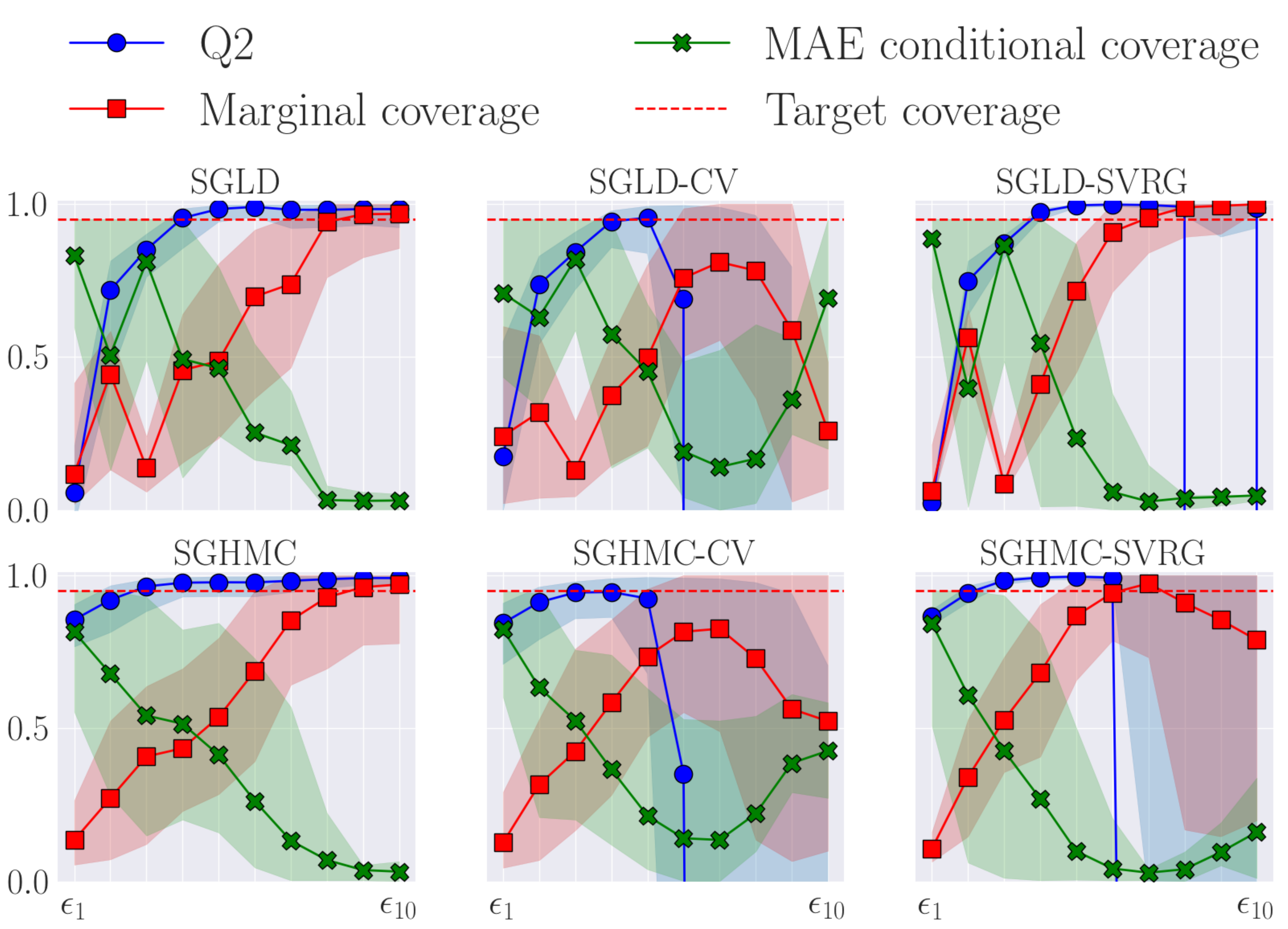}
\includegraphics[width=0.42\textwidth]{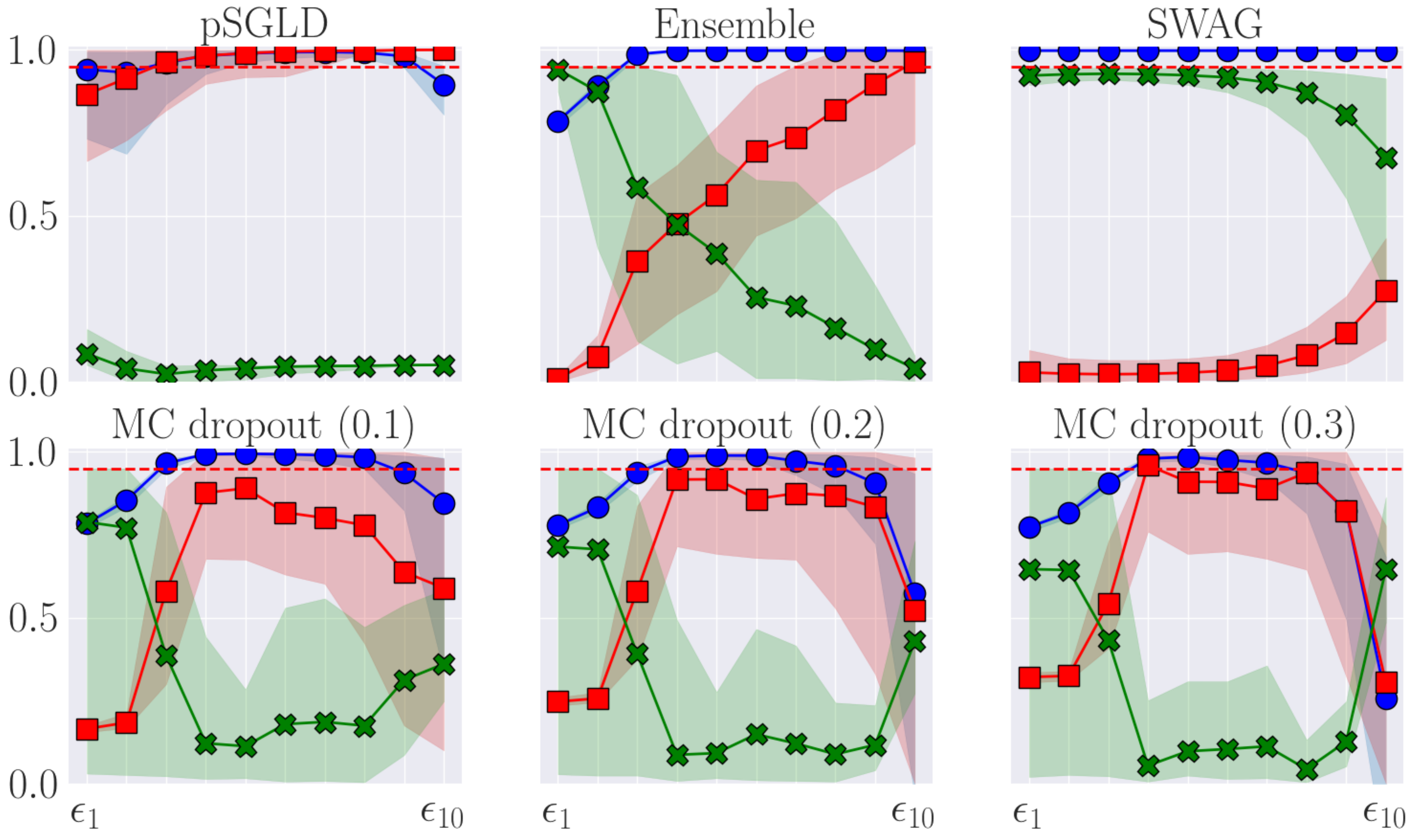}
\caption{Problem AF\#2. Coverage metrics and $Q^2$ coefficient with respect to the step size $\epsilon$. The target coverage is set to $0.95$.}
\label{fig:q2_coverage_af2_sgmcmc.pdf}
\end{figure}

\paragraph{Coverage probabilities.}
The coverage probabilities metrics and the regression coefficients $Q^2$ are reported in Figure \ref{fig:q2_coverage_af2_sgmcmc.pdf}. Compared to the previous experiment of section \ref{sec:without_ood_testing}, we see that achieving the target coverage level is more challenging for most of the approximation methods due to the test inputs that are out of the distribution of the training data. Several \textbf{SGMCMC} methods (such as \textbf{SGLD}, \textbf{SGHMC}, and their cyclical variants) easily yield high regression coefficients but struggle to reach the target confidence level. A similar behavior is observed for \textbf{deep ensembles}. This suggests that the width of the confidence intervals are not big enough to properly cover the test predictions. In this experiment, we see that \textbf{MC-dropout} has slightly better performances than in the previous section but remains less effective than the other approximation methods. Finally, we find that the performances of \textbf{pSGLD} and \textbf{SWAG} remain similar to AF\#1.

\paragraph{MMD distances to the HMC reference.}
In terms of MMD distances, we find that \textbf{deep ensembles} and \textbf{SGLD-SVRG} yield the closest approximations to the HMC reference in function space. For brevity, the graphs of the MMD distances have been reported in Appendix \ref{app:experiments} (see Figure \ref{fig:mmd_weight_function_spaces_af2}). Here again, we find that the MMD distance in weight space increases with the step size while it mainly decreases in function space.


\section{Discussion about coverage probabilities}
\begin{figure}[h]
\centering
\includegraphics[width=0.42\textwidth]{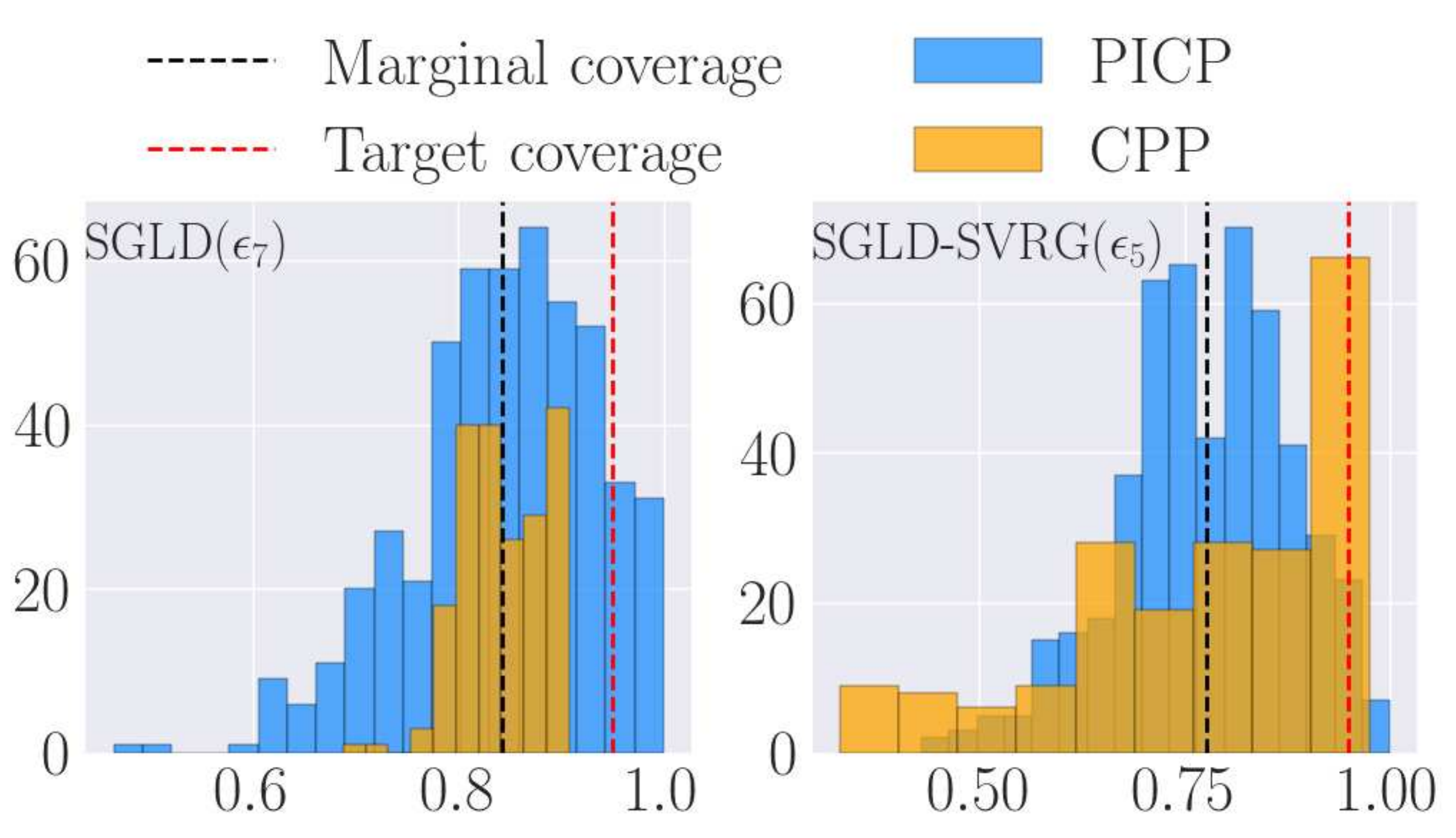}
\caption{Problem AF\#1. Comparison of the three coverage probabilities obtained with \textbf{SGLD} and \textbf{SGLD-SVRG}.}
\label{fig:picp_versus_mcp_ccp}
\end{figure}

Previous works that assess the quality of the confidence intervals in BNNs rely on the PICP (see, \textit{e.g.}, \citet{yao2019quality}). In contrast to MCP and CCP, PICP only integrates over the test set and does not take into account the variability of the confidence interval with respect to the training set. This is convenient in practice given that we generally do not have access to the distribution of the training data. However, PICP may lead to wrong conclusions about the validity of the confidence intervals. To support our claim, in Figure \ref{fig:picp_versus_mcp_ccp} we compare the three coverage metrics: the PICP histogram obtained by considering all the PICPs (mean over test set) for each training set, while the CPP histogram gathers all the CCPs (mean over training set) for each sample from the test set. We also represent the MCP (mean over both training and test set) and the target confidence level. In both cases, the MCP and CPP metrics clearly indicate that the target coverage is far from being attained. On the contrary, the PICP exhibits large variability over the training sets, with a significant probability of being close to the target coverage for \textbf{SGLD} on the left panel. This implies that, depending on the random sampling of the training set, the PICP may lead to consider \textbf{SGLD} as an efficient approximation of the posterior, while in reality the MCP and CPP are way more pessimistic and tend to conclude the opposite. Consequently, we advocate to use the PICP very cautiously.

Since in practice the MCP and CPP cannot be computed by sampling many independent training sets as we did in our experiments, we suggest instead to estimate them by resampling methods such as the bootstrap, in order to compensate for the PICP limitations illustrated above.

\section{Summary}
This work compares several approximation methods for Bayesian neural networks with 14 algorithms including $9$ SGMCMC methods, \textbf{deep ensembles}, \textbf{LA-FKAC}, \textbf{MC-Dropout}, \textbf{SWAG}, and \textbf{HMC} which is taken as a reference. In contract to previous works, we assess the performance of the selected methods by evaluating the validity of the generated confidence intervals, the quality of the approximation in weight and function spaces, and the precision of the mean prediction on test data. We find that \textbf{SGMCMC} methods and \textbf{deep ensembles} are able to achieve good coverage probabilities and predictive performances. Most notably, in our experiments, \textbf{deep ensembles} provide the best approximation to the HMC reference in both weight and function spaces. While \textbf{SWAG} has excellent performances in terms of regression, it tends to underestimate the uncertainties. We also propose a novel similarity analysis between the algorithms which shows that \textbf{LA-KFAC} and \textbf{pSGLD} with high step sizes provide close approximations in weight space. \textbf{SGMCMC-SVRG} and \textbf{deep ensembles} are also very similar in weight space. We finally consider the KSD but unfortunately we observe that it cannot exclusively be relied on to assess the quality of an approximation to the target posterior. In summary, our experiments show that several approximations methods can be efficient, but selecting appropriate hyperparameters yielding valid confidence intervals and regression accuracy is still challenging. A promising way would be to rely on resampling methods on the training set to estimate the MCP and the CCP or use the rencetly introduced conformal prediction \citep{jackknife}. We also hope that the proposed new comparisons in weight and function spaces will give insights on how the different algorithms explore the posterior distribution.


\bibliography{bibfile}
\bibliographystyle{icml2023}

\newpage
\appendix
\onecolumn

\vspace*{0.3cm}
\hspace*{1cm} \textbf{\LARGE{Supplementary Material}}
\vspace*{0.6cm}

\FloatBarrier

\section{Considered algorithms} \label{app:implementation}
We briefly synthesize the considered \textbf{SGMCMC} methods that we consider in our study, with details on software implementation and practical considerations in our numerical experiments.

\subsection{Stochastic gradient Markov chain Monte Carlo algorithms} \label{app:sgmcmc}
Stochastic gradient Markov chain Monte Carlo (SGMCMC) methods are a family of gradient-based algorithms that rely on the resolution of an It\^{o} stochastic differential equation \citep{ma2015complete} (see also \citep{nemeth2021} for a review). Amongst this family of methods, we consider the stochastic gradient Langevin dynamics (SGLD) \citep{welling2011bayesian} and the stochastic gradient Hamiltonian Monte Carlo (SGHMC) \citep{chen2014stochastic} methods, together with other variants that include variance reduction \citep{dubey2016variance,baker2019control} or preconditioning techniques \citep{li2016preconditioned,springenberg2016bayesian}.

\paragraph{Stochastic gradient Langevin dynamics.} The \textbf{SGLD} algorithm proposed by \citet{welling2011bayesian} takes the form
\begin{align*}
k \geq 0\,, \quad \btheta^{k+1} = \btheta^k - \displaystyle\epsilon_k \widehat{\nabla}U(\btheta^k) + \sqrt{2\epsilon_k} \Delta \mathbf{W}^{k+1}\,,
\end{align*}
where $\Delta \mathbf{W}^{k+1}$ is centered and normalized Gaussian random variable. Herein, \textbf{SGLD} is implemented with a constant step size $\epsilon_k = \epsilon$ for all $k \geq 0$.

\paragraph{Stochastic gradient Hamiltonian Monte Carlo.} The \textbf{SGHMC} algorithm has been proposed by \citet{chen2014stochastic} where one leapfrog step reads as:
\begin{equation*}
k \geq 0:
\left\{
\begin{aligned}
& \btheta^{k+1} = \btheta^k + \eta_k \mathbf{M}^{-1} \mathbf{r}^k\,, \\
& \mathbf{r}^{k+1} = \mathbf{r}^k - \eta_k \widehat{\nabla} U(\btheta^k) - \epsilon_k \mathbf{C}\mathbf{M}^{-1}\mathbf{r}^k + \sqrt{2\mathbf{C}\eta_k}\Delta \mathbf{W}^{k+1}\,,
\end{aligned}
\right.
\end{equation*}
where $\mathbf{r}$ denotes the momentum and $\Delta \mathbf{W}^{k+1} \sim \mathcal{N}(0,\mathbf{I})$. By introducing the change of variable $\mathbf{v}^k = \epsilon_k \mathbf{M}^{-1}\mathbf{r}^k$, the above equations can be rewritten as
\begin{equation*}
k \geq 0:
\left\{
\begin{aligned}
& \btheta^{k+1} = \btheta^k + \mathbf{v}^k \,, \\
& \mathbf{v}^{k+1} = (1 - \alpha) \mathbf{v}^k - \epsilon_k \widehat{\nabla} U(\btheta^k) + \sqrt{2\alpha\epsilon_k}\Delta \mathbf{W}^{k+1}\,,
\end{aligned}
\right.
\end{equation*}
where $\epsilon_k = \eta_k^2\mathbf{M}^{-1}$ and $\alpha = \eta\mathbf{M}^{-1}\mathbf{C}$. The step size $\epsilon_k$ is constant in our experiments with \textbf{SGHMC}, and the momentum $\mathbf{v}$ is resampled every $10$ leapfrog steps following a centered and normalized Gaussian distribution.

\paragraph{Preconditioned SGLD.} \cite{ma2015complete} proposed the Riemannian SGLD by building upon the Riemannian manifold Langevin and Hamitonian Monte Carlo methods \cite{girolami2011riemann}. The Riemannian SGLD takes the form:
\begin{equation} \label{eq:rSGLD}
k \geq 0\,, \quad \btheta^{k+1} = \btheta^k - \displaystyle\epsilon_k \left( D(\btheta^k)\widehat{\nabla} U(\btheta^k) + \Gamma(\btheta^k) \right) + \sqrt{2\epsilon_k D(\btheta^k)} \Delta \mathbf{W}^{k+1}\,,
\end{equation}
where $\mathbf{D} \in \mathbb{R}^{d \times d}$ can be seen as a preconditioning matrix and the vector $\mathbf{\Gamma} \in \mathbb{R}^d$ is given by 
\begin{equation}
\mathbf{\Gamma}_i(\btheta) = \sum_j \displaystyle\frac{\partial}{\partial \mathbf{\varsigma}_j} \mathbf{D}_{ij}(\btheta)\,.
\end{equation}
Several forms of the preconditioner $\mathbf{D}$ can be used, the most common choice being a constant diagonal matrix $\mathbf{D}_{\mathrm{diag}}$ of the form $\mathbf{D}_{\mathrm{diag}} = N^{-1}\mathbf{I}_d$, where $N$ denotes the size of the training dataset. The expected Fisher information matrix is also a natural choice but it is usually intractable, and a few alternatives have been proposed in the literature. Herein we consider the \textbf{pSGLD} method of \citet{li2016preconditioned} that neglects the vector $\mathbf{\Gamma}$ and uses a preconditioner inspired by the RMSprop optimization algorithm. The gradient is scaled using a moving average of its norm at each iteration:
\begin{equation*}
\begin{aligned}
& \mathbf{D}(\btheta^k) = \mathrm{diag}\left( \lambda \mathbf{I} + \sqrt{\mathbf{L}(\btheta^k)} \right)^{-1}\,, \\
& \mathbf{L}(\btheta^k) = \alpha \mathbf{L}(\btheta^{k-1}) + (1 - \alpha) \widehat{\nabla} U(\btheta^{k-1}) \circ \widehat{\nabla} U(\btheta^{k-1})\,, 
\end{aligned}
\end{equation*}
where $\alpha$ is a parameter with values in $[0,1]$, $\lambda$ is a regularization constant, and $\circ$ denotes the Hadamard (element-wise) product.

\paragraph{\textbf{SGMCMC-CV} and \textbf{SGMCMC-SVRG}.} As described in section \ref{sec:bnns}, we consider two variants of the \textbf{SGLD} and \textbf{SGHMC} algorithms that rely on control variates to reduce the variance of stochastic approximation $\widehat{\nabla}U$ of $\nabla U$. Following \citet{baker2019control}, \textbf{SGLD-CV} and \textbf{SGHMC-CV} use an estimation of the maximum a posteriori (MAP) for the control variable $\mathbf{\eta}$. Hence, the neural network is first trained in a classical fashion using an optimization algorithm, and the resulting values of the network parameters are stored. The remaining variants \textbf{SGLD-SVRG} and \textbf{SGHMC-SVRG} use the strategy proposed by \citet{dubey2016variance} which consists in setting $\mathbf{\eta}$ to a MAP estimate as well, but then updating $\mathbf{\eta}$ every $m$ iterations as follows: $\mathbf{\eta} = \btheta^\ell$ if $\mathrm{mod}(\ell,m) = 0$. The update frequency $m$ has been fixed to $100$ in every experiment.

\paragraph{Cyclical SGMCMC.} The cyclical variants of \textbf{SGLD} and \textbf{SGHMC} use a step size of the form \citep{zhang2019cyclical}
\begin{align*}
    \epsilon_k = \frac{\epsilon_0}{2}\left( \cos\left( \displaystyle\frac{\pi\mathrm{mod}(k-1,\lceil K/M \rceil)}{\lceil K/M \rceil} \right) + 1 \right)\,,
\end{align*}
where $\epsilon_0$ is the initial step size, $K$ denotes total number of iterations, and $M$ is the cycle length. The performance of the cyclical variants is studied with respect to the initial step size and the cycle length, as described in section \ref{sec:evaluation_metrics}.

\subsection{Implementation} \label{app:code}
Our experiments are implemented with JAX \citep{jax2018github} and PyTorch \citep{Paszke_PyTorch_An_Imperative_2019}. For the \textbf{LA-FKAC} approximation, we use the \texttt{Laplace} package that implements various types of Laplace approximations \citep{daxberger2021laplace} and relies on PyTorch. The remaining methods are implemented with Google's JAX framework. Regarding the \textbf{SGMCMC} methods, we rely on \texttt{BlackJAX} \citep{blackjax2020github}, where \textbf{SGLD} and \textbf{SGHMC} are already implemented. We use our own implementations of variants with variance reduction, integrated within \texttt{BlackJAX}. The remaining approximation methods are implemented from scratch with JAX.

\subsection{Hyperparameters for numerical experiments}\label{app:hyperparameters}
The selected step sizes for each \textbf{SGMCMC} algorithm are given in Table \ref{app:tab:learning_rates}.
Any other hyperparameter, such as the number of iterations or batch size, is held fixed. For all \textbf{SGMCMC} methods, the number of iterations is set to $10^5$ and the first $50^4$ iterations are discarded as burn-in. The remaining iterations are automatically thinned by selecting $2000$ particles which are obtained by minimizing the maximum mean discrepancy \citep{gretton2006kernel}. This procedure, referred to as MMD thinning, is applied to every \textbf{SGMCMC} output and works as follows. Given $T$ samples $\btheta_1,\dots,\btheta_T$ of a network parameters $\btheta \in \mathbb{R}^d$, we find a sequence of indices $\pi \in \{1,\dots,T\}^m$ such that $\btheta_{\pi(1)},\dots,\btheta_{\pi(m)}$ represent the selected samples. The sequence of indices $\pi$ is obtained thanks to a greedy quantization algorithm \citep{teymur2021optimal} that minimizes the maximum mean discrepancy. Let then $\bP_T$ be the empirical distribution of the $T$ samples $\btheta_1,\dots,\btheta_T$, $\bP_T = \sum_{i=1}^{T} \delta(\btheta_i)$. At the $(i+1)$-th iteration of the quantization algorithm, the index $\pi(i+1)$ is obtained by minimizing the MMD as follows:
\begin{equation*}
\pi(i+1) = \arg\min_{j \in \{1,\dots,T\}} \quad \mathrm{MMD}^2(\bP_T, \bQ_{i+1}(j))\,,
\end{equation*}
where $\bQ_{i+1}(j)$ denotes the empirical measure of the already $i$ selected samples $\btheta_1, \dots, \btheta_{i}$, and an additional sample $\btheta_j$ to be determined:
\begin{equation*}
\bQ_{i+1}(j) =\frac{1}{i+1} \sum_{\ell=1}^{i} \delta(\btheta_{\pi(\ell)}) + \frac{1}{i+1} \delta(\btheta_j)\,.
\end{equation*}
The underlying kernel function $k$ is chosen as the following characteristic distance-based kernel \citep{sejdinovic2013equivalence} $k(\btheta,\btheta^{\prime}) = \|\btheta\|_2 + \|\btheta^{\prime}\|_2 - \|\btheta-\btheta^{\prime}\|_2$. In our all experiments, we select $m = 2000$ samples in order to represent each \textbf{SGMCMC} output.
In this case of the variance reduction technique \textbf{SVRG}, the control variates are updated every $100$ iterations. We use $10$ leapfrog steps in \textbf{SGHMC} and each of its variants. For \textbf{deep ensembles}, $200$ neural networks are independently trained and their weights are initialized from a centered normalized Gaussian distribution. For each of the remaining approximation methods (\textbf{LA-KFAC}, \textbf{MC-Dropout}, and \textbf{SWAG}), samples of sizes $2000$ are saved.

\begin{table}[h]
  \caption{Selected step sizes for each algorithm, equally log-spaced.}
  \label{app:tab:learning_rates}
  \centering
  \begin{tabular}{cccccc}
    \toprule
Step size & SGLD variants  & SGHMC variants & pSGLD, Ensemble, MC Dropout &      LA-KFAC   & SWAG \\
\hline
$\epsilon_1$  & 1e-08  & 1e-08  & 0.0001  & 0.000001  & 1.000000e-07 \\
$\epsilon_2$  & 2.154435e-08  & 1.668101e-08  & 0.000215  & 0.000004  & 1.668101e-07 \\
$\epsilon_3$  & 4.641589e-08  & 2.782559e-08  & 0.000464  & 0.000013  & 2.782559e-07 \\
$\epsilon_4$  & 1.000000e-07  & 4.641589e-08  & 0.001000  & 0.000046  & 4.641589e-07 \\
$\epsilon_5$  & 2.154435e-07  & 7.742637e-08  & 0.002154  & 0.000167  & 7.742637e-07 \\
$\epsilon_6$  & 4.641589e-07  & 1.291550e-07  & 0.004642  & 0.000599  & 1.291550e-06 \\
$\epsilon_7$  & 1.000000e-06  & 2.154435e-07  &  0.010000 &  0.002154 &  2.154435e-06 \\
$\epsilon_8$  & 2.154435e-06  & 3.593814e-07   & 0.021544  & 0.007743  & 3.593814e-06 \\
$\epsilon_9$  & 4.641589e-06  & 5.994843e-07   & 0.046416  & 0.027826  & 5.994843e-06 \\
$\epsilon_{10}$  &1e-05 &  1e-06  &  0.1 &  0.1  & 1e-05 \\    
    \bottomrule
  \end{tabular}
\end{table}

\FloatBarrier

\section{Description of the training and test datasets} \label{app:datasets_regression_problems}
The synthetic regression problems are described in more details below.

\paragraph{Regression problem AF\#1.}
This first regression test case is a homoscedastic regression of a one-dimensional function. The dataset $\mathcal{D} = \{ X_i, Y_i \}_{i=1}^{N}$ is such that $X_i \sim \mathcal{U}([-3,3])$, and $Y_i = \cos(2X_i) + \sin(X_i) + \epsilon_i$, where $\epsilon_i \sim \mathcal{N}(0,\sigma)$, with $\sigma = 0.2$. The test dataset $\mathcal{D}^{\star}$ has the same distribution.

\paragraph{Regression problem AF\#2.}
This test case is taken from \citep{yao2019quality} which consists in a homoscedastic regression of a one-dimensional function as well. The inputs $X_1, \dots, X_N$ of a training dataset $\mathcal{D}$ are uniformly sampled from $[-4,-1] \cup [1,4]$, while test inputs are uniformly distributed between $[-4,4]$. The observations are defined as $Y_i = 0.1X_i^3 + \epsilon_i$ where $\epsilon_i \sim \mathcal{N}(0,0.25)$. 

\paragraph{Regression problem AF\#3.}
This test case is taken from \citep{yao2019quality} which consists in a homoscedastic regression of a one-dimensional function. The first $80$ inputs $X_1, \dots, X_{80}$ of a training dataset $\mathcal{D}$ are uniformly samples in $[-6,-2] \cup [2,6]$, and the last two samples $X_{81},X_{82}$ are uniformly sampled in $[-2,2]$. In contrast, the inputs of a test dataset $\mathcal{D}^{\star}$ are all uniformly distributed in $[-6,6]$. The output observations are given by $Y_i = -(1+X_i)\sin(1.2X_i) + \epsilon_i$ where $\epsilon \sim \mathcal{N}(0,0.25)$.

\paragraph{Regression problem AF\#4.}
This last regression problem is taken from the work of \citet{izmailov2021bayesian}. The inputs of a training dataset $\mathcal{D} = \{(X_i,Y_i)\}_{i=1}^{N}$ are uniformly distributed in $[-10,-6]\cup[6,10]\cup[14,18]$. The observations $Y_i$ are defined as $Y_i = f(X_i; \btheta_0) + \epsilon_i$, with $\epsilon_i \sim \mathcal{N}(0,0.02)$, and where $f(\cdot; \btheta_0)$ denotes a feed-forward neural network with three hidden layers of sizes $100$ (and thus $20,501$ parameters in total). The weight parameters $\btheta_0$ are sampled once from a centered normalized Gaussian distribution $\mathcal{N}(0,\mathbf{I}_d)$. The inputs of the test dataset are uniformly distributed in $[-12,22]$.

\section{Additional results} \label{app:experiments}
In this section, we gather additional results about the empirical experiments, and a few additional details about the networks architecture. For the first synthetic problem (AF \#1), the architecture is made of two hidden layers with $100$ hidden features, leading to a network with $d =10,401$ parameters. Following \citep{yao2019quality}, we consider two hidden layers with $50$ features for the second and third regression problems (AF \#2 and AF \#3), such that the network is made of $d = 2,651$ parameters. Finally, following \citet{izmailov2021bayesian}, a network made of three hidden layers with $100$ features each is use for the last synthetic problem (AF \#4), leading to a network made of $d = 20,501$ parameters.

\subsection{Best marginal coverage probabilities and regression coefficients} \label{app:best_mcp_q2}

We first gather below in Tables \ref{tab2:best_mcp_q2}-\ref{tab4:best_mcp_q2} the best marginal coverages and best $Q^2$ coefficients obtained for the experiments AF\#2, AF\#3, and AF\#4.

\begin{table*}[h]
    \setlength{\tabcolsep}{1pt}
    \renewcommand{\arraystretch}{1}
    \centering
    \caption{AF\#2: Best marginal coverage probabilities (MCP) and best $Q^2$ coefficients obtained with each algorithm.}
\begin{tabular}{|l|cc|cc||l|cc|cc|}
   \hline\hline 
    Method & Best MCP & $Q^2$ & Best $Q^2$ & MCP & Method & Best MCP & $Q^2$ & Best $Q^2$ & MCP \\ 
   \hline 
SGLD & \textbf{0.94} & 0.98 & 0.989 & 0.696 & CSGHMC(10) & \textbf{0.947} & 0.988 & 0.988 & 0.947 \\
SGLD-CV & 0.809 & -4e8 & 0.954 & 0.497 & CSGHMC(100) & \textbf{0.946} & 0.990 & \textbf{0.99} & 0.946 \\
SGLD-SVRG & \textbf{0.954} & 0.995 & \textbf{0.997} & 0.908 & CSGHMC(1000) & 0.928 & 0.99 & \textbf{0.99} & 0.928 \\
SGHMC & \textbf{0.96} & 0.99 & \textbf{0.99} & 0.960 & pSGLD & 0.962 & 0.961 & \textbf{0.993} & 0.996 \\
SGHMC-CV & 0.825 & -13.05 & 0.944 & 0.583 & Deep ensemble & 0.964 & 0.997 & \textbf{0.997} & 0.736 \\
SGHMC-SVRG & \textbf{0.942} & 0.992 & \textbf{0.995} & 0.867 & SWAG & 0.273 & 0.997 & \textbf{0.997} & 0.273 \\
CSGLD(10) & \textbf{0.949} & 0.978 & 0.989 & 0.655 & MC Drop.(0.1) & 0.890 & 0.994 & \textbf{0.994} & 0.89 \\
CSGLD(100) & 0.92 & 0.976 & 0.99 & 0.652 & MC Drop.(0.2) & 0.917 & 0.989 & 0.989 & 0.917 \\
CSGLD(100) & 0.91 & 0.980 & 0.989 & 0.648 & MC Drop.(0.3) & \textbf{0.959} & 0.979 & 0.984 & 0.909 \\
   \hline\hline
\end{tabular}
    \label{tab2:best_mcp_q2}
\end{table*}

\begin{table*}[h]
    \setlength{\tabcolsep}{1pt}
    \renewcommand{\arraystretch}{1}
    \centering
    \caption{AF\#3: Best marginal coverage probabilities (MCP) and best $Q^2$ coefficients obtained with each algorithm.}
\begin{tabular}{|l|cc|cc||l|cc|cc|}
   \hline\hline 
    Method & Best MCP & $Q^2$ & Best $Q^2$ & MCP & Method & Best MCP & $Q^2$ & Best $Q^2$ & MCP \\ 
   \hline 
SGLD & 0.921 & 0.803 & 0.823 & 0.885 & CSGHMC(10) & 0.834 & 0.825 & 0.867 & 0.781 \\
SGLD-CV & 0.551 & -29.357 & -0.091 & 0.034 & CSGHMC(100) & 0.803 & 0.794 & 0.850 & 0.753 \\
SGLD-SVRG & 0.919 & 0.859 & 0.879 & 0.877 & CSGHMC(1000) & 0.805 & 0.817 & 0.851 & 0.740 \\
SGHMC & 0.837 & 0.788 & 0.847 & 0.801 & pSGLD & 0.919 & 0.851 & 0.879 & 0.907 \\
SGHMC-CV & 0.387 & -28.251 & -0.165 & 0.131 & Deep ensemble & \textbf{0.94} & 0.933 & 0.944 & 0.821 \\
SGHMC-SVRG & 0.736 & -1e9 & 0.655 & 0.614 & SWAG & 0.751 & 0.538 & nan & 0.654 \\
CSGLD(10) & 0.902 & 0.829 & 0.833 & 0.856 & MC Drop.(0.1) & 0.624 & 0.792 & 0.799 & 0.588 \\
CSGLD(100) & 0.893 & 0.739 & 0.803 & 0.840 & MC Drop.(0.2) & 0.554 & 0.662 & 0.662 & 0.554 \\
CSGLD(1000) & 0.900 & 0.818 & 0.859 & 0.851 & MC Drop.(0.3) & 0.415 & 0.436 & 0.437 & 0.397 \\
   \hline\hline
\end{tabular}
    \label{tab3:best_mcp_q2}
\end{table*}

\begin{table*}[h!]
    \setlength{\tabcolsep}{1pt}
    \renewcommand{\arraystretch}{1}
    \centering
    \caption{AF\#4: Best marginal coverage probabilities (MCP) and best $Q^2$ coefficients obtained with each algorithm.}
\begin{tabular}{|l|cc|cc||l|cc|cc|}
   \hline\hline 
    Method & Best MCP & $Q^2$ & Best $Q^2$ & MCP & Method & Best MCP & $Q^2$ & Best $Q^2$ & MCP \\ 
   \hline 
SGLD & 0.646 & -2.136 & 0.785 & 0.188 & CSGHMC(10) & 0.589 & -2.316 & 0.814 & 0.133 \\
SGLD-CV & 0.696 & -57.602 & 0.781 & 0.130 & CSGHMC(100) & 0.836 & 0.446 & 0.780 & 0.123 \\
SGLD-SVRG & 0.529 & 0.488 & 0.869 & 0.390 & CSGHMC(1000) & 0.475 & 0.189 & 0.830 & 0.266 \\
SGHMC & 0.661 & -1.575 & 0.792 & 0.098 & pSGLD & 0.957 & -20858.904 & -324.041 & 1.0 \\
SGHMC-CV & 0.616 & -1284.229 & 0.786 & 0.148 & Deep ensemble & 0.413 & 0.420 & 0.925 & 0.203 \\
SGHMC-SVRG & 0.607 & -0.367 & 0.829 & 0.176 & swag & 0.537 & 0.862 & 0.862 & 0.537 \\
CSGLD(10) & 0.639 & -40.265 & 0.769 & 0.087 & MC Drop.(0.1) & 0.232 & 0.802 & 0.851 & 0.227 \\
CSGLD(100) & 0.654 & -5.929760 & 0.785 & 0.139 & MC Drop.(0.2) & 0.400 & 0.796 & 0.860 & 0.270 \\
CSGLD(1000) & 0.563 & -0.639 & 0.769 & 0.126 & MC Drop.(0.3) & 0.453 & 0.549 & 0.821 & 0.314 \\
   \hline\hline
\end{tabular}
    \label{tab4:best_mcp_q2}
\end{table*}

\subsection{Similarities between the algorithms} \label{app:similarities_afi}

\begin{figure}[h!]
\centering
\begin{subfigure}[t]{0.43\textwidth}
\centering
\includegraphics[width=\textwidth]{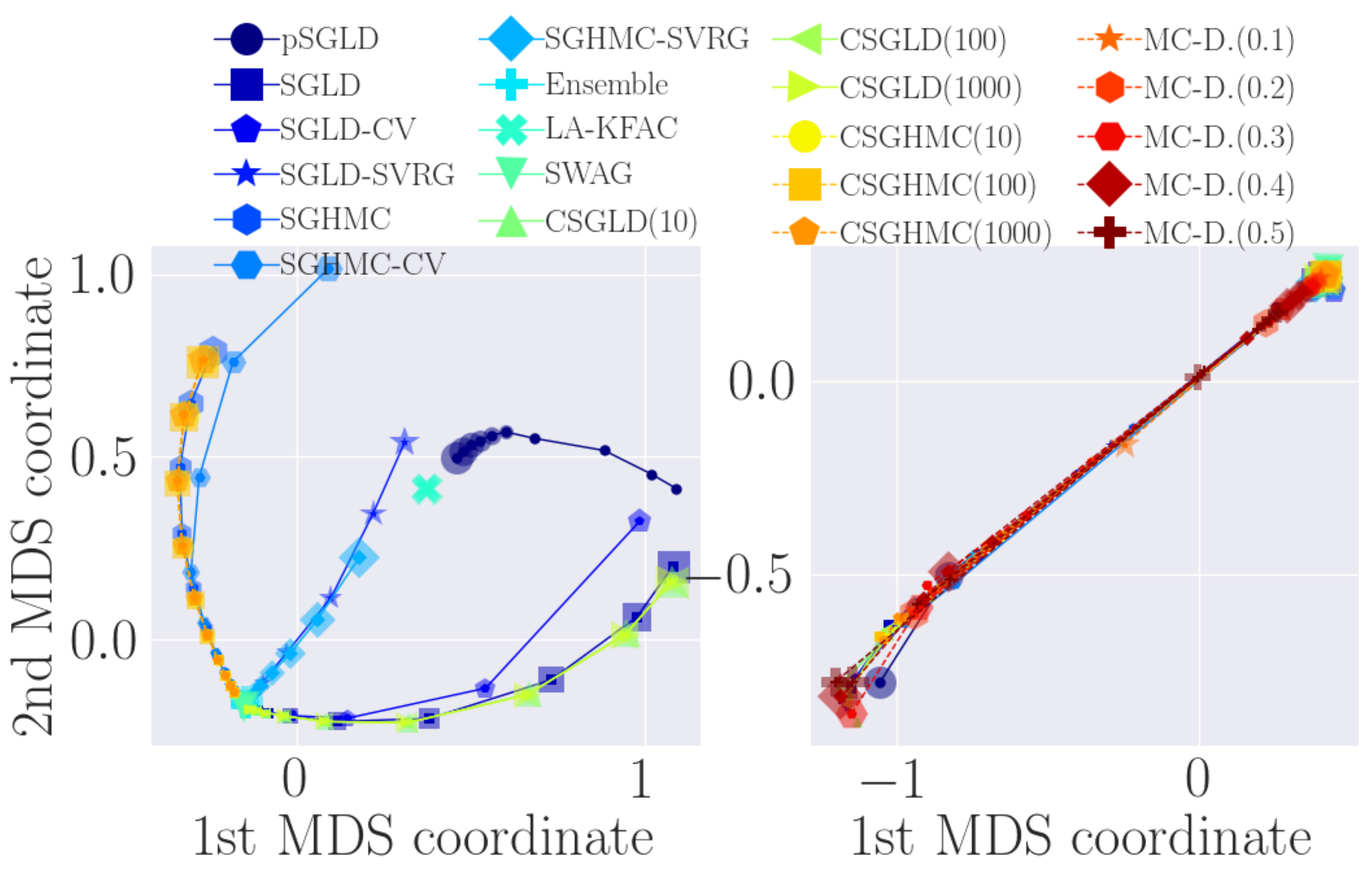}
\caption{Problem AF\#1}
\end{subfigure}
\begin{subfigure}[t]{0.43\textwidth}
\centering
\includegraphics[width=\textwidth]{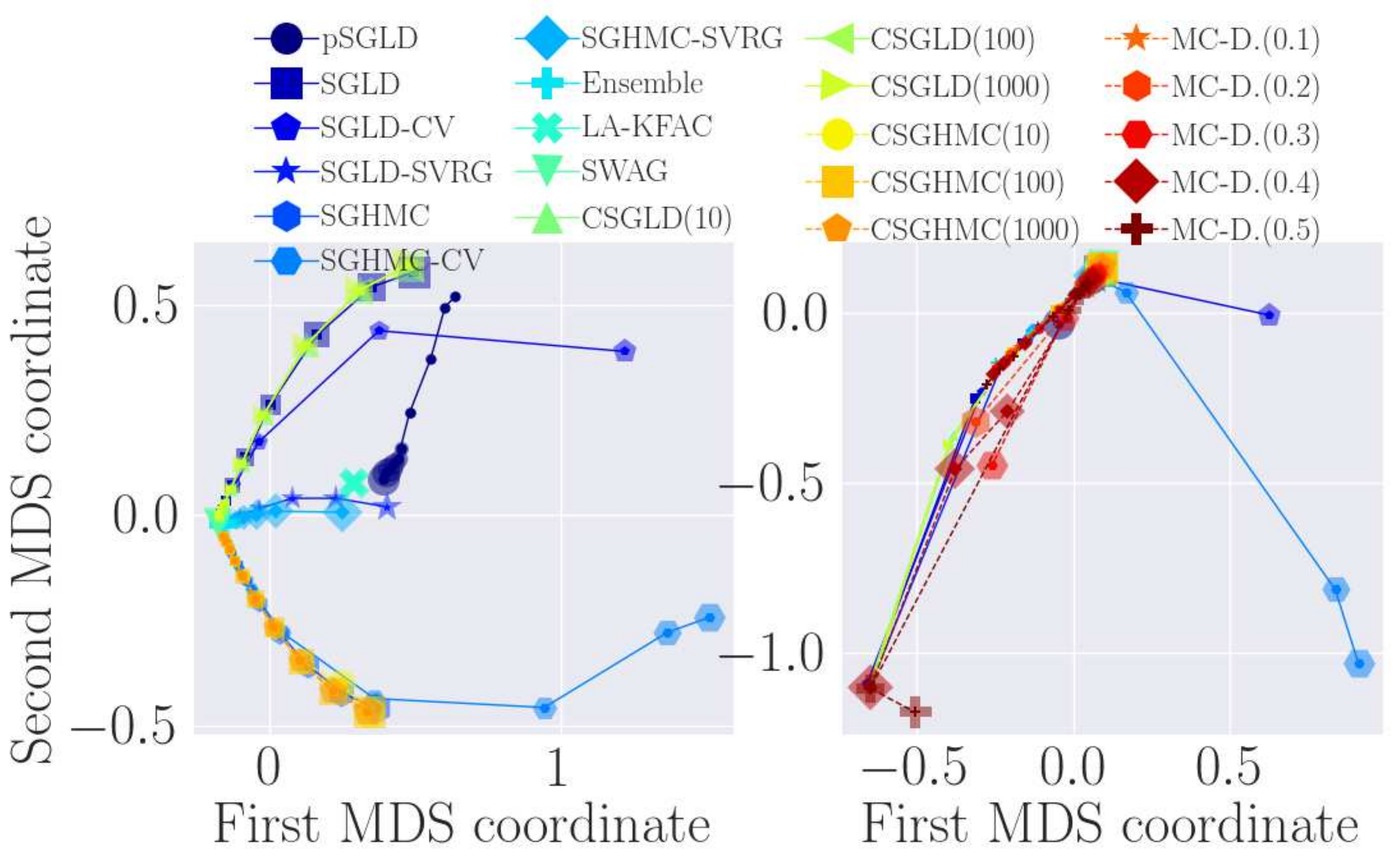}
\caption{Problem AF\#2}
\end{subfigure}

\begin{subfigure}[t]{0.43\textwidth}
\centering
\includegraphics[width=\textwidth]{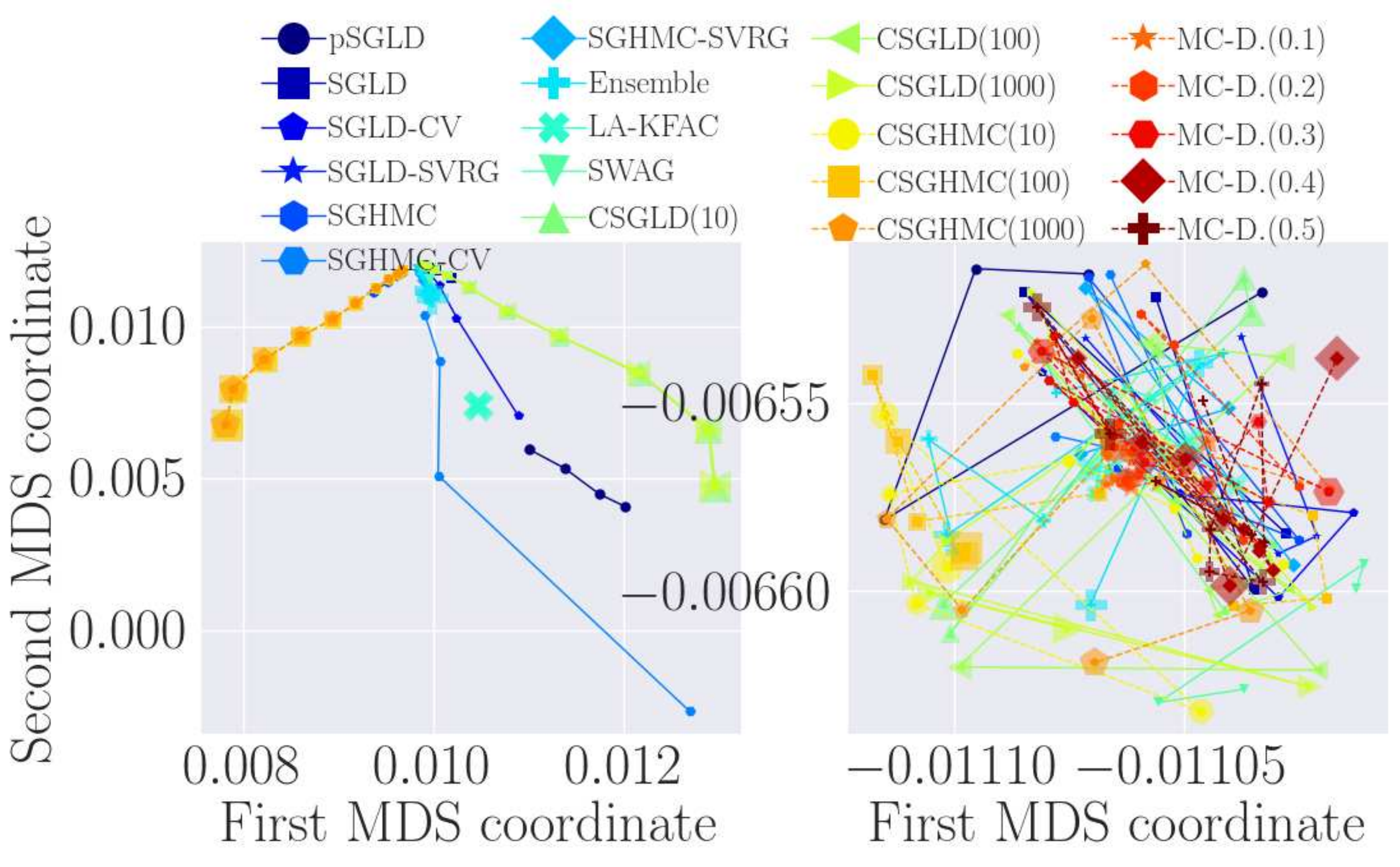}
\caption{Problem AF\#3}
\end{subfigure}
\begin{subfigure}[t]{0.43\textwidth}
\centering
\includegraphics[width=\textwidth]{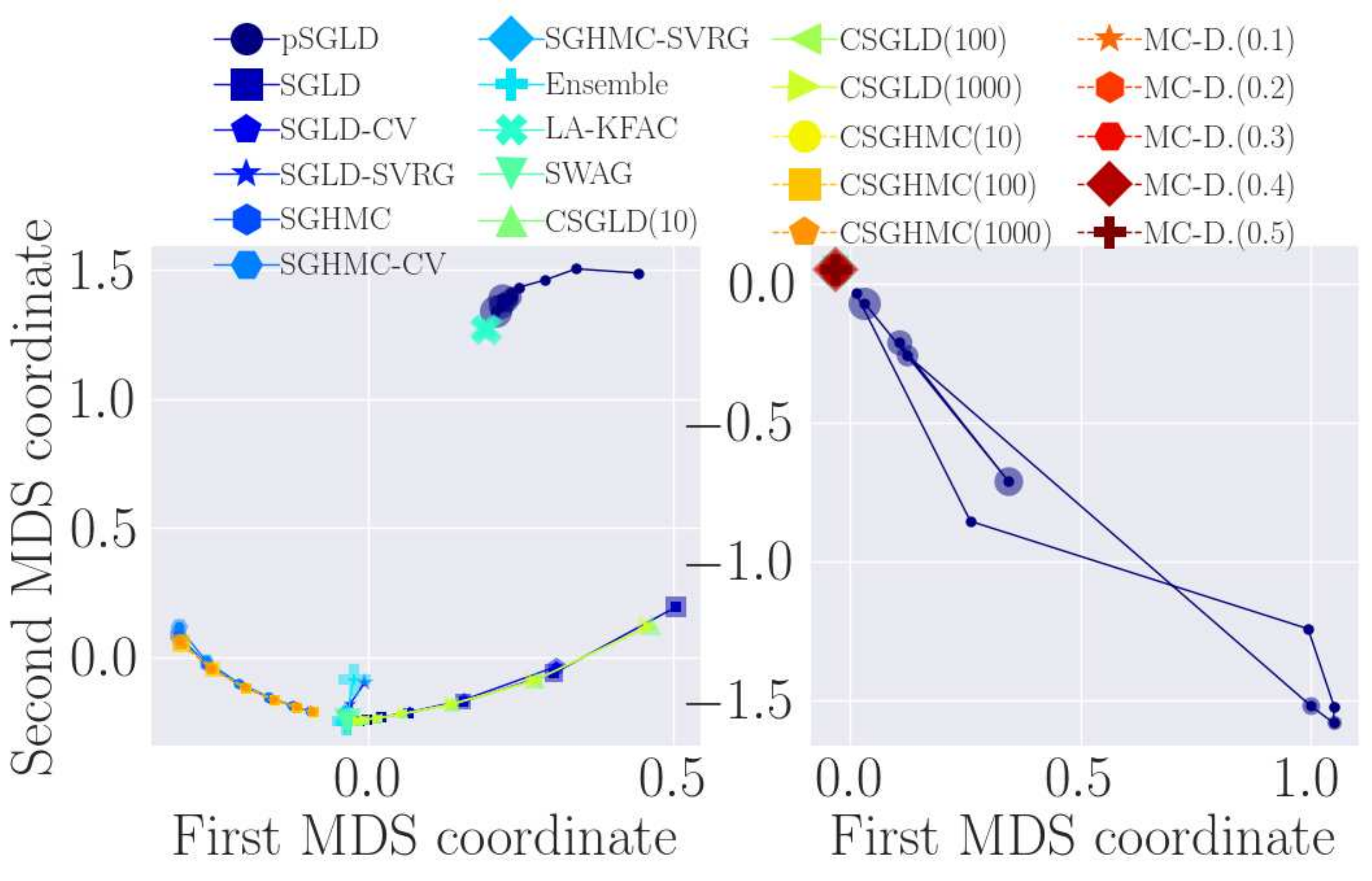}
\caption{Problem AF\#4}
\end{subfigure}
\caption{Similarities between the algorithms as measured by the MMD and represented in a two-dimensional space build with multidimensional scaling. Left panels: weight space, right panels: function space. The markers sizes are proportional to the value of the underlying step size $\epsilon$.} 
\label{app:fig:similarities_afi}
\end{figure}

The similarities between the algorithms are gathered for all the experiments in Figure \ref{app:fig:similarities_afi}. The similarities in weight space seem to exhibit the same structure in our four experiments. However, the similarities in function space do not exhibit any particular structure.

\subsection{Additional results for regression problem AF\#1}

\begin{figure}[h]
\centering
\includegraphics[width=0.42\textwidth]{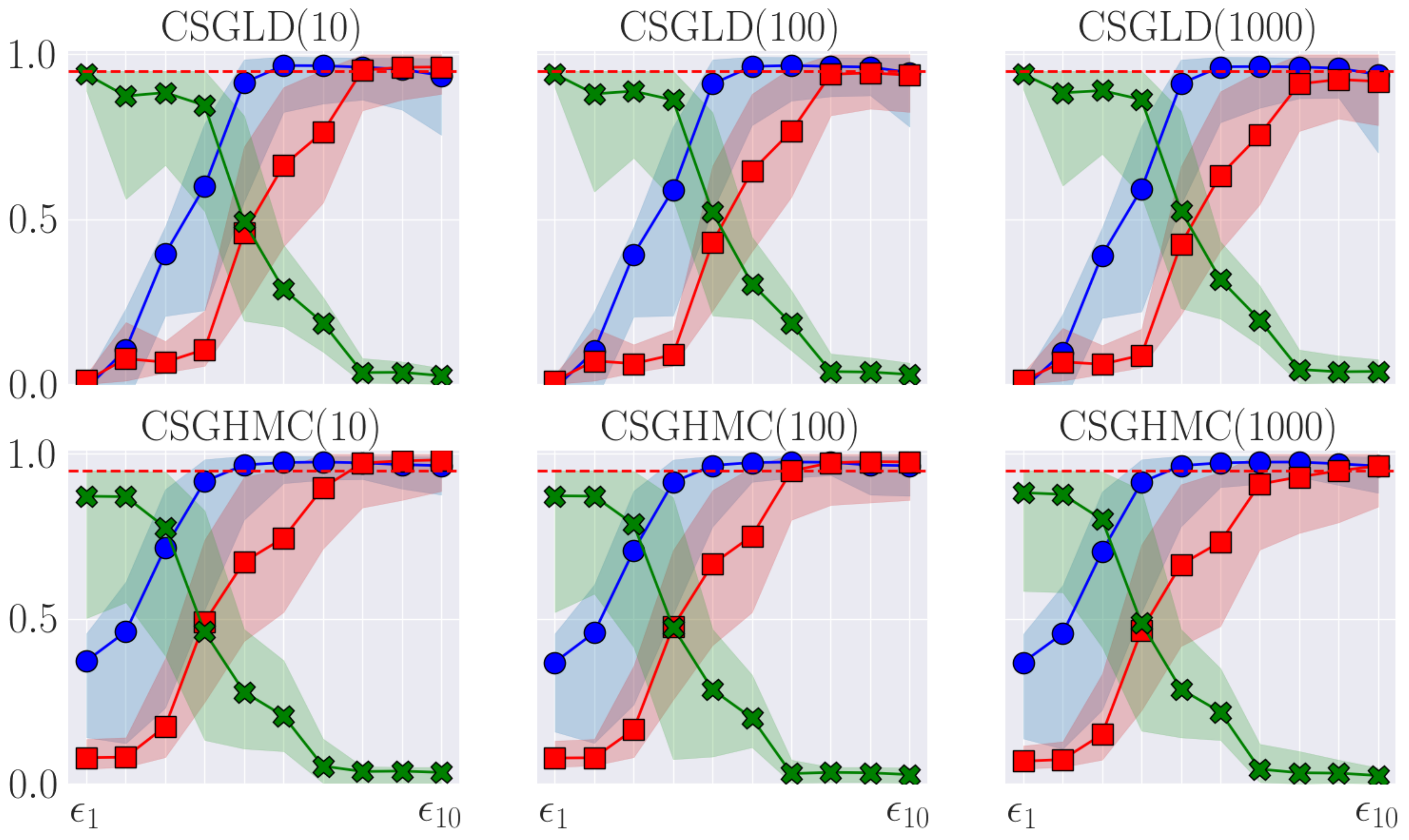}
\caption{Problem AF\#1. Coverage metrics and $Q^2$ coefficient with respect to the step size $\epsilon$. The target coverage is set to $0.95$.}
\label{fig:q2_coverage_af1_csgmcmc.pdf}
\end{figure}

Results obtained for the cyclical variants of \textbf{SGMCMC} have been reported herein. Figure \ref{fig:q2_coverage_af1_csgmcmc.pdf} shows the $Q^2$ regression cofficient and coverage probabilities obtained with \textbf{CSGMCMC}.

\begin{figure}[h]
\centering
\includegraphics[width=0.49\textwidth]{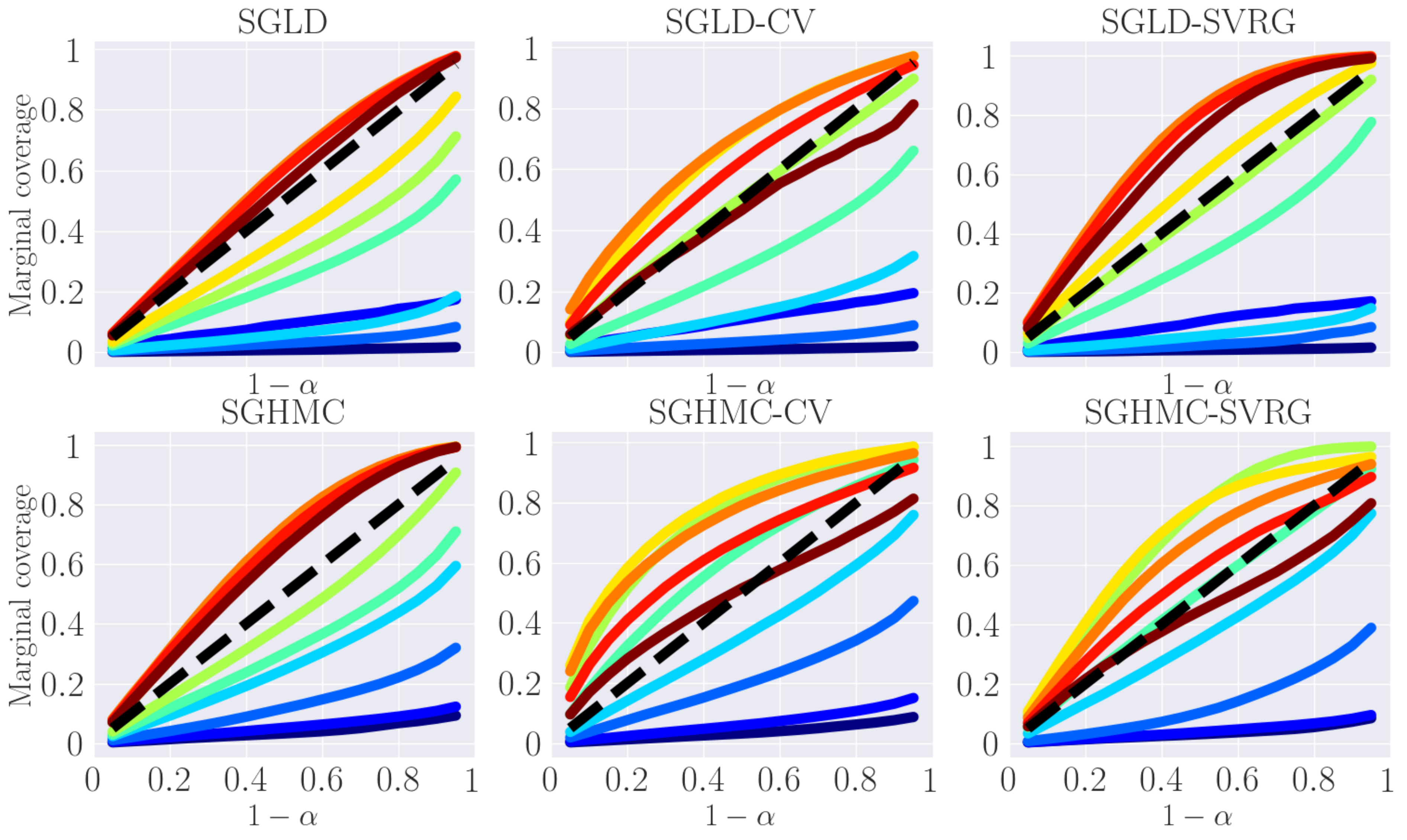}
\includegraphics[width=0.49\textwidth]{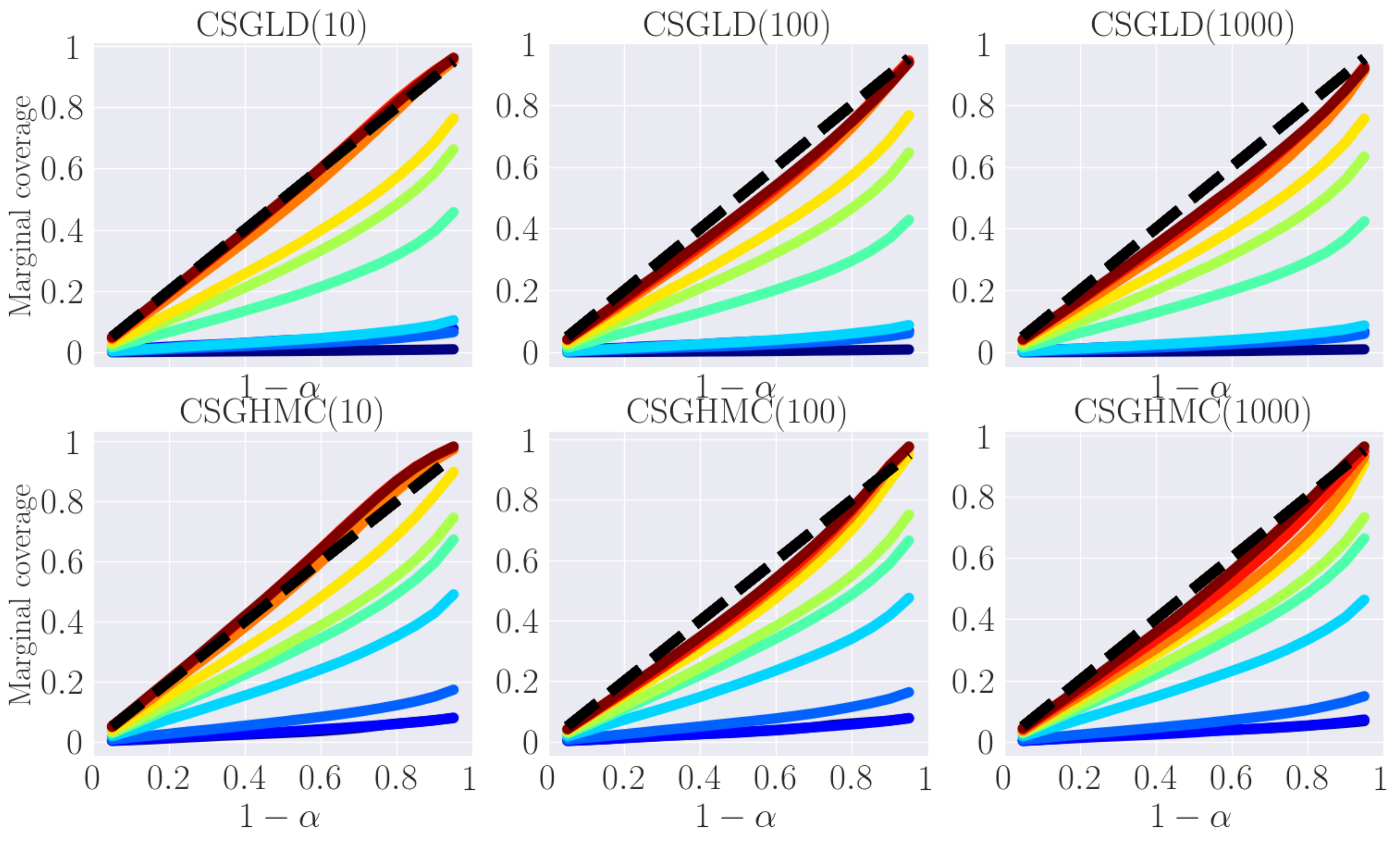}
\caption{Problem AF\#1. Graphs of the marginal coverage probability with respect to the target level for \textbf{SGMCMC} methods. The curves are colored by the value of the underlying step size: dark blue corresponds to the lowest, while dark red corresponds to the highest step size.}
\label{fig:mcp_graphs_sgmcmc_af1}
\end{figure}

\begin{figure}[h]
\centering
\includegraphics[width=\textwidth]{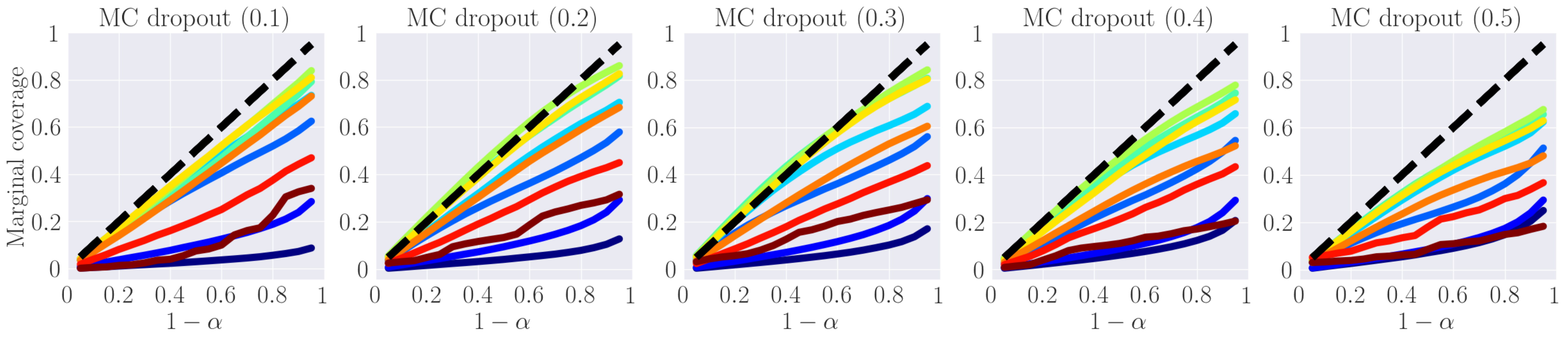}

\includegraphics[width=0.62\textwidth]{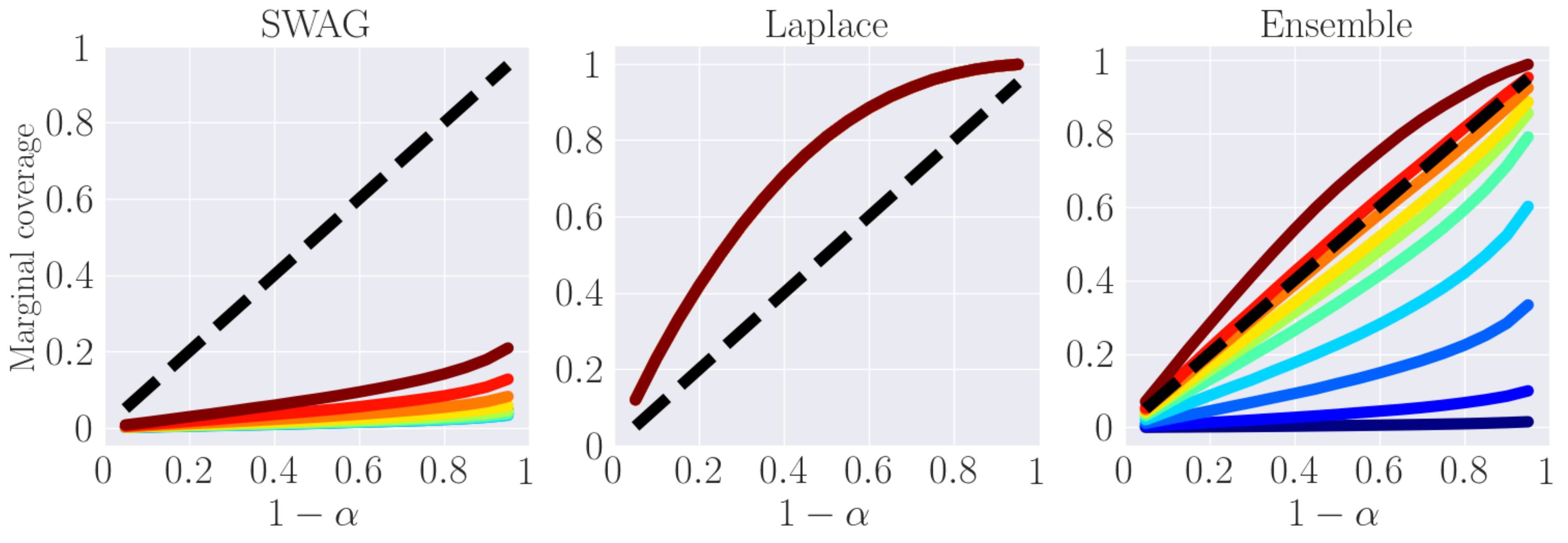}
\includegraphics[width=0.21\textwidth]{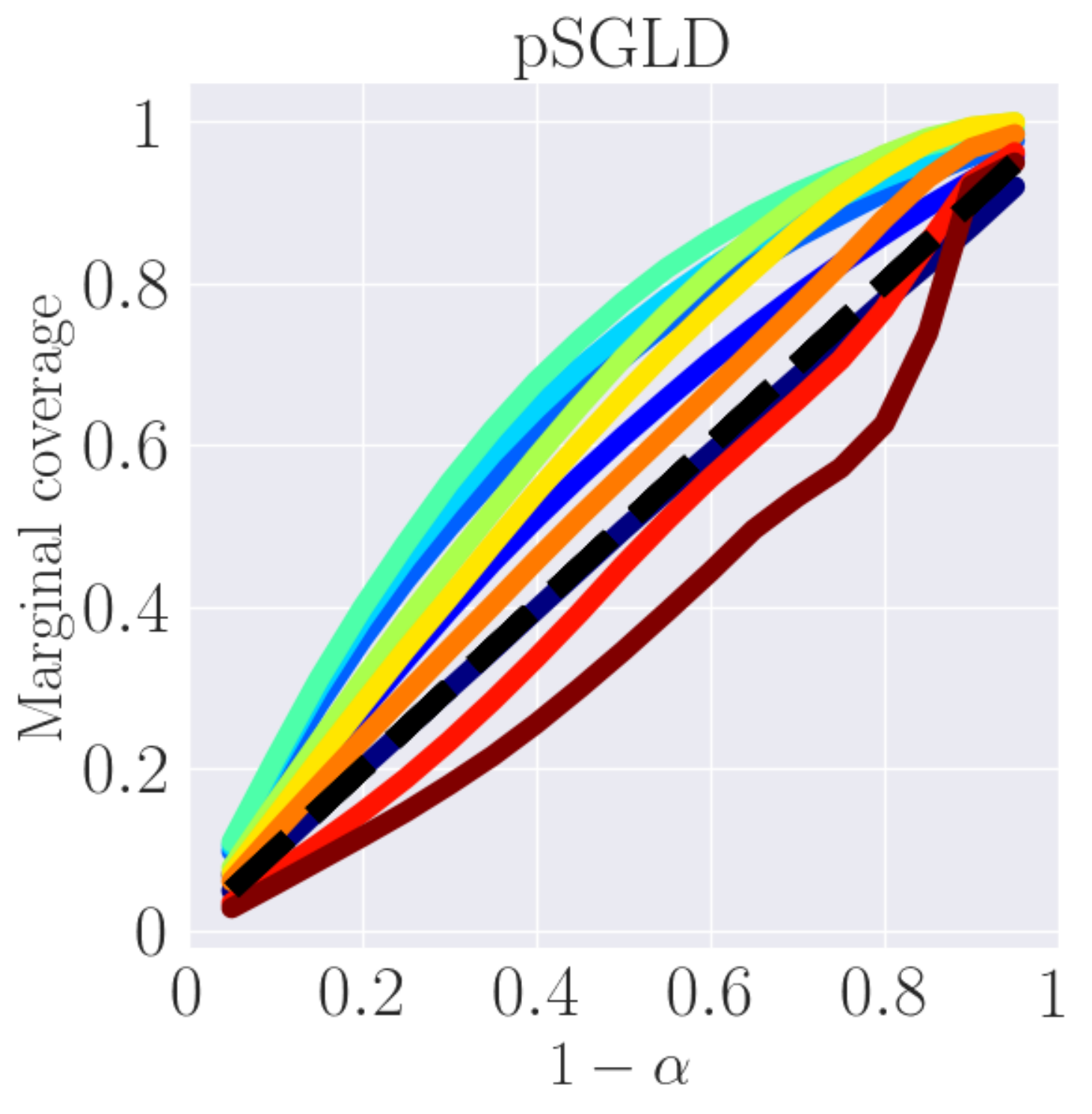}
\caption{Problem AF\#1. Graphs of the marginal coverage probability with respect to the target level for \textbf{MC-Dropout}, \textbf{SWAG}, \textbf{LA-KFAC}, \textbf{deep ensembles}, and \textbf{pSGLD}.}
\label{fig:mcp_graphs_mcdropout_af1}
\end{figure}

\begin{figure}
\centering
\includegraphics[width=0.60\textwidth]{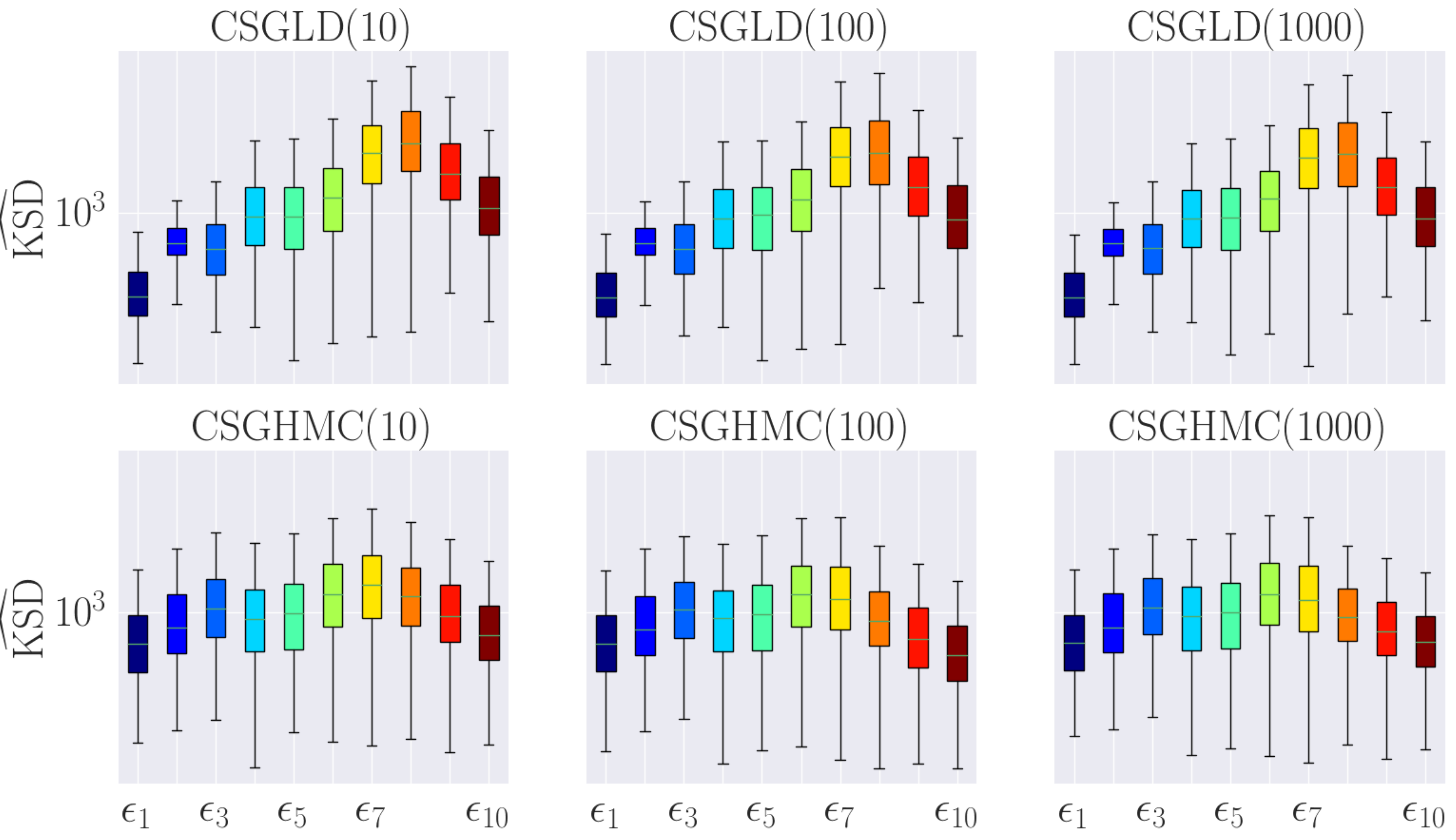}
\caption{Problem AF\#1. Kernelized Stein discrepancies $\mathrm{KSD}(\bP,\bQ)$ between the target posterior measure $\bP$ and the approximated posteriors $\bQ$.}
\label{fig:ksd_af1_2}
\end{figure}

We also report here graphs of the marginal coverage probability with respect to the target confidence level $(1-\alpha)$, for some $\alpha \in [0.05, 0.95]$, and all the considered values for the underlying hyperparameters, which are gathered in Figures \ref{fig:mcp_graphs_sgmcmc_af1}-\ref{fig:q2_coverage_af2_csgmcmc.pdf}. 

We find that most \textbf{SGMCMC} methods (without variance reduction) give rise to marginal coverage probabilities that do not increase monotonically with the target level, and that are easily much higher than the target level. The cyclical step size reduces the marginal coverage probabilities. The \textbf{pSGLD} methods yield very different results as the target level is easily achieved, even for the lowest step sizes. \textbf{LA-FKAC} shows a similar behavior, where the marginal coverage probability is always above the target level.

The \textbf{MC-Dropout} method is able to reach the target level whenever $\alpha$ is small enough, but struggles to reach an appropriate marginal coverage as $\alpha$ increases. Finally, we find that the marginal coverage obtained with \textbf{deep ensembles} varies monotonically with the target level for sufficiently high step sizes. 

\subsection{Additional results for regression problem AF\#2}
\begin{figure}[h]
\centering
\includegraphics[width=0.5\textwidth]{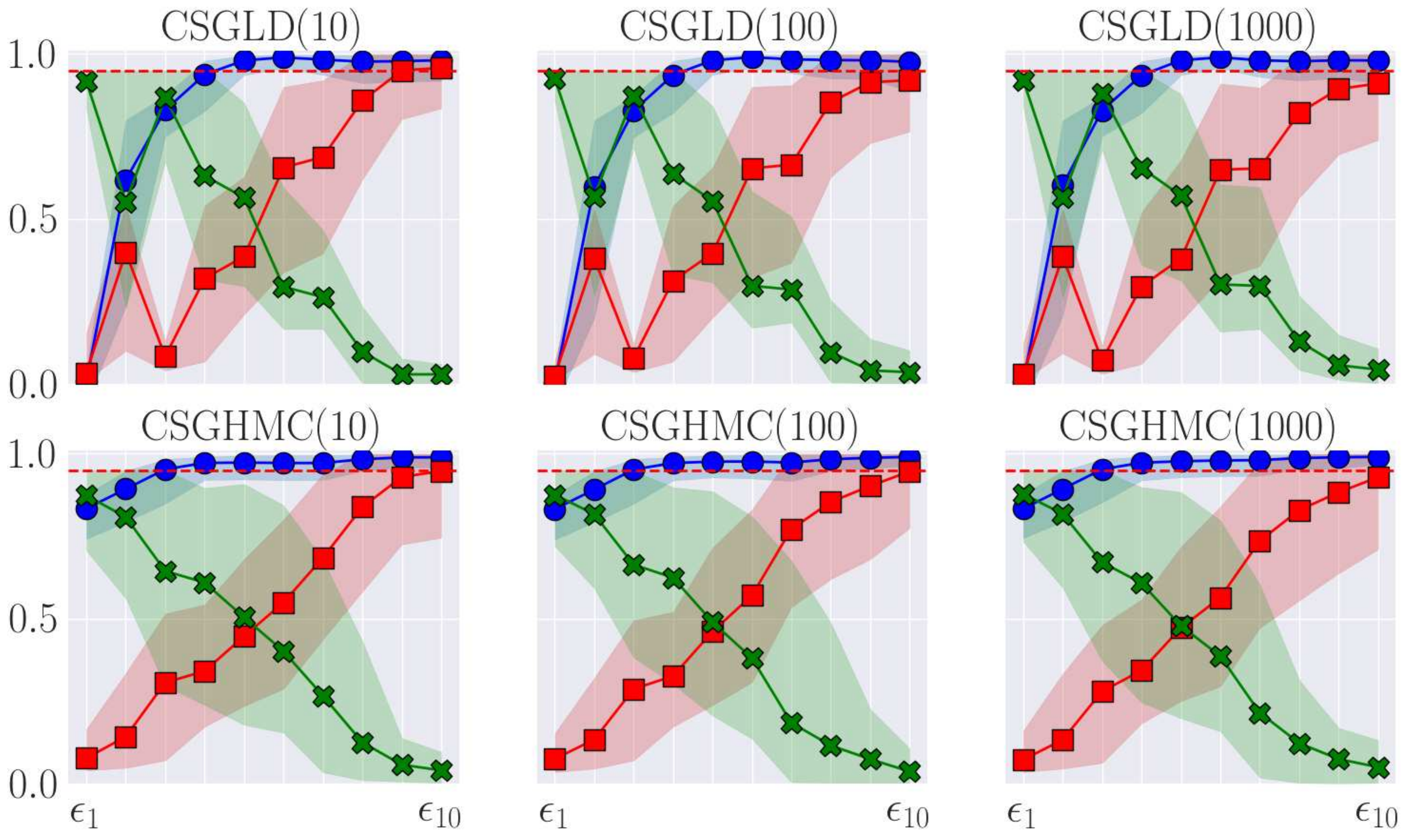}
\caption{Problem AF\#2. Coverage metrics and $Q^2$ coefficient with respect to the step size $\epsilon$. The target coverage is set to $0.95$.}
\label{fig:q2_coverage_af2_csgmcmc.pdf}
\end{figure}

\begin{figure}
\centering
\includegraphics[width=0.44\textwidth]{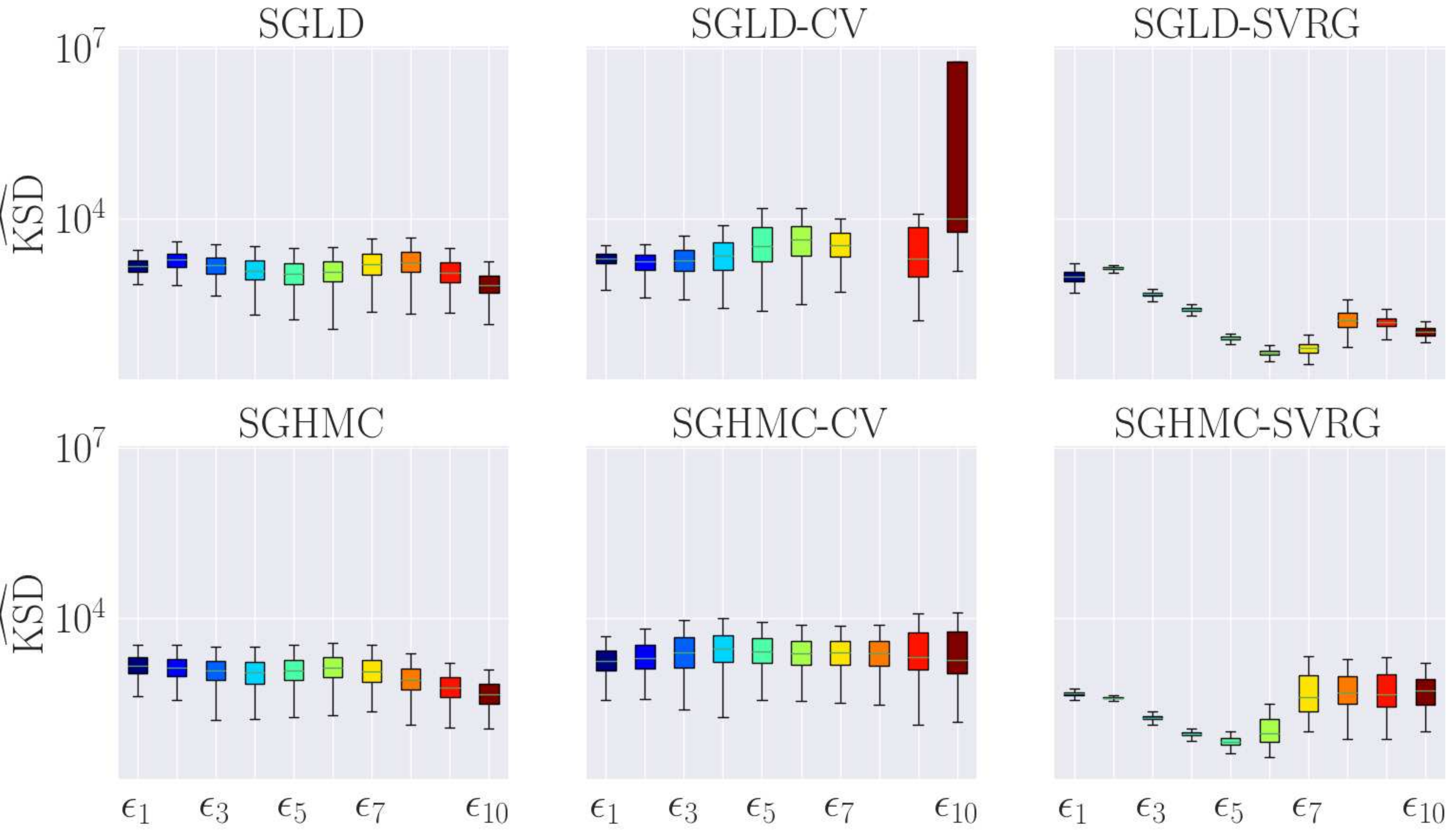}
\includegraphics[width=0.44\textwidth]{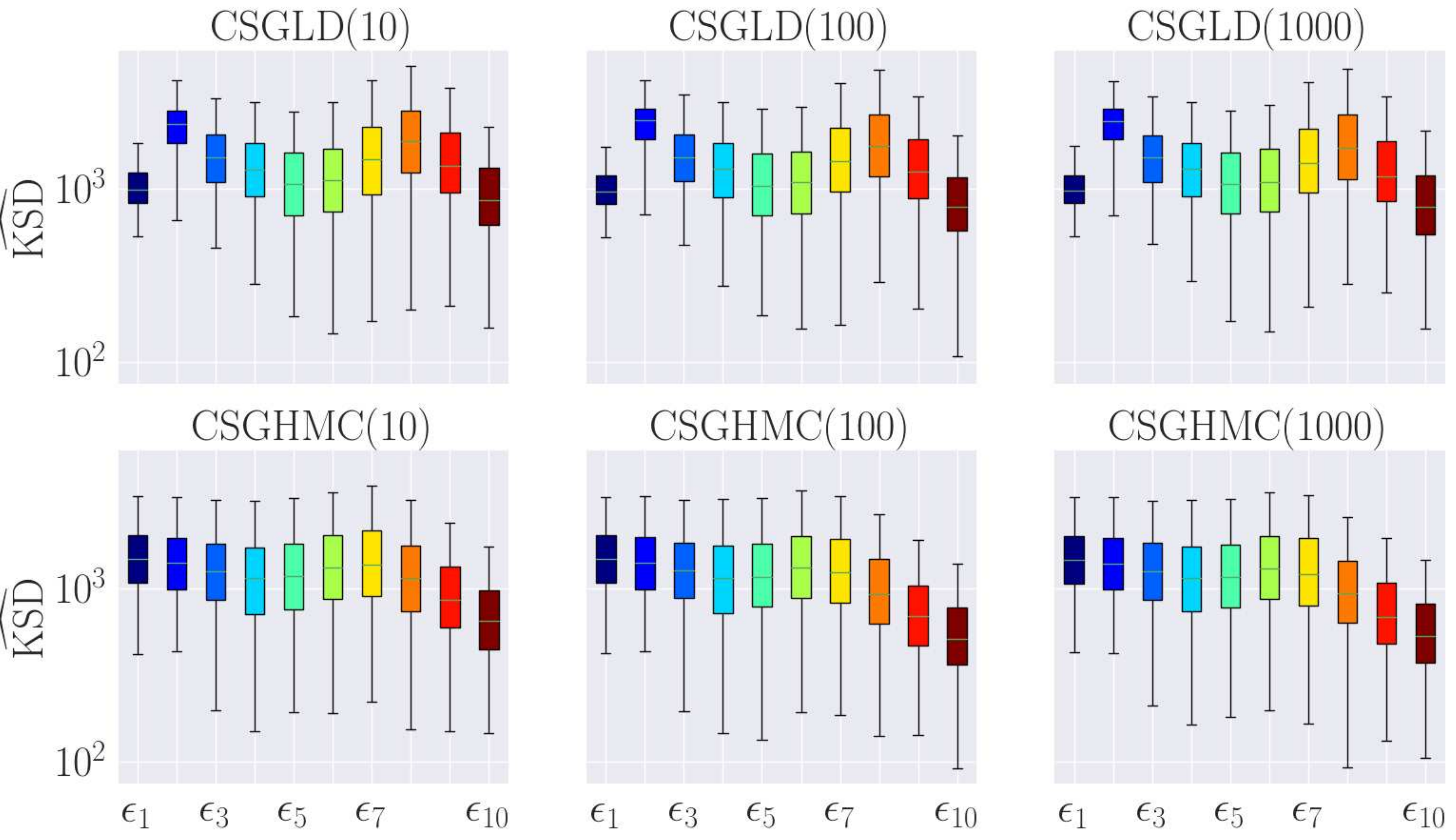}
\includegraphics[width=0.44\textwidth]{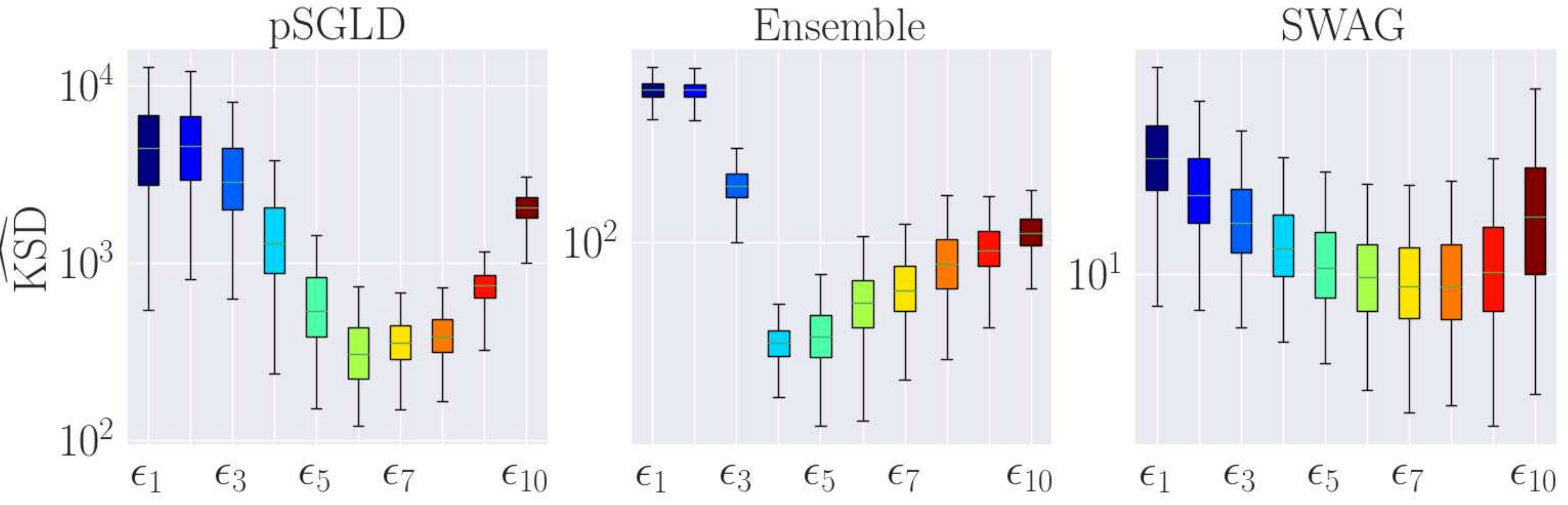}
\caption{Problem AF\#2. Kernelized Stein discrepancies $\mathrm{KSD}(\bP,\bQ)$ between the target posterior measure $\bP$ and the approximated posteriors $\bQ$.}
\label{fig:ksd_af2}
\end{figure}

\begin{figure}[h]
\centering
\includegraphics[width=0.38\textwidth]{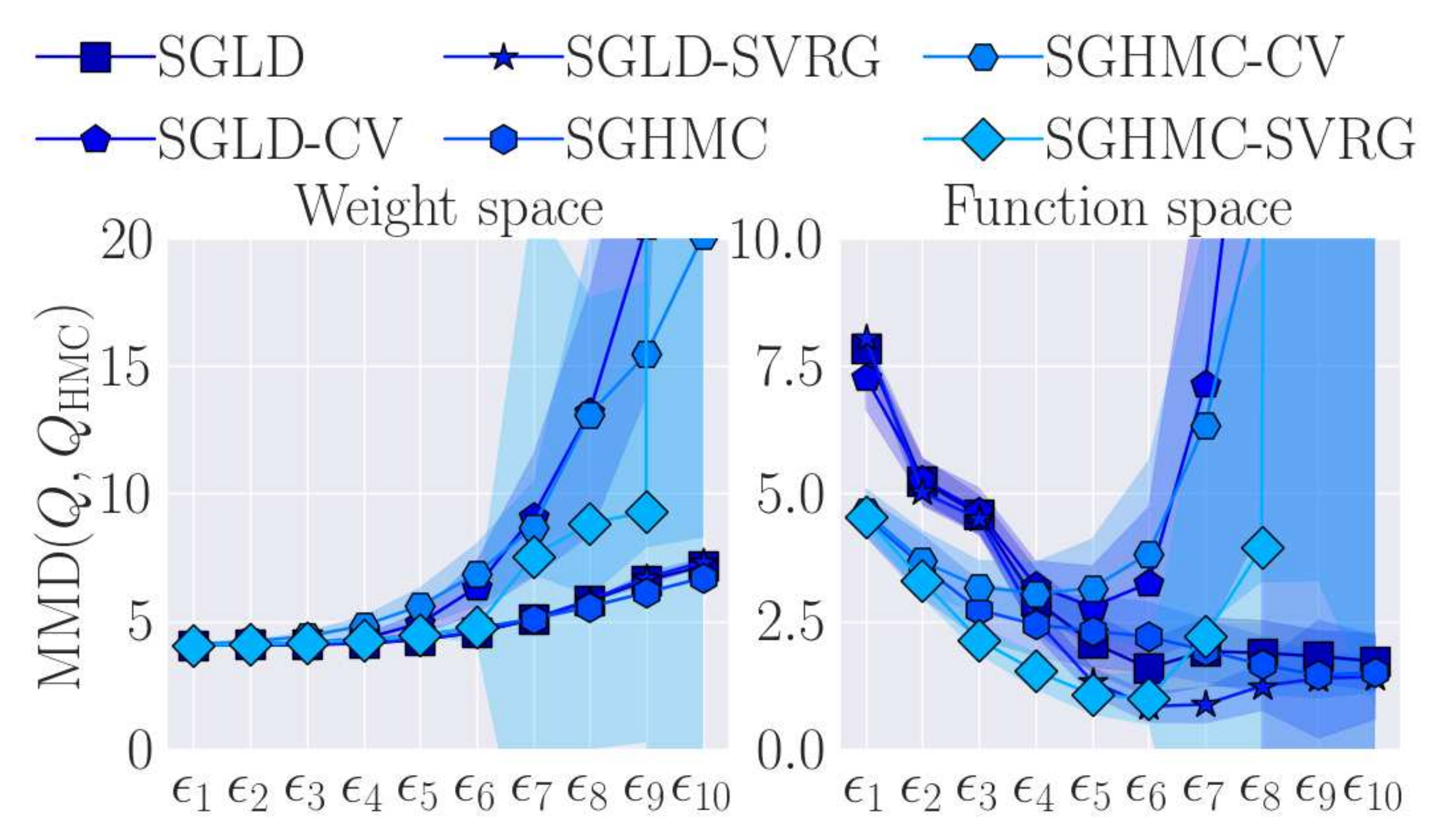}
\includegraphics[width=0.38\textwidth]{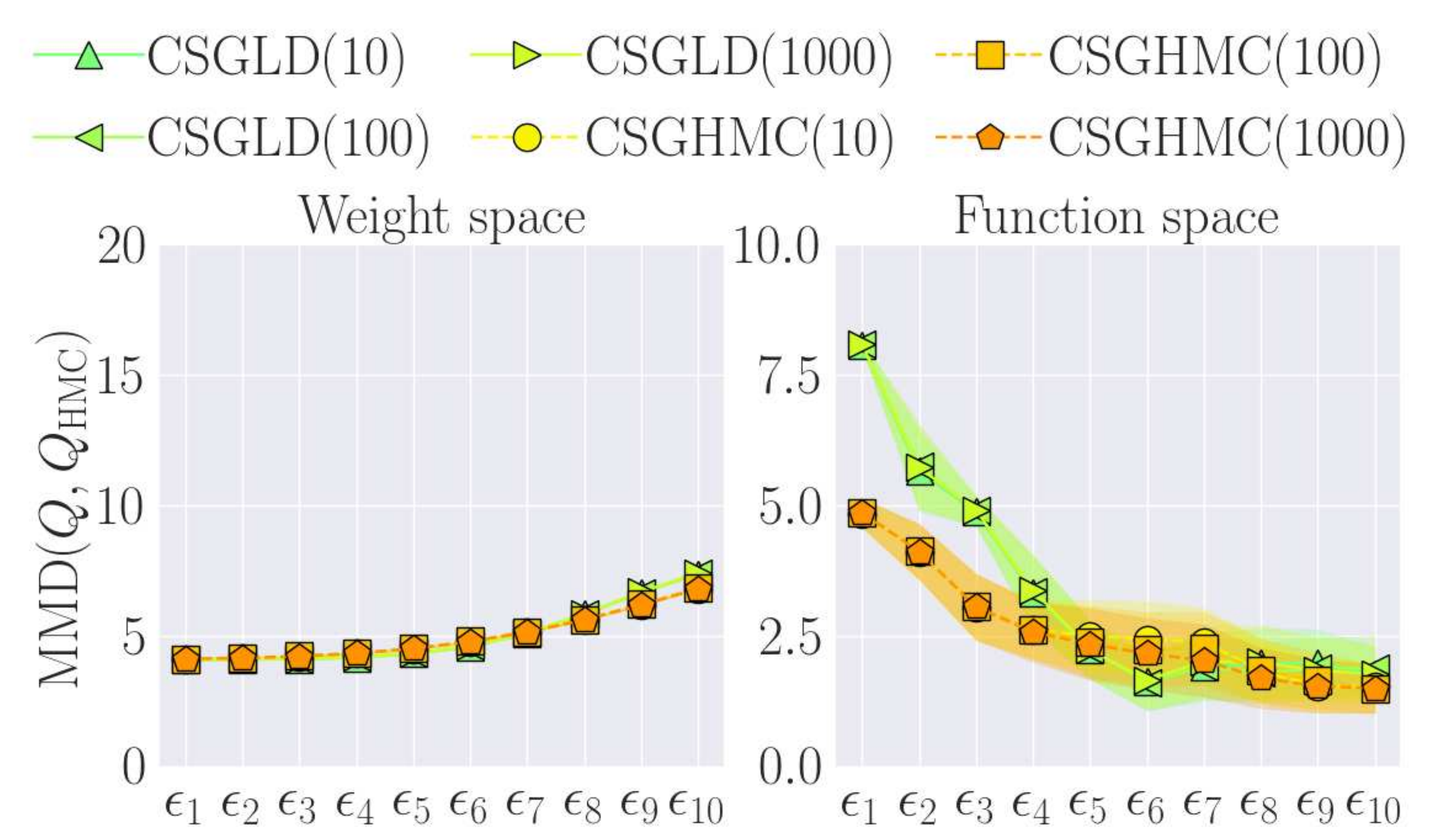}
\includegraphics[width=0.38\textwidth]{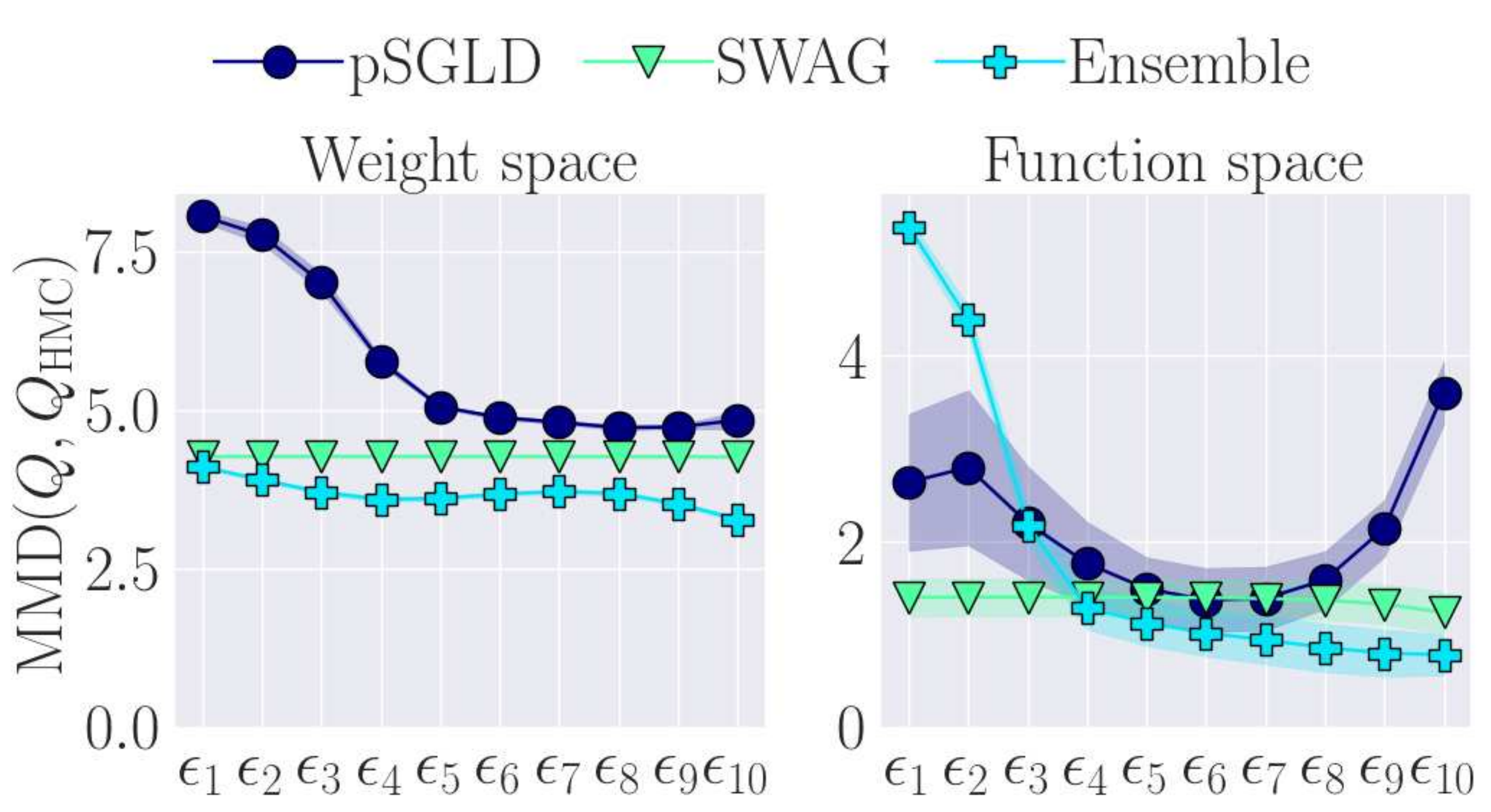}
\caption{Problem AF\#2. Maximum mean discrepancies $\mathrm{MMD}(\bQ,\bQ_{\mathrm{HMC}})$ between the approximated posterior distributions $\bQ$ and the reference HMC sample, in weight and function spaces.}
\label{fig:mmd_weight_function_spaces_af2}
\end{figure}

\begin{figure}[h]
\centering
\includegraphics[width=0.42\textwidth]{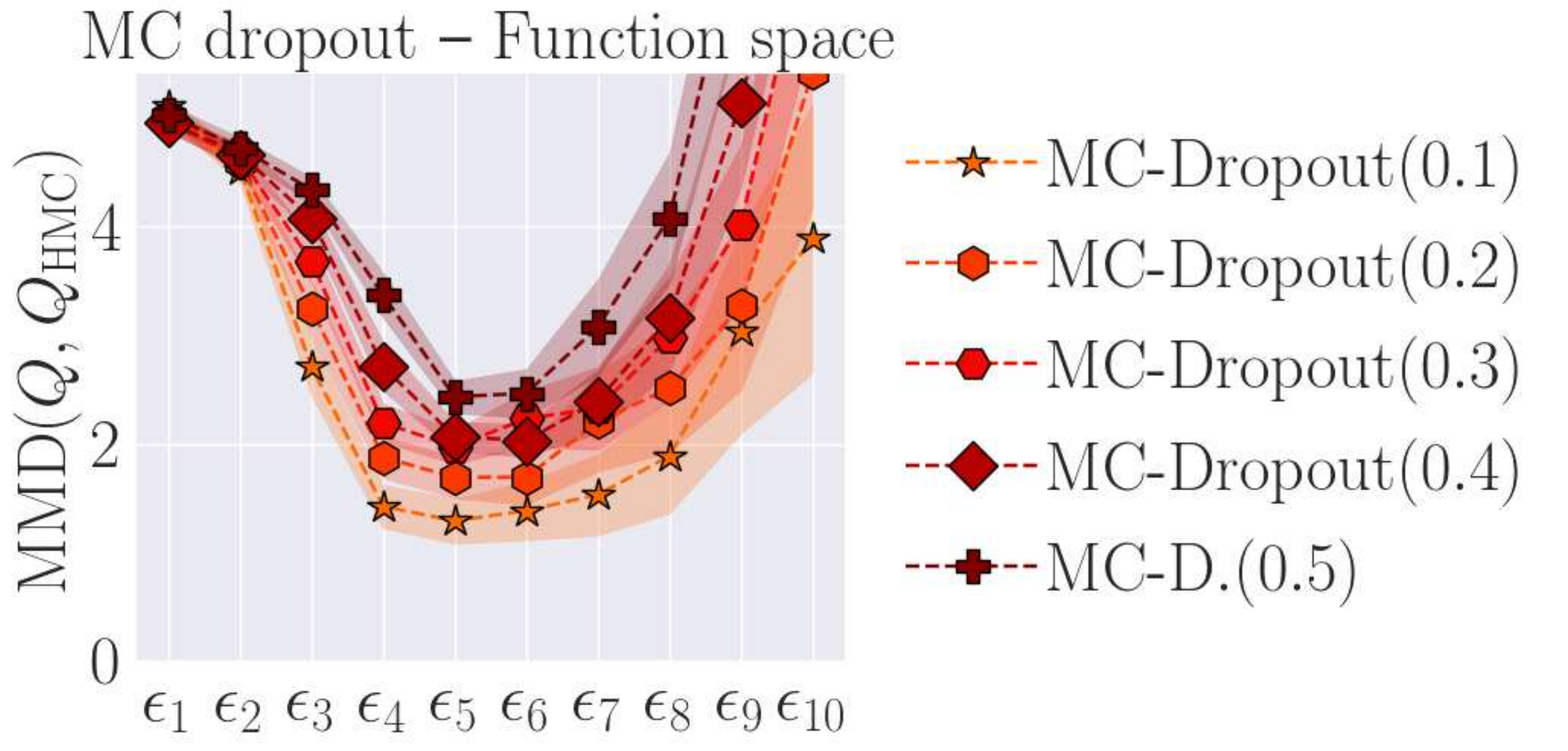}
\caption{Problem AF\#2. Maximum mean discrepancies between the approximated posterior distributions and the reference HMC sample, in weight and function spaces.}
\label{fig:mmd_weight_function_spaces_mcdropout_af2}
\end{figure}

The coverage metrics and regression coefficient obtained with \textbf{CSGLD} and \textbf{CSGHMC} are shown in Figure \ref{fig:q2_coverage_af2_csgmcmc.pdf}. The performances of the cyclical variants are similar to those of \textbf{SGLD} and \textbf{SGHMC} (see Figure \ref{fig:q2_coverage_af2_sgmcmc.pdf}). The graphs of the kernelized Stein discrepancies and maximum mean discrepancies are shown in Figures \ref{fig:ksd_af2} and \ref{fig:mmd_weight_function_spaces_af2}-\ref{fig:mmd_weight_function_spaces_mcdropout_af2}.

\begin{figure}[h]
\centering
\includegraphics[width=0.49\textwidth]{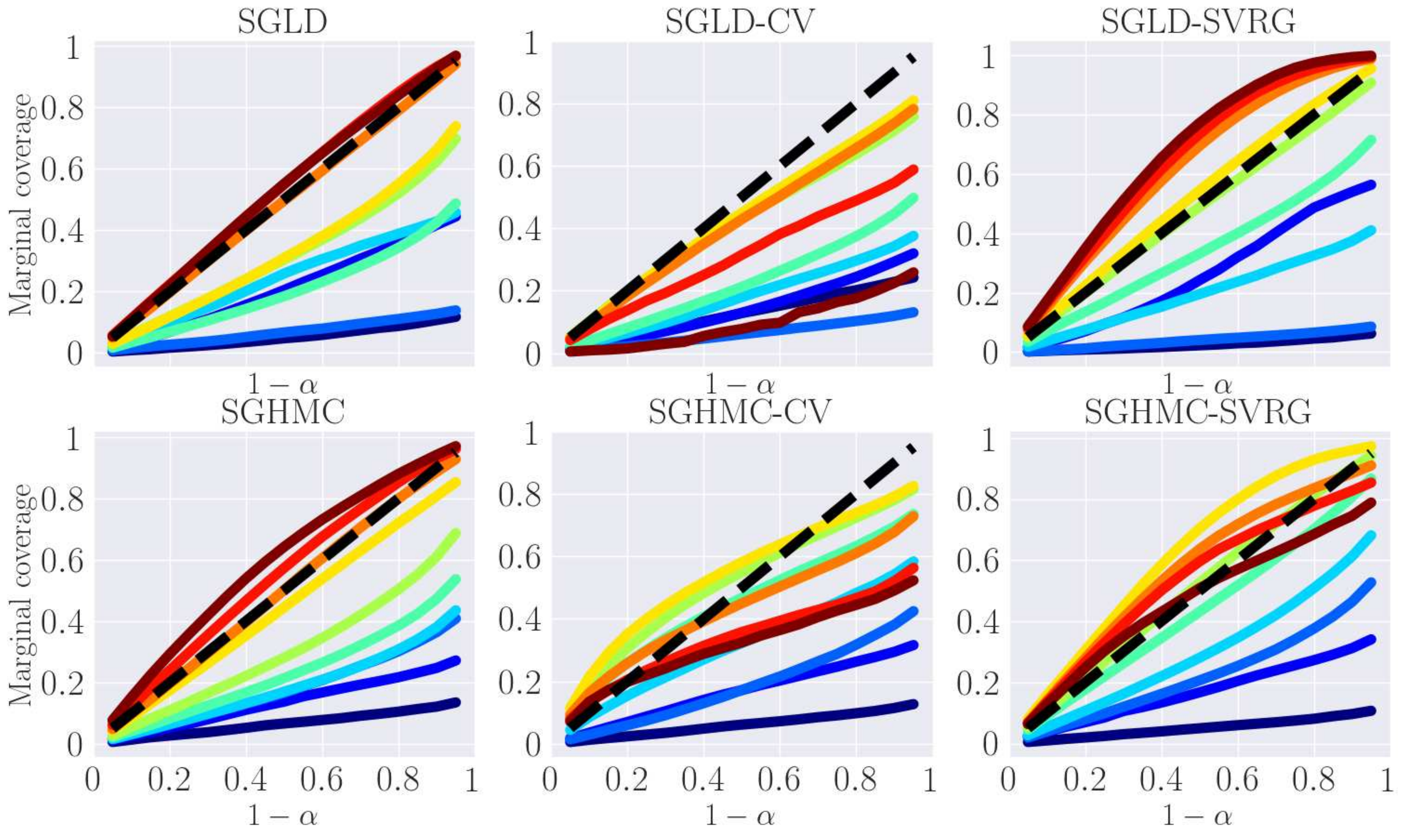}
\includegraphics[width=0.49\textwidth]{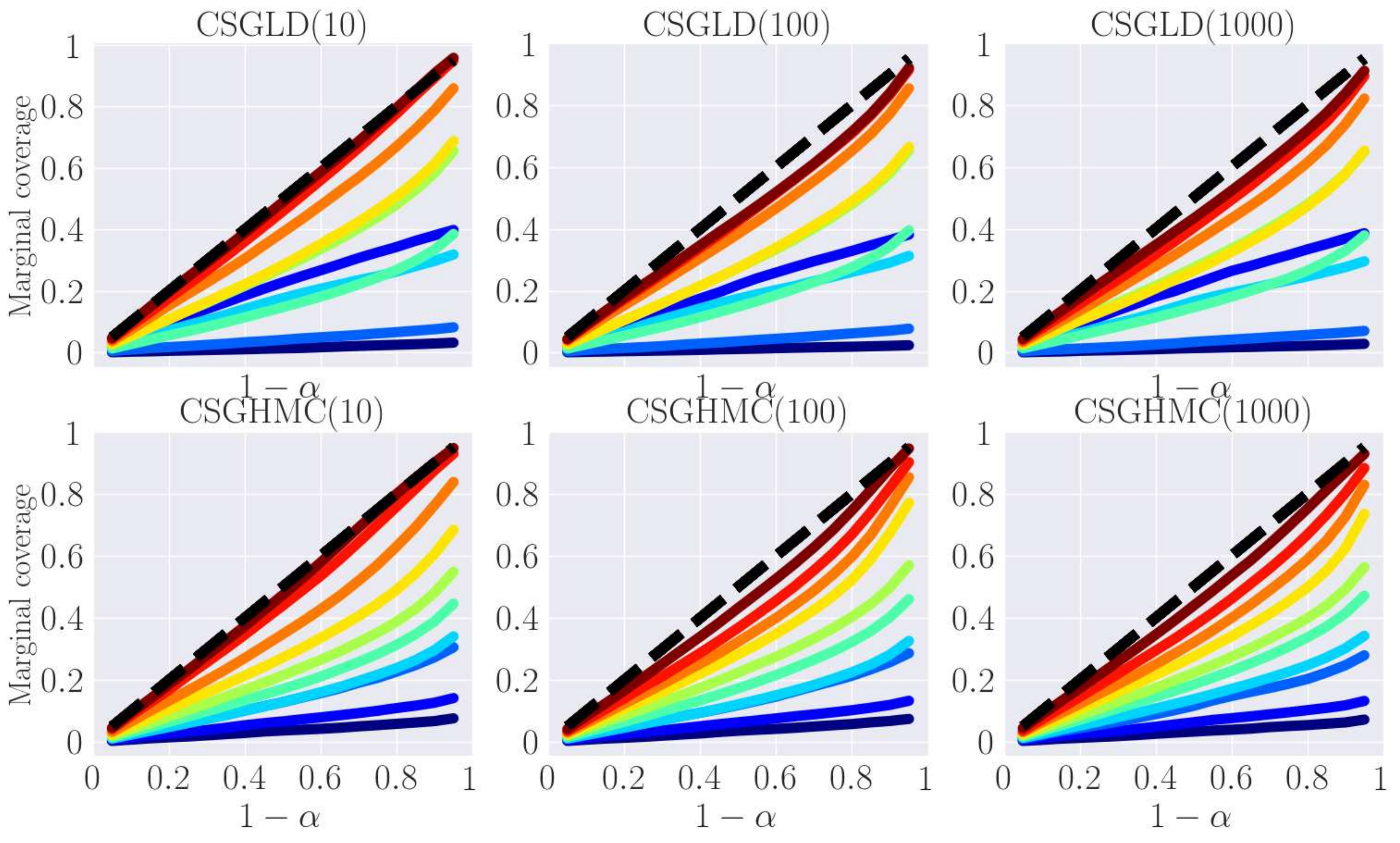}
\caption{Problem AF\#2. Graphs of the marginal coverage probability with respect to the target level for \textbf{SGMCMC} methods. The curves are colored by the value of the underlying step size: dark blue corresponds to the lowest, while dark red corresponds to the highest step size.}
\label{fig:mcp_graphs_sgmcmc_af2}
\end{figure}

\begin{figure}[h]
\centering
\includegraphics[width=\textwidth]{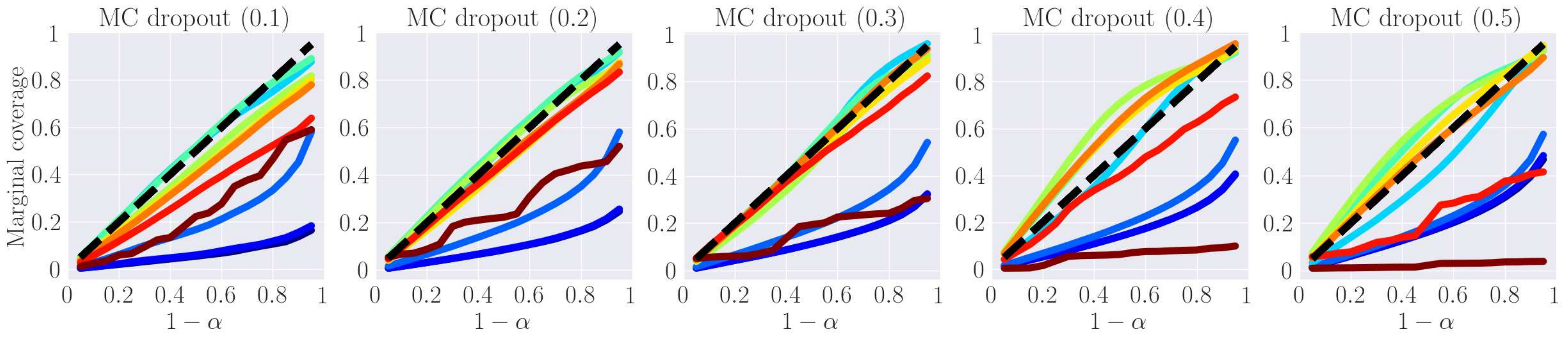}

\includegraphics[width=0.62\textwidth]{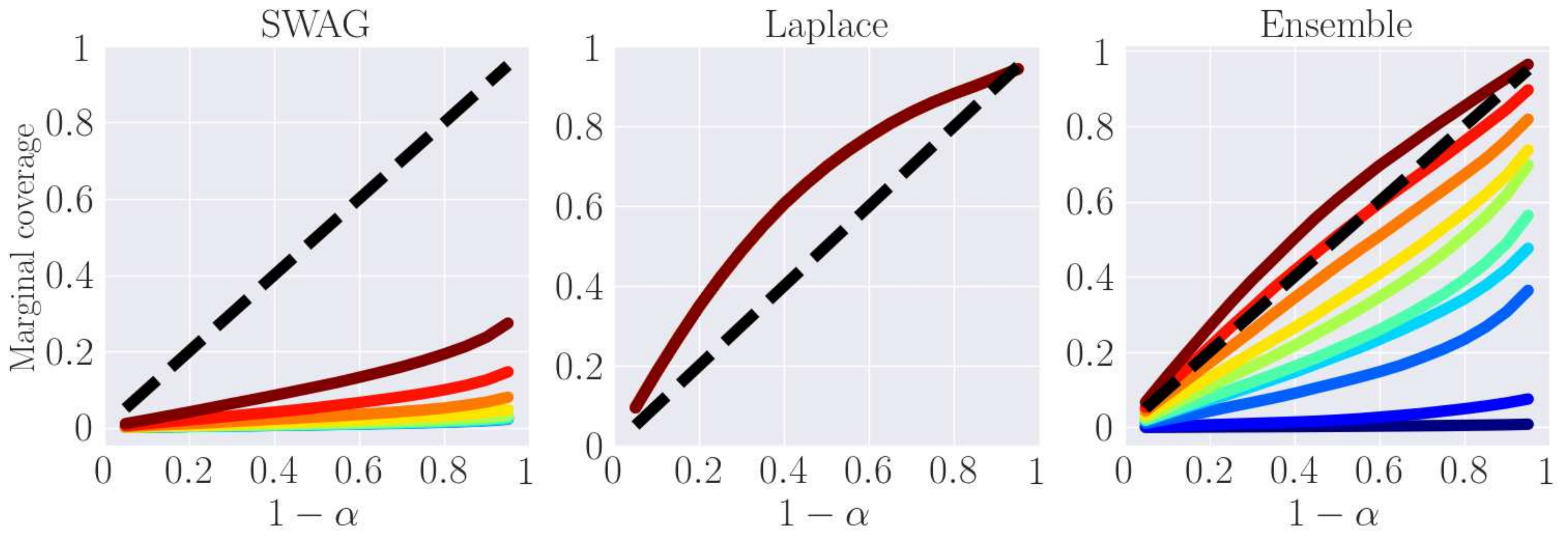}
\includegraphics[width=0.21\textwidth]{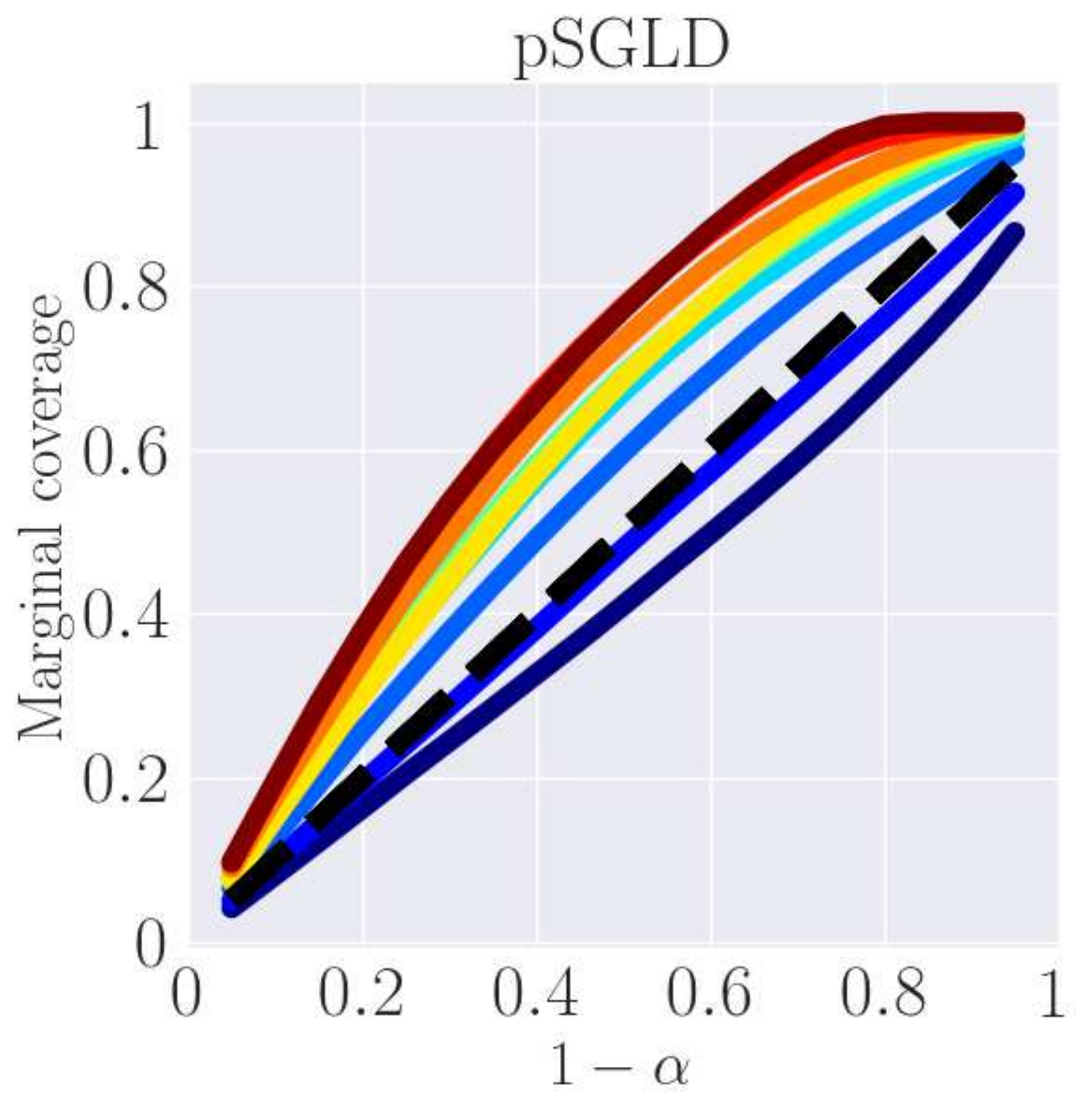}
\caption{Problem AF\#2. Graphs of the marginal coverage probability with respect to the target level for \textbf{MC-Dropout}, \textbf{SWAG}, \textbf{LA-KFAC}, \textbf{deep ensembles}, and \textbf{pSGLD}.}
\label{fig:mcp_graphs_mcdropout_af2}
\end{figure}

\subsection{Additional results for regression problems AF\#3 and AF\#4}
\begin{figure}[h]
\centering
\includegraphics[width=0.42\textwidth]{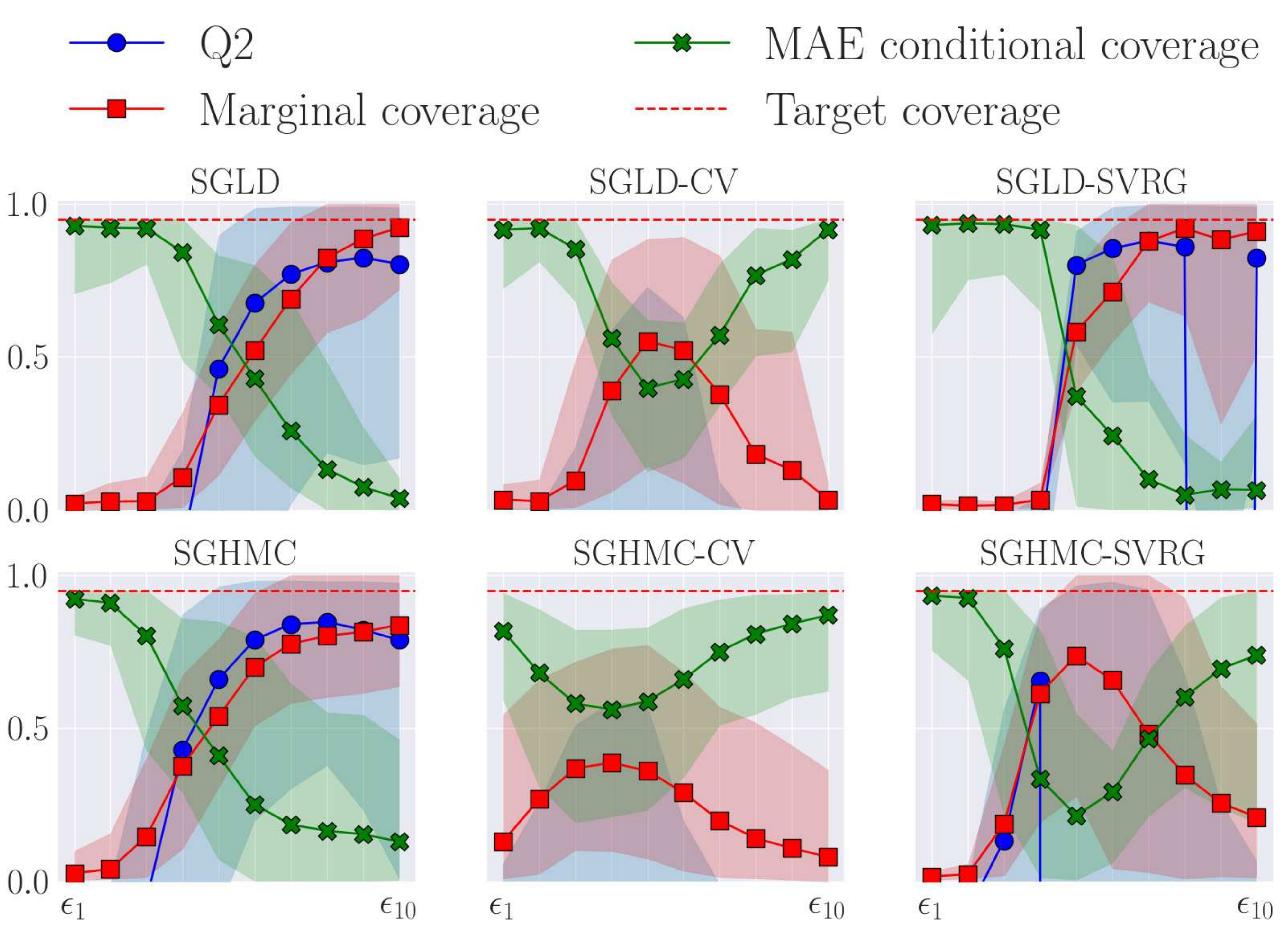}
\includegraphics[width=0.42\textwidth]{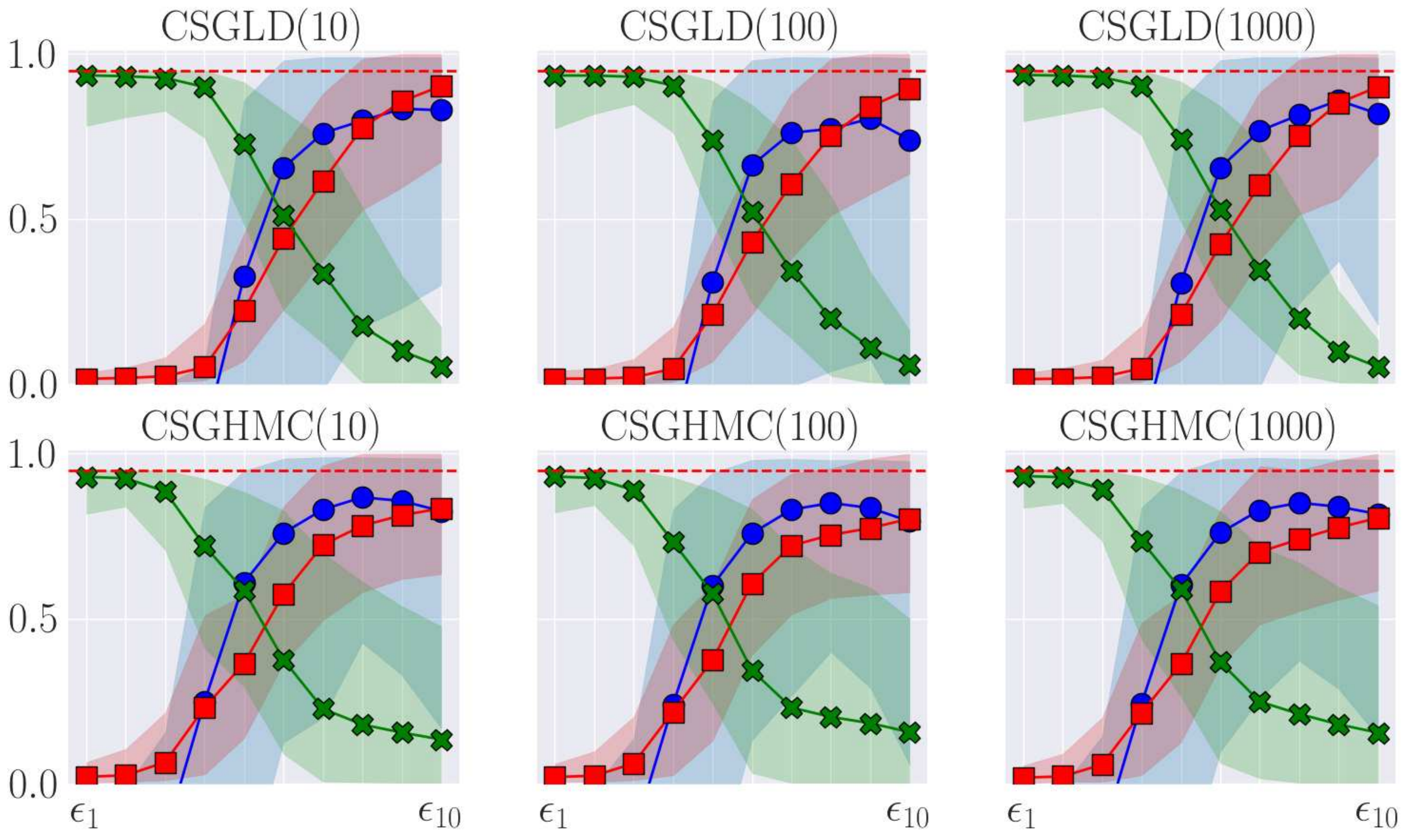}
\includegraphics[width=0.42\textwidth]{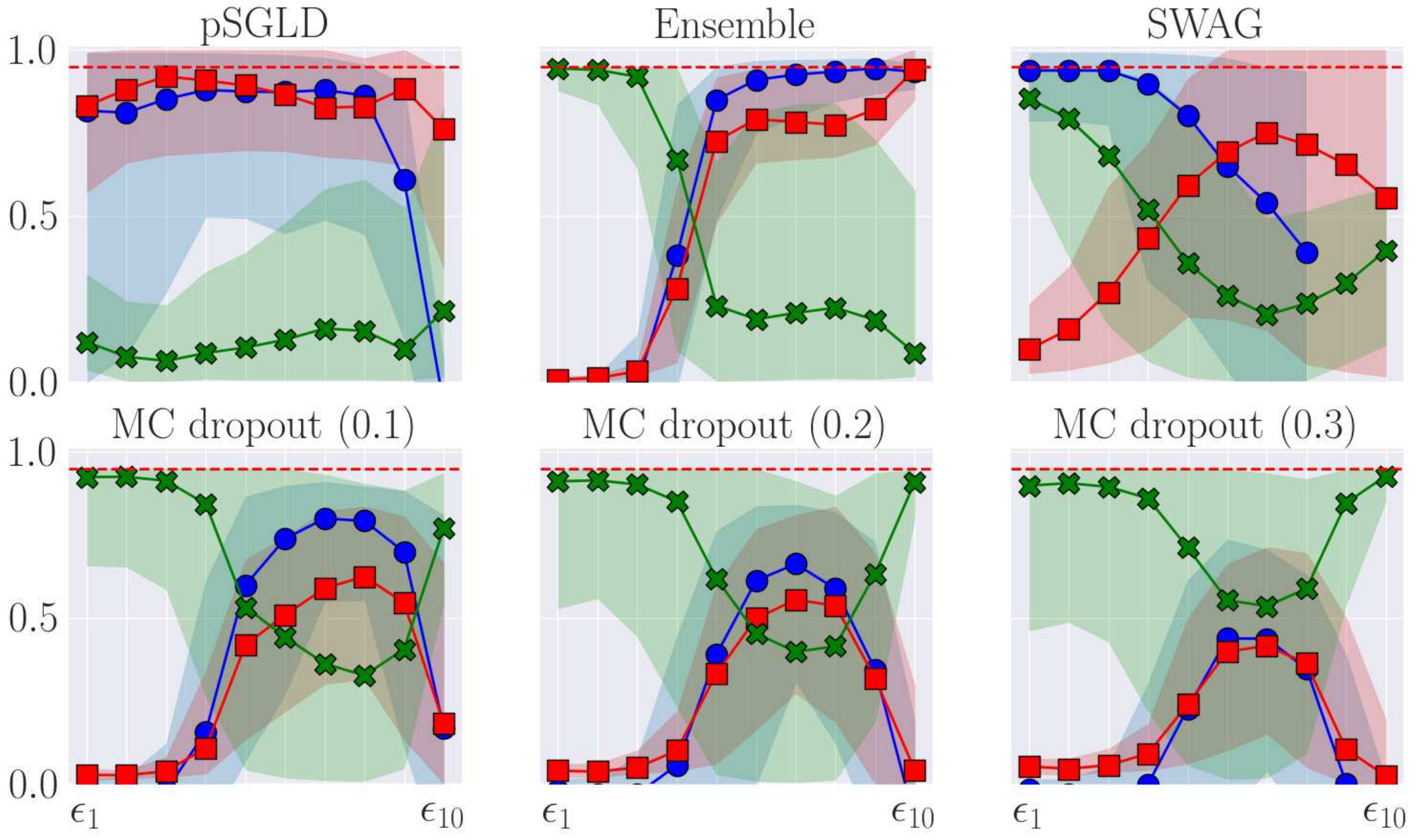}
\caption{Problem AF\#3. Coverage metrics and $Q^2$ coefficient with respect to the step size $\epsilon$. The target coverage is set to $0.95$.}
\label{fig:q2_coverage_af3_sgmcmc.pdf}
\end{figure}

\begin{figure}[h]
\centering
\includegraphics[width=0.38\textwidth]{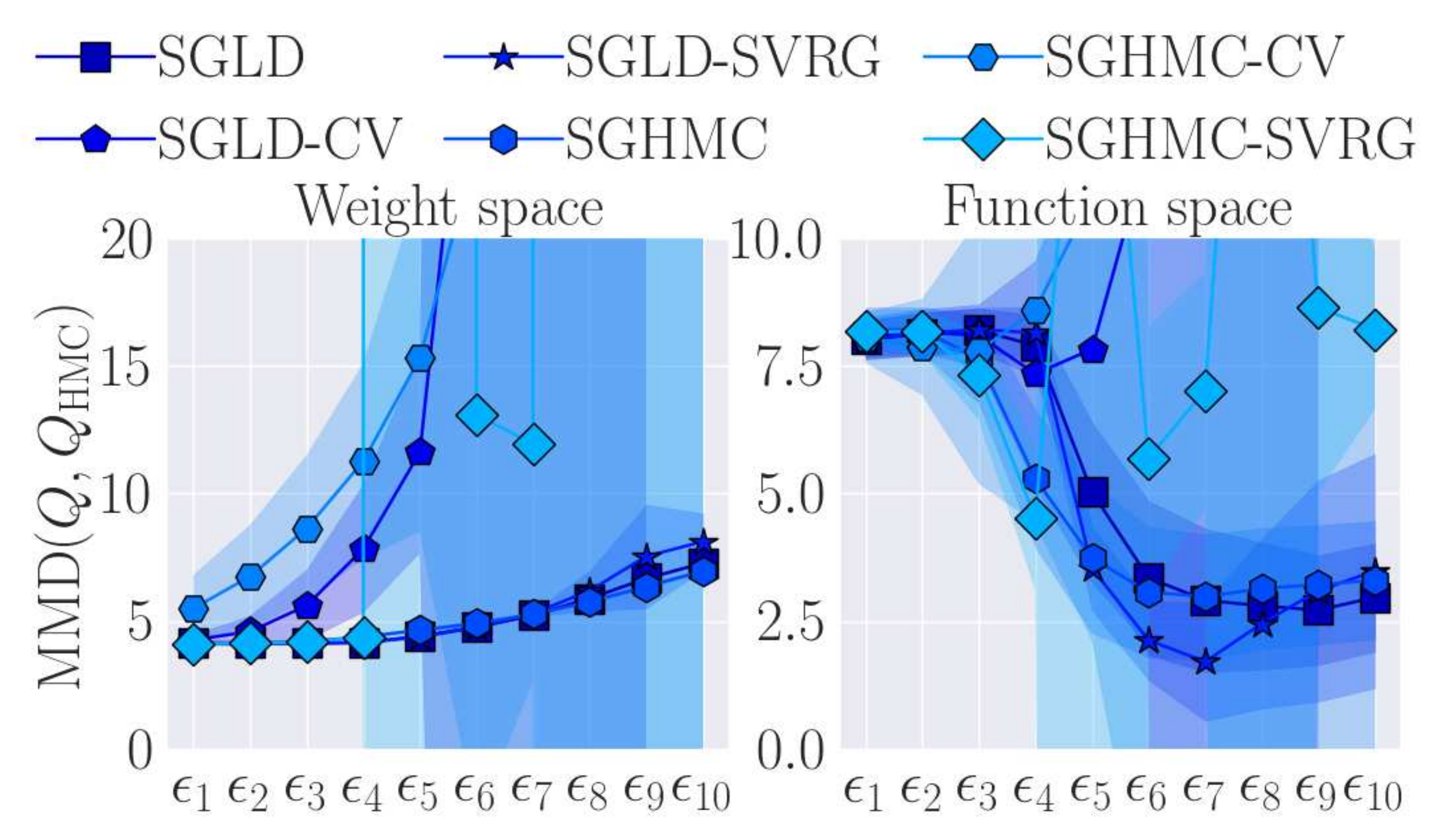}
\includegraphics[width=0.38\textwidth]{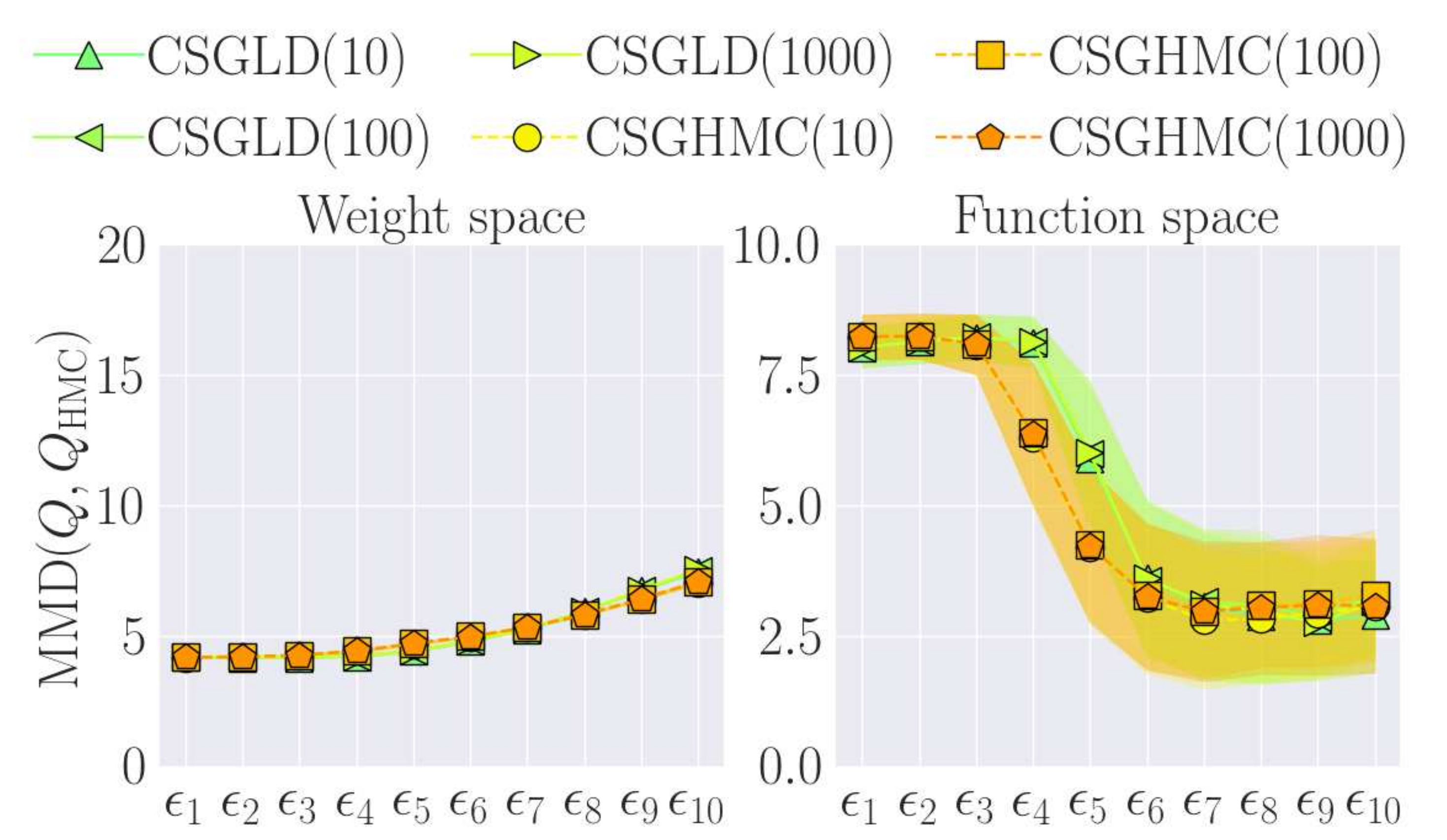}
\includegraphics[width=0.38\textwidth]{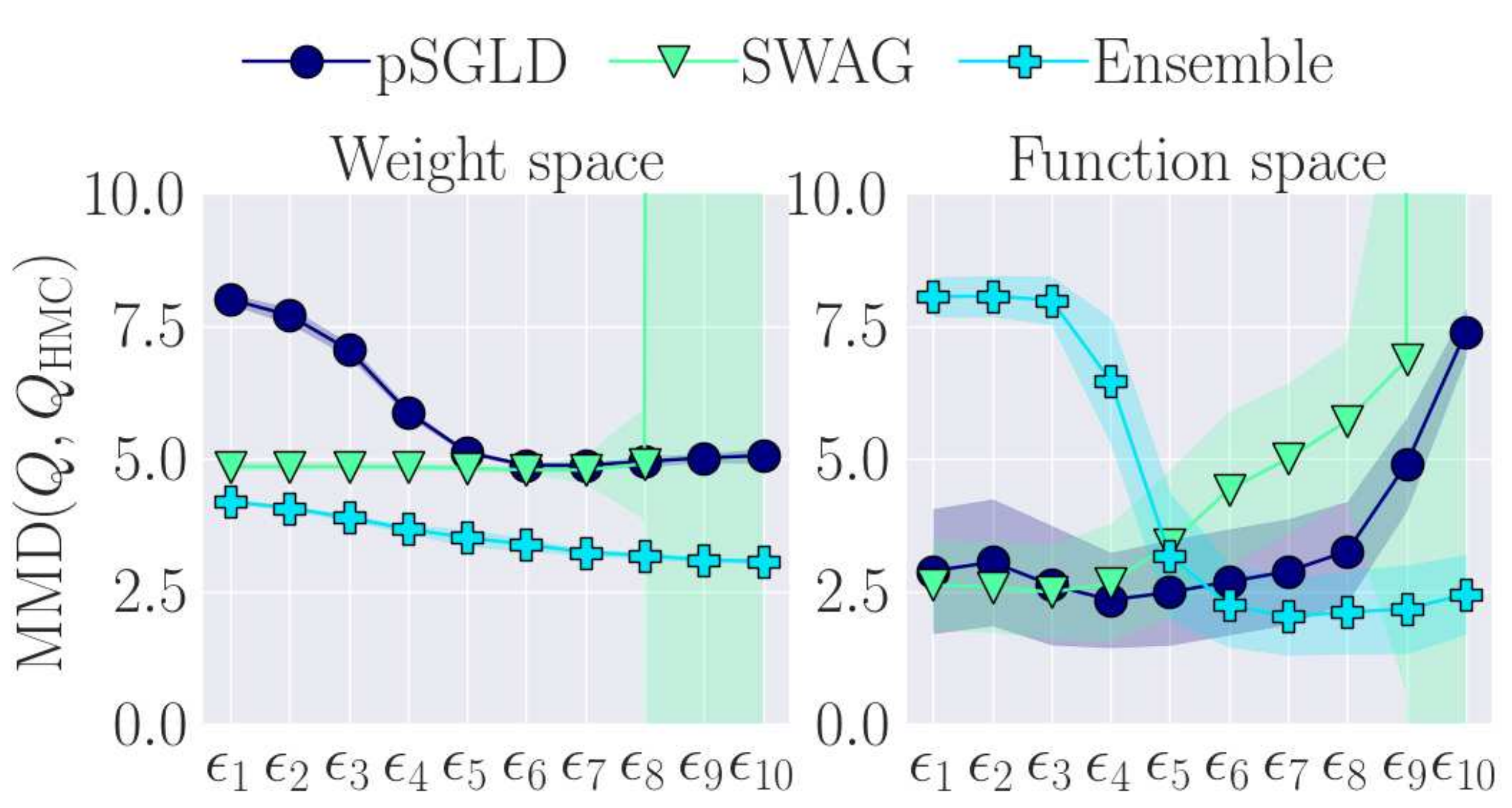}
\caption{Problem AF\#3. Maximum mean discrepancies $\mathrm{MMD}(\bQ,\bQ_{\mathrm{HMC}})$ between the approximated posterior distributions $\bQ$ and the reference HMC sample, in weight and function spaces.}
\label{fig:mmd_weight_function_spaces_af3}
\end{figure}

\begin{figure}[h]
\centering
\includegraphics[width=0.42\textwidth]{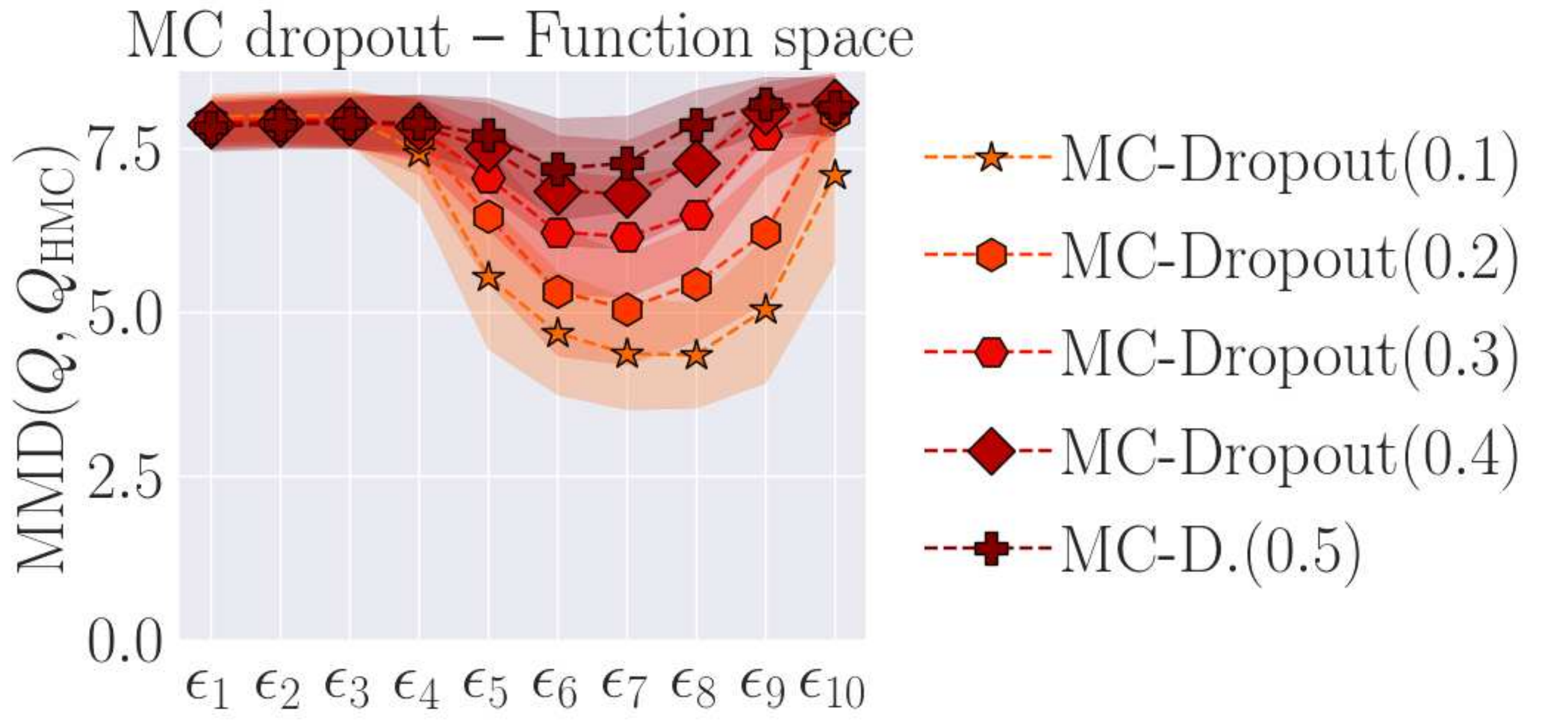}
\caption{Problem AF\#3. Maximum mean discrepancies between the approximated posterior distributions and the reference HMC sample, in weight and function spaces.}
\label{fig:mmd_weight_function_spaces_mcdropout_af3}
\end{figure}

\begin{figure}
\centering
\includegraphics[width=0.44\textwidth]{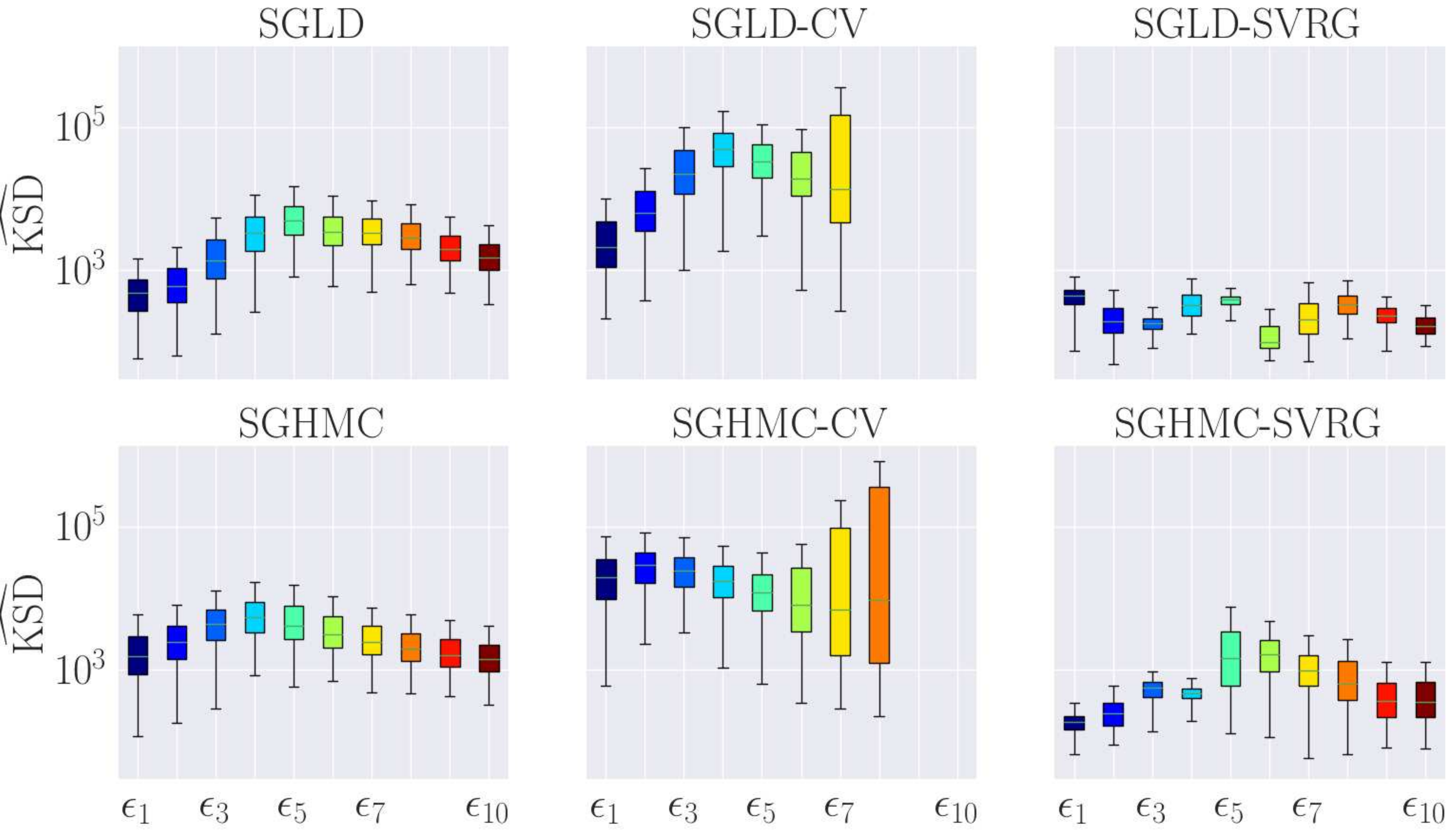}
\includegraphics[width=0.44\textwidth]{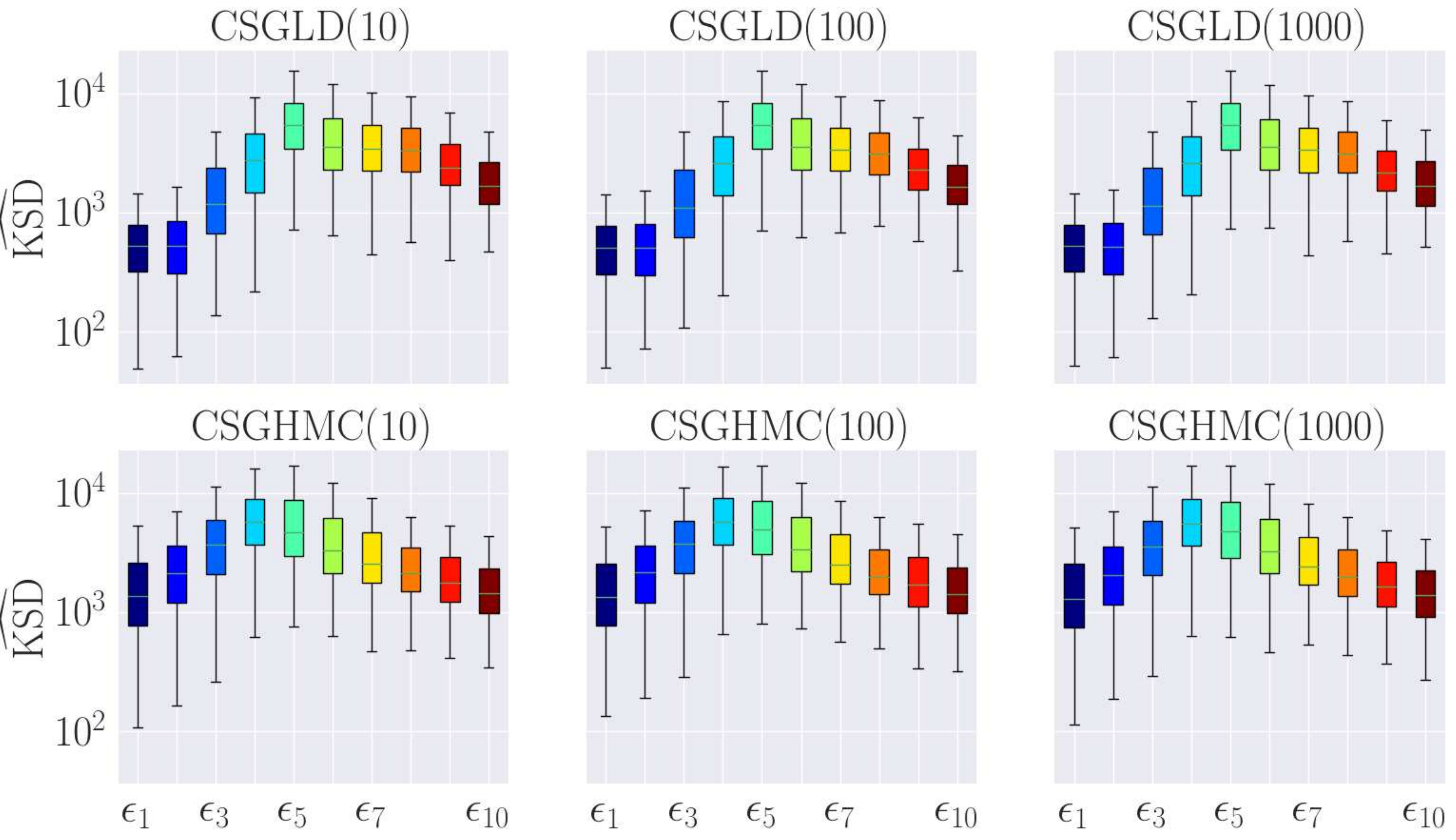}
\includegraphics[width=0.44\textwidth]{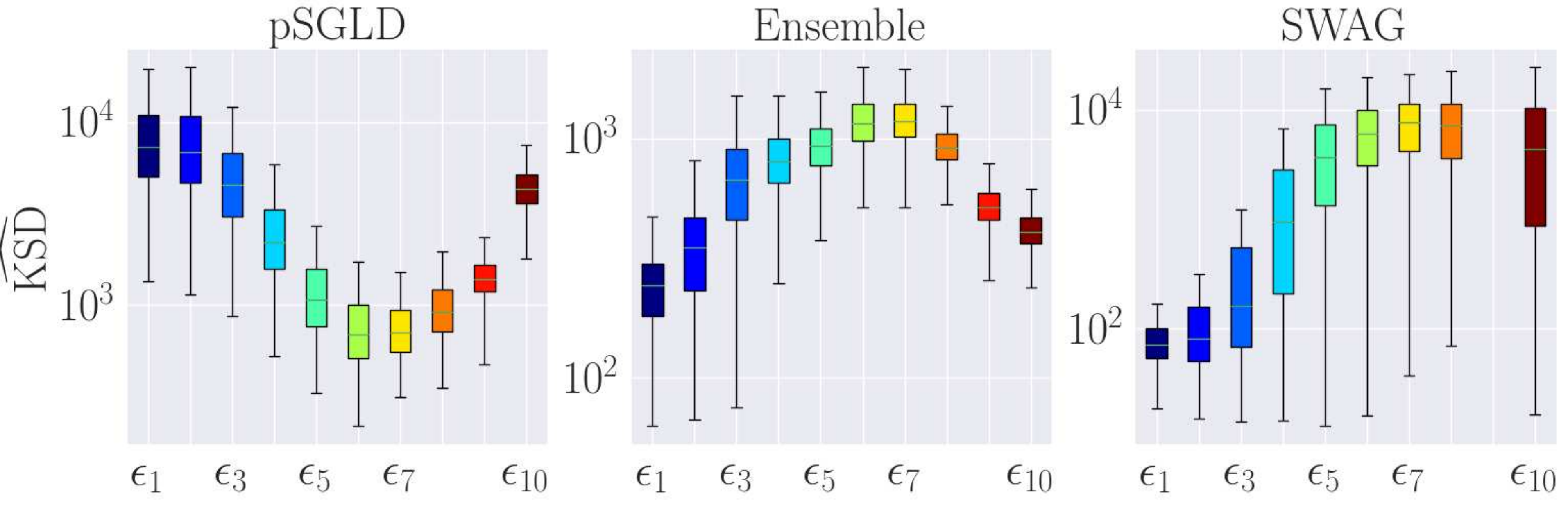}
\caption{Problem AF\#3. Kernelized Stein discrepancies $\mathrm{KSD}(\bP,\bQ)$ between the target posterior measure $\bP$ and the approximated posteriors $\bQ$.}
\label{fig:ksd_af3}
\end{figure}

\begin{figure}[h]
\centering
\includegraphics[width=0.49\textwidth]{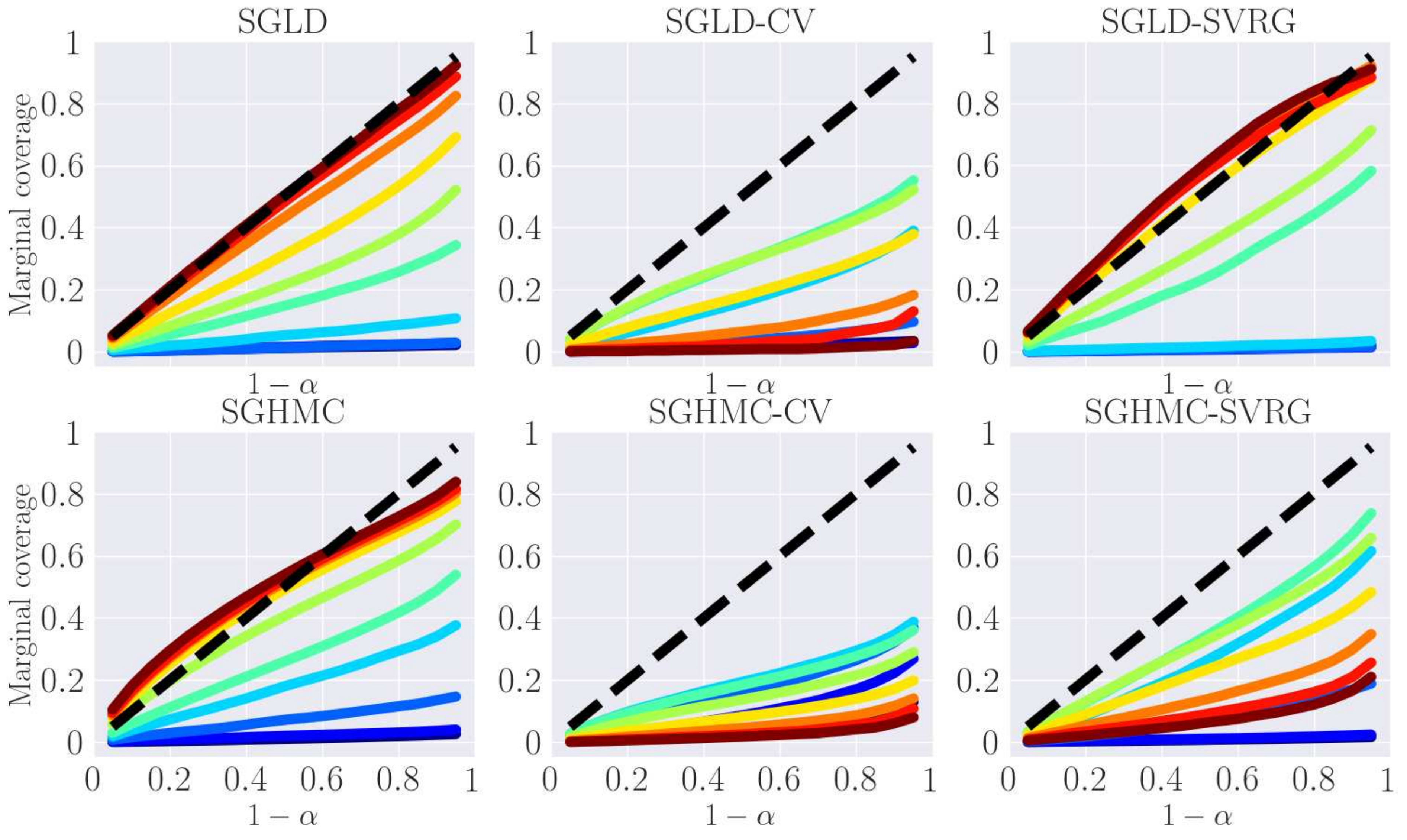}
\includegraphics[width=0.49\textwidth]{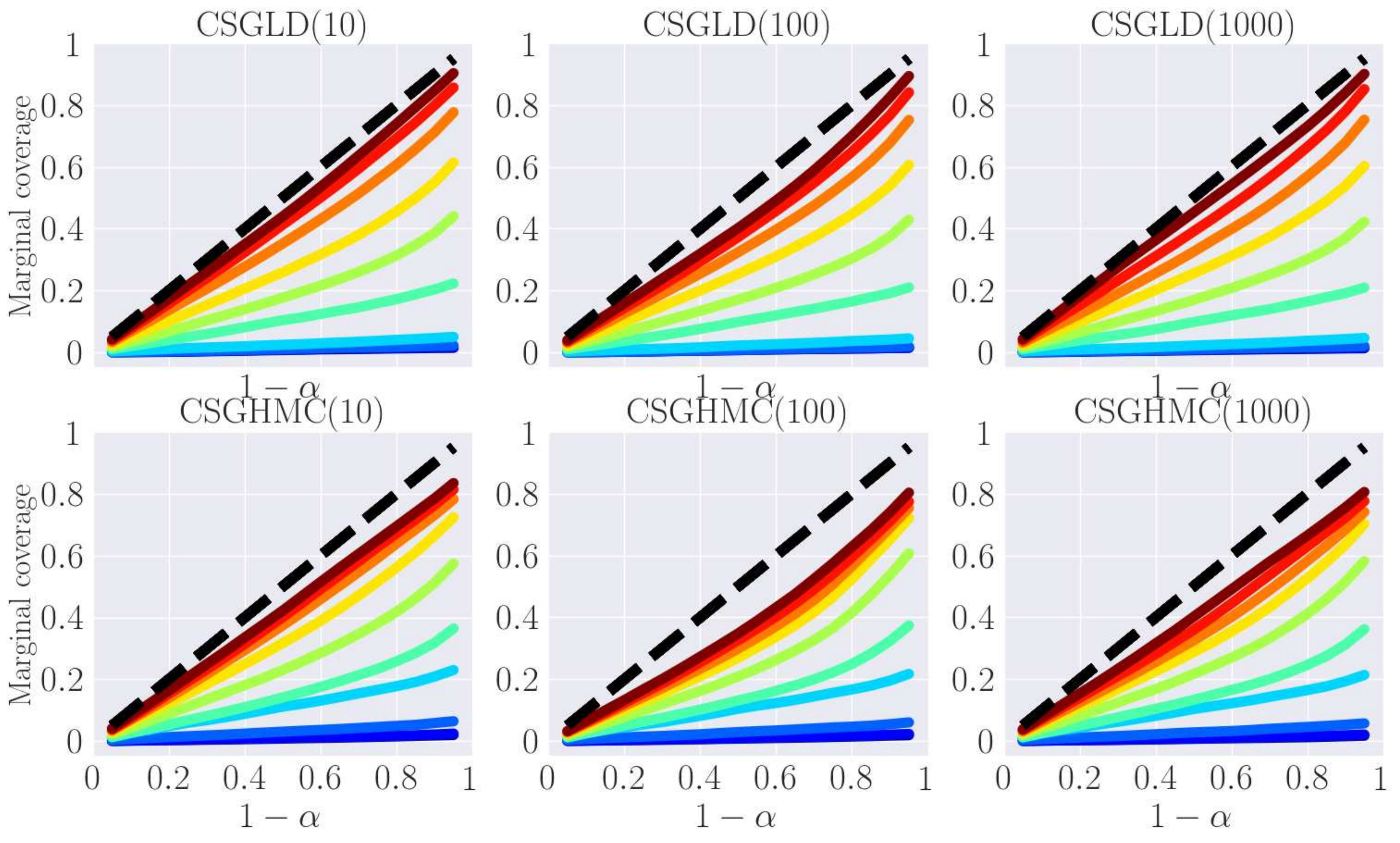}
\caption{Problem AF\#3. Graphs of the marginal coverage probability with respect to the target level for \textbf{SGMCMC} methods. The curves are colored by the value of the underlying step size: dark blue corresponds to the lowest, while dark red corresponds to the highest step size.}
\label{fig:mcp_graphs_sgmcmc_af3}
\end{figure}

\begin{figure}[h]
\centering
\includegraphics[width=\textwidth]{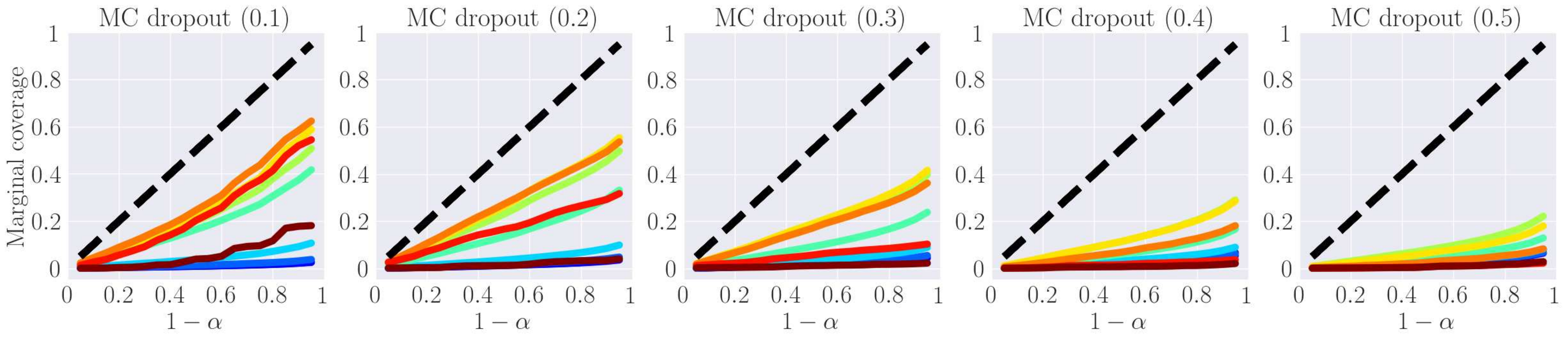}

\includegraphics[width=0.62\textwidth]{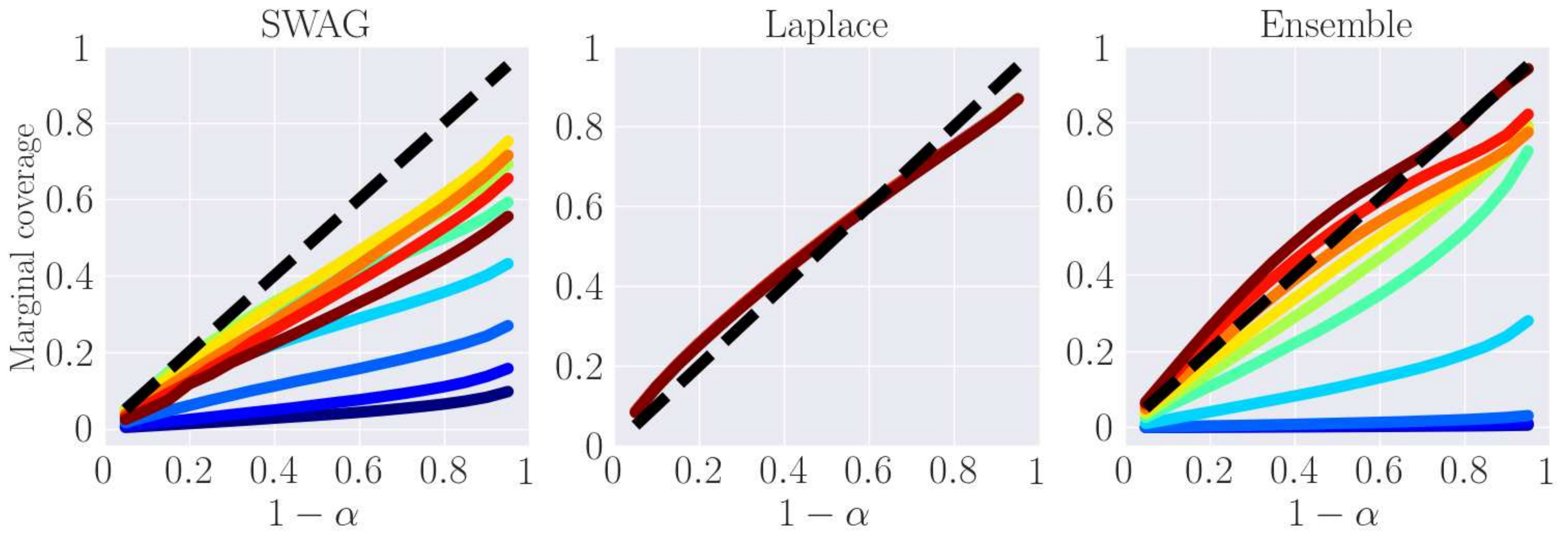}
\includegraphics[width=0.21\textwidth]{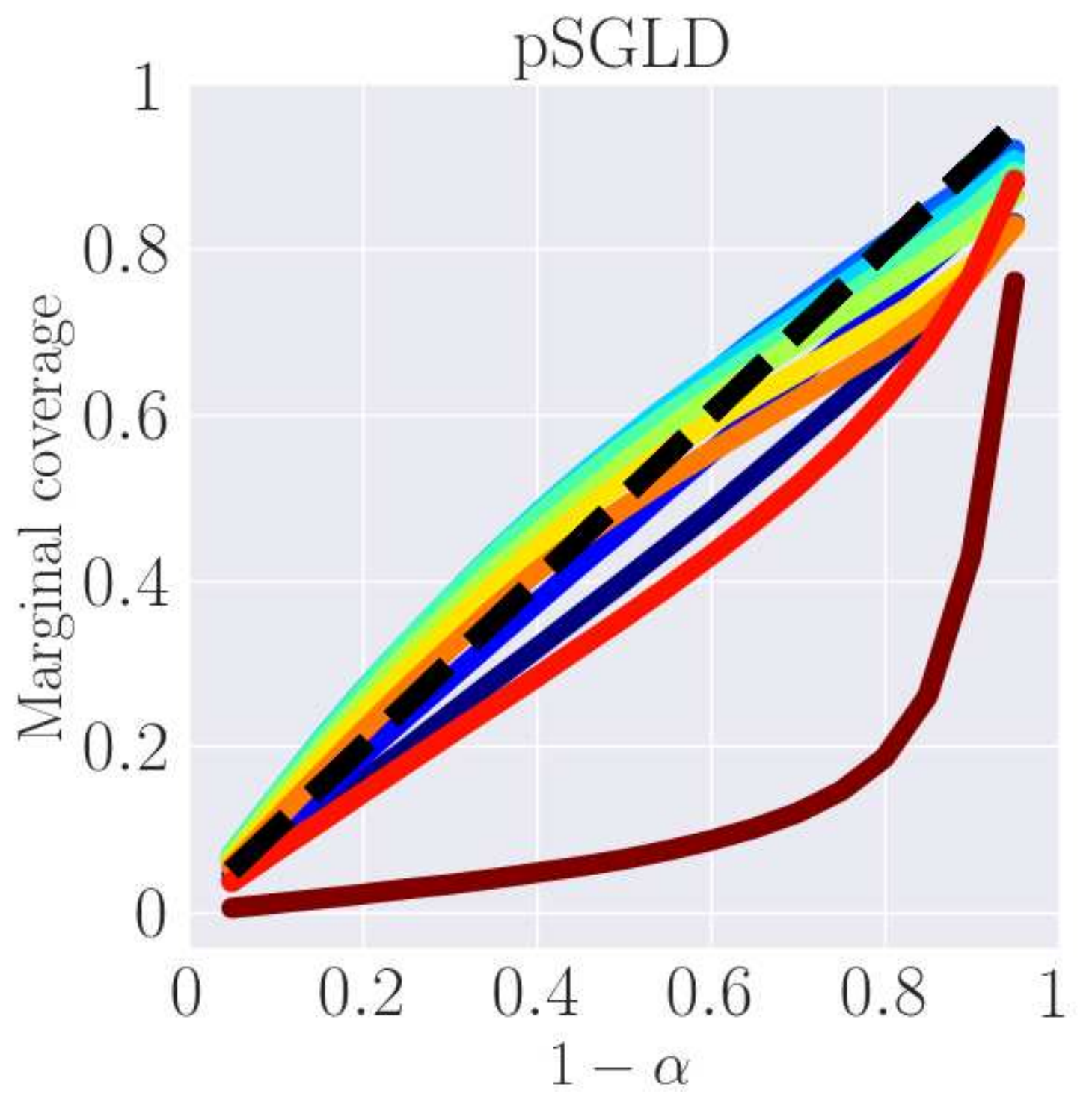}
\caption{Problem AF\#3. Graphs of the marginal coverage probability with respect to the target level for \textbf{MC-Dropout}, \textbf{SWAG}, \textbf{LA-KFAC}, \textbf{deep ensembles}, and \textbf{pSGLD}.}
\label{fig:mcp_graphs_mcdropout_af3}
\end{figure}

\begin{figure}[h]
\centering
\includegraphics[width=0.42\textwidth]{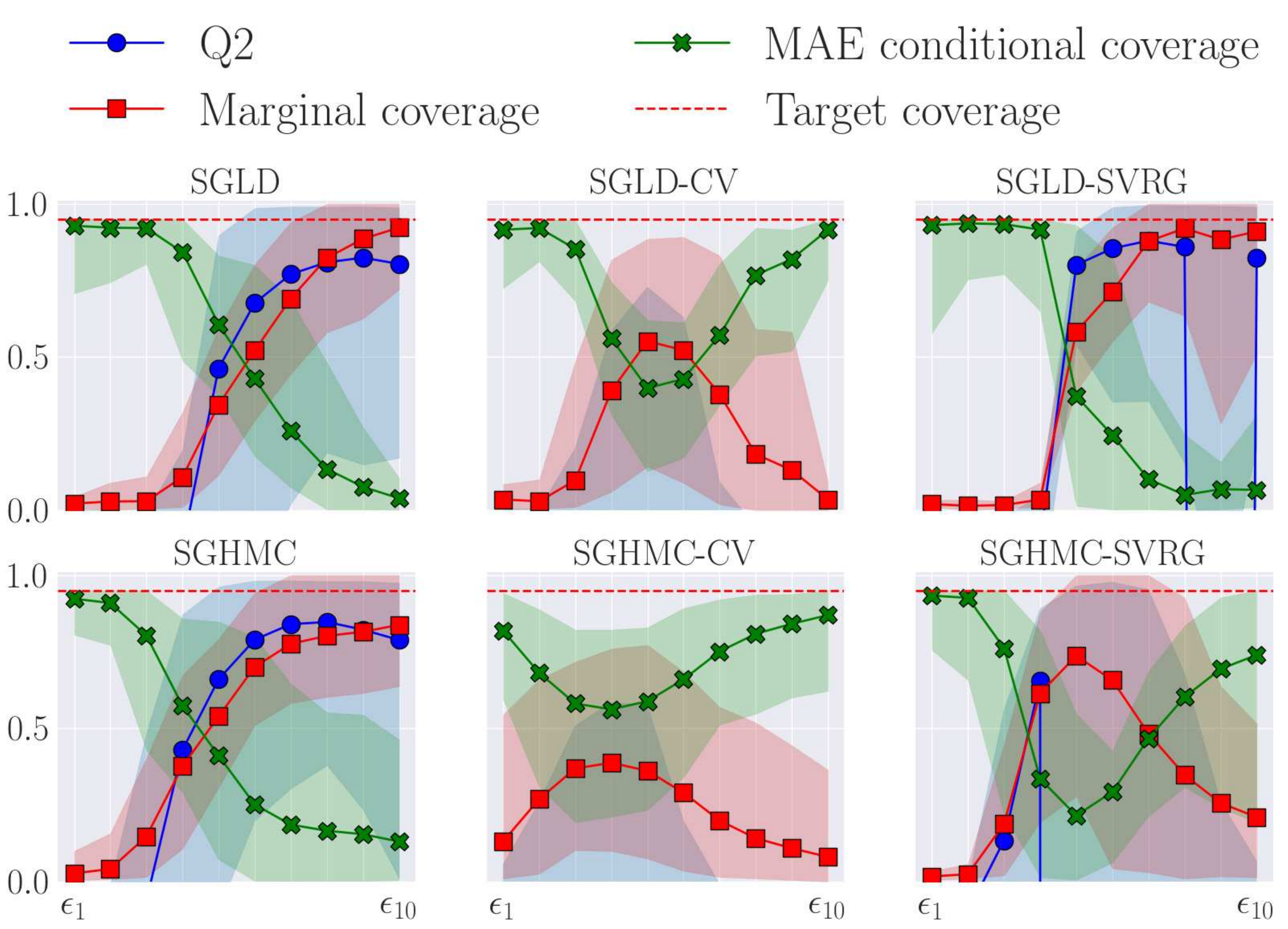}
\includegraphics[width=0.42\textwidth]{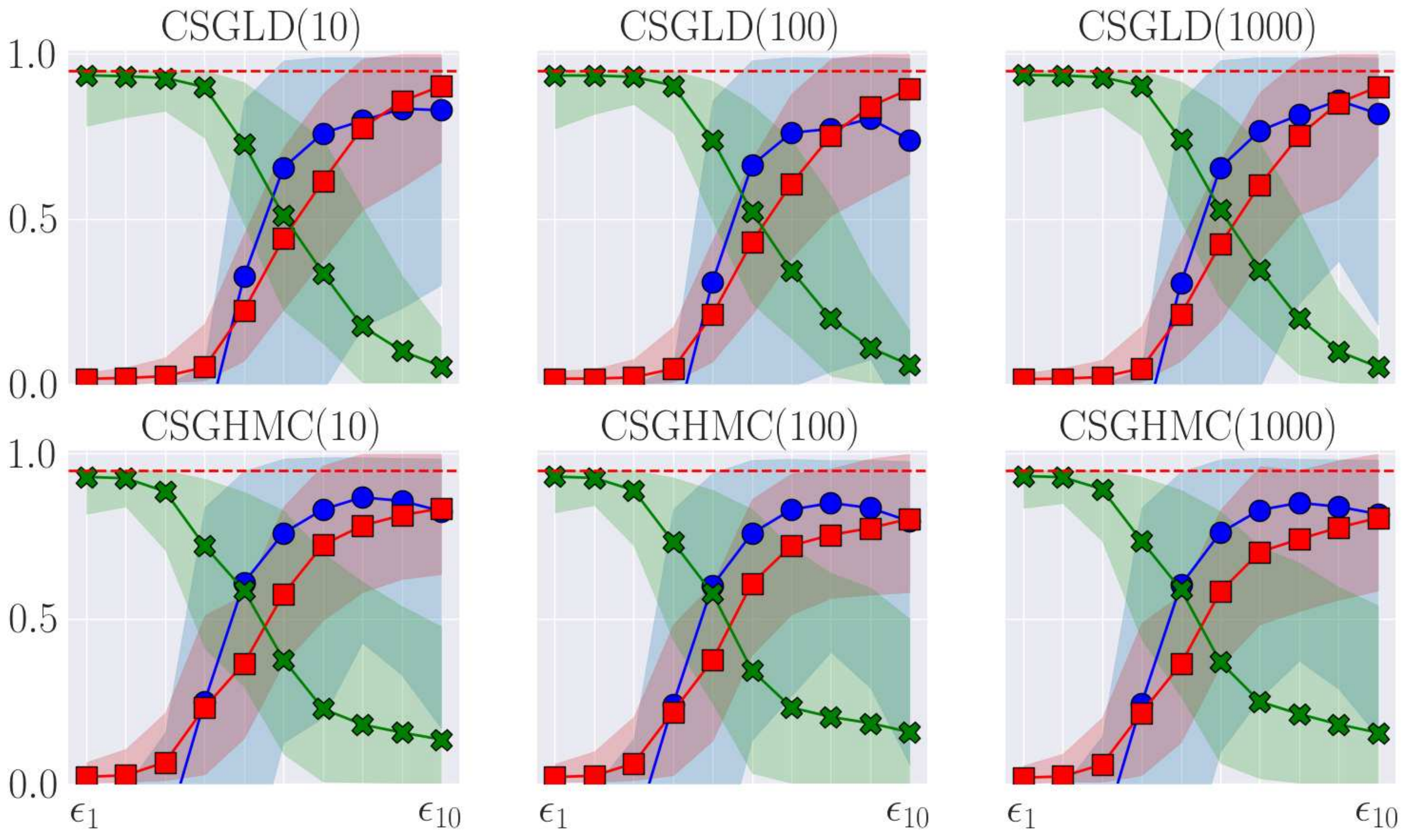}
\includegraphics[width=0.42\textwidth]{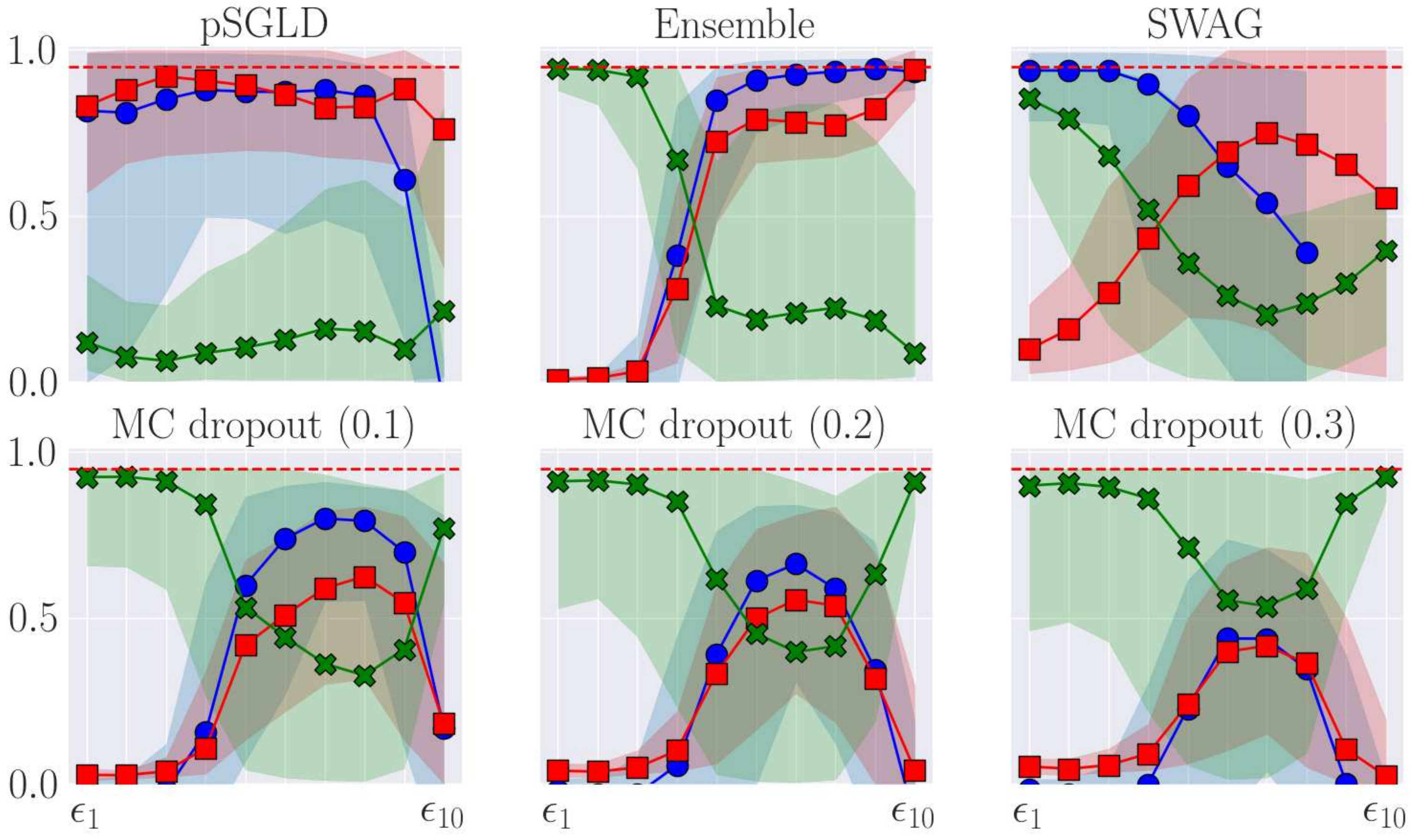}
\caption{Problem AF\#4. Coverage metrics and $Q^2$ coefficient with respect to the step size $\epsilon$. The target coverage is set to $0.95$.}
\label{fig:q2_coverage_af4_sgmcmc.pdf}
\end{figure}

\begin{figure}[h]
\centering
\includegraphics[width=0.38\textwidth]{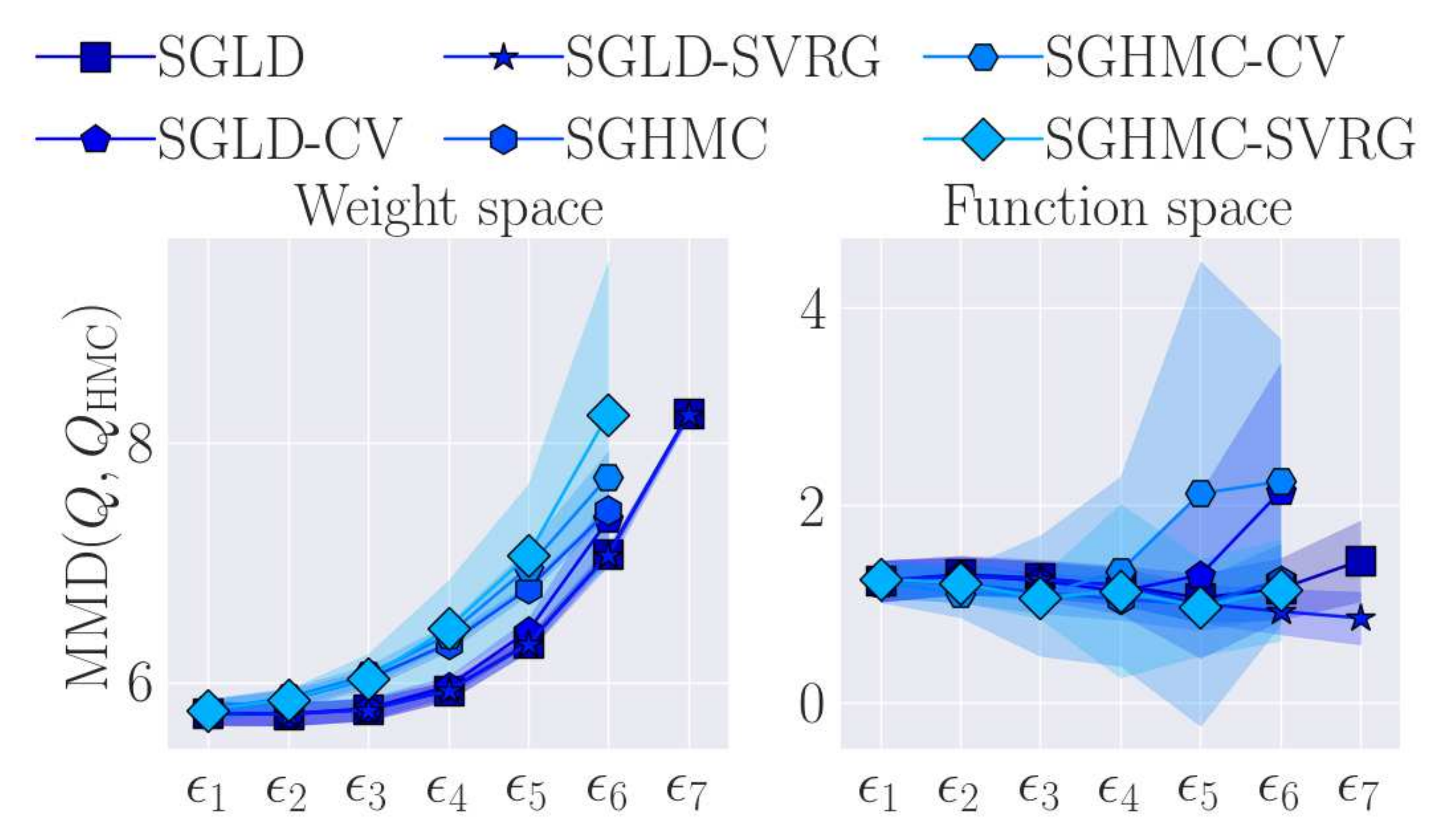}
\includegraphics[width=0.38\textwidth]{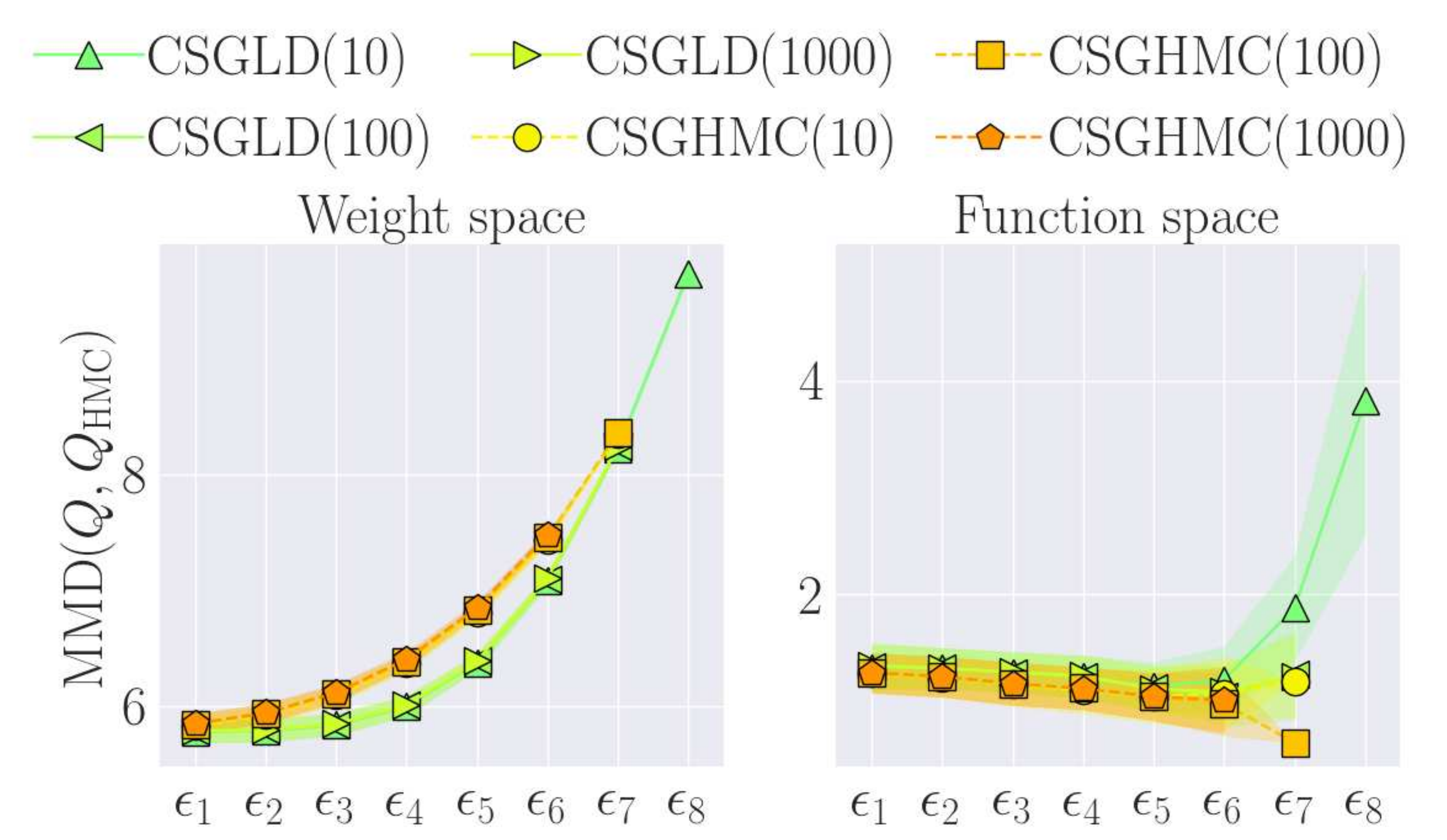}
\includegraphics[width=0.38\textwidth]{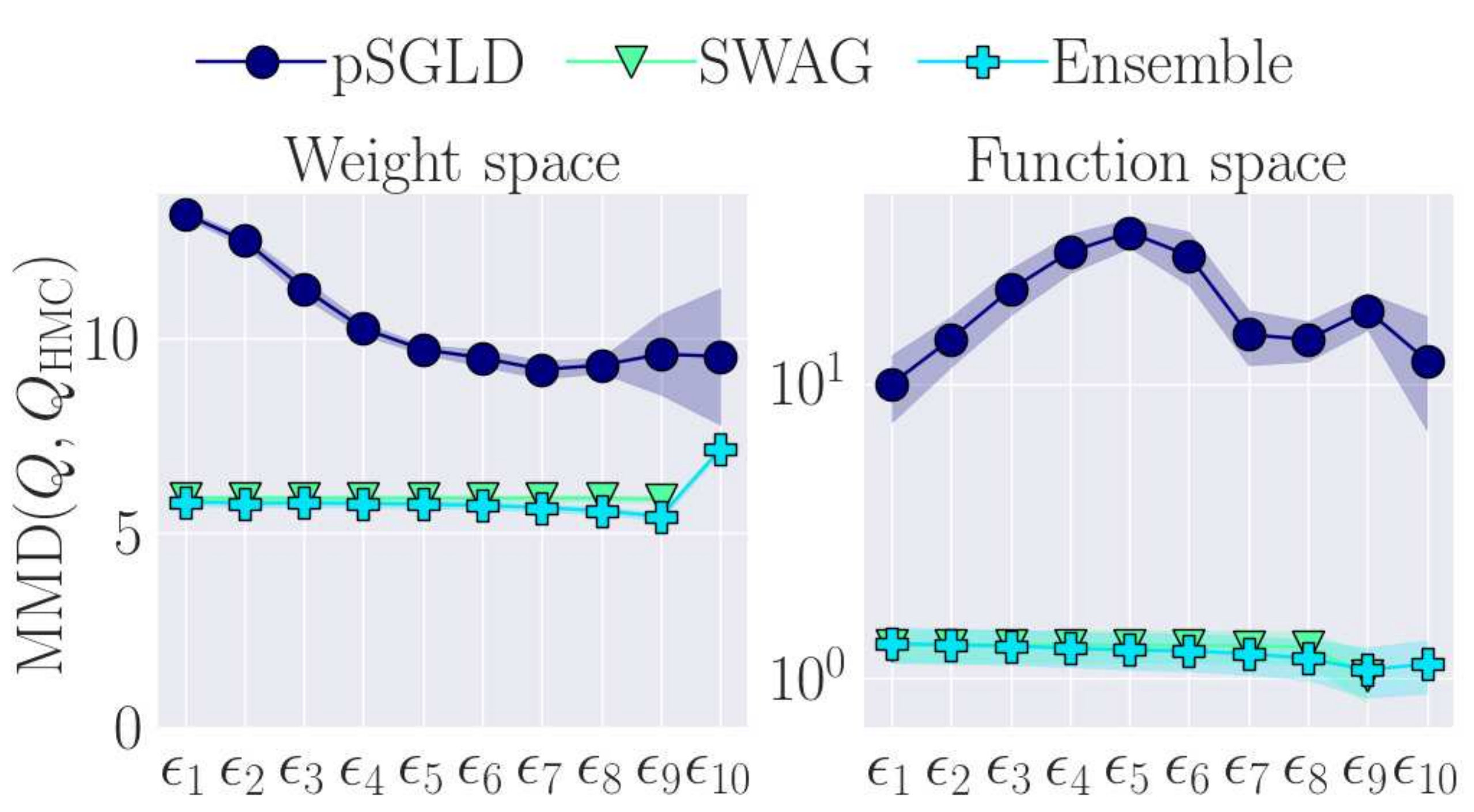}
\caption{Problem AF\#4. Maximum mean discrepancies $\mathrm{MMD}(\bQ,\bQ_{\mathrm{HMC}})$ between the approximated posterior distributions $\bQ$ and the reference HMC sample, in weight and function spaces.}
\label{fig:mmd_weight_function_spaces_af4}
\end{figure}

\begin{figure}[h]
\centering
\includegraphics[width=0.42\textwidth]{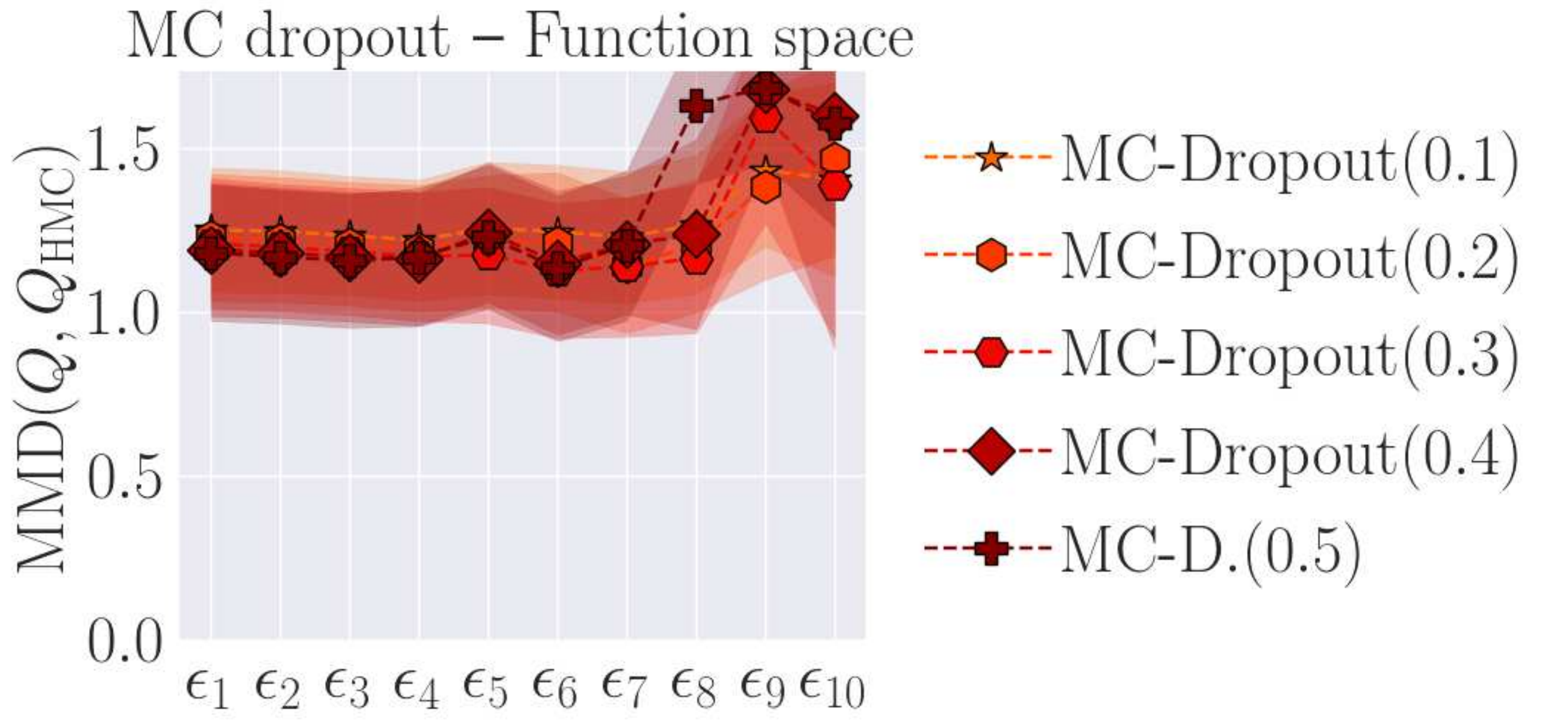}
\caption{Problem AF\#4. Maximum mean discrepancies between the approximated posterior distributions and the reference HMC sample, in weight and function spaces.}
\label{fig:mmd_weight_function_spaces_mcdropout_af4}
\end{figure}

\begin{figure}
\centering
\includegraphics[width=0.44\textwidth]{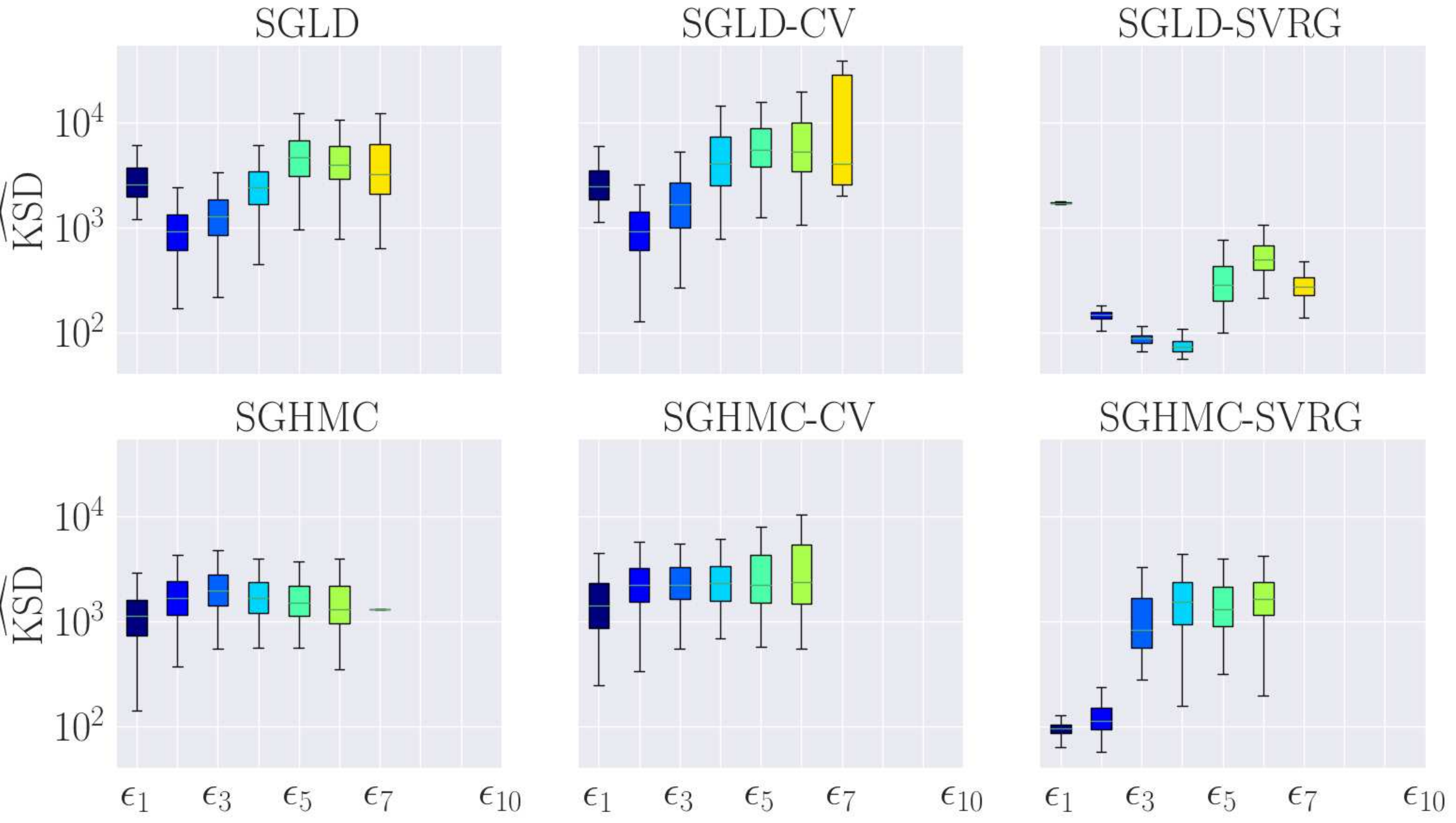}
\includegraphics[width=0.44\textwidth]{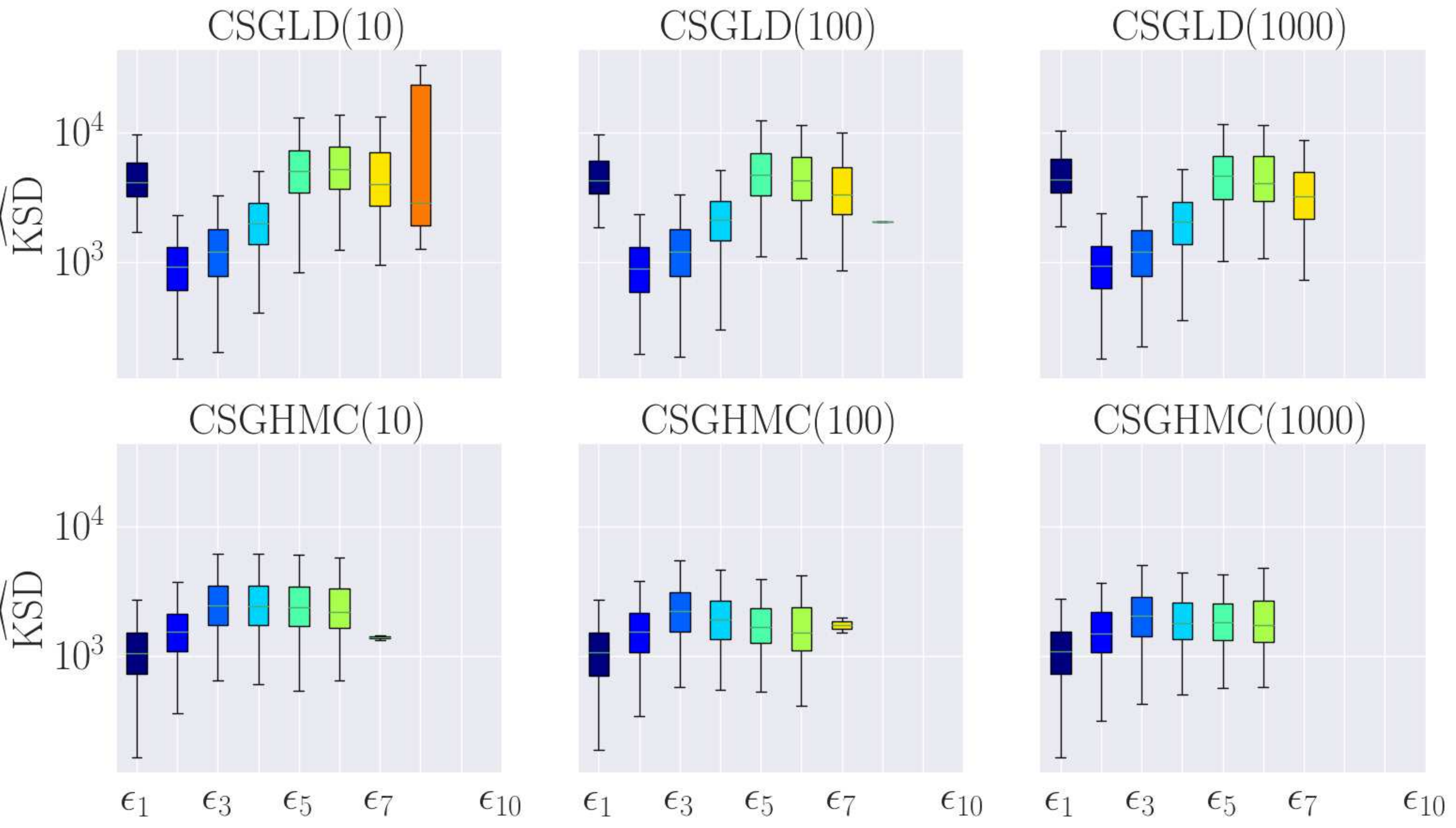}
\includegraphics[width=0.44\textwidth]{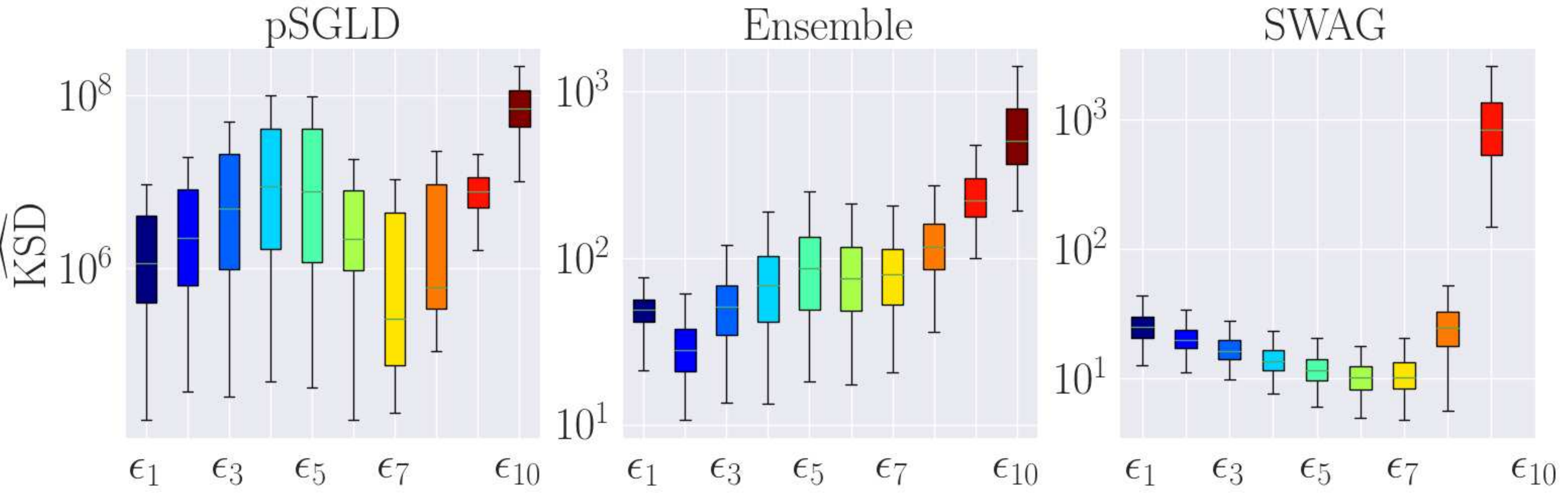}
\caption{Problem AF\#4. Kernelized Stein discrepancies $\mathrm{KSD}(\bP,\bQ)$ between the target posterior measure $\bP$ and the approximated posteriors $\bQ$.}
\label{fig:ksd_af4}
\end{figure}

\begin{figure}[h]
\centering
\includegraphics[width=0.49\textwidth]{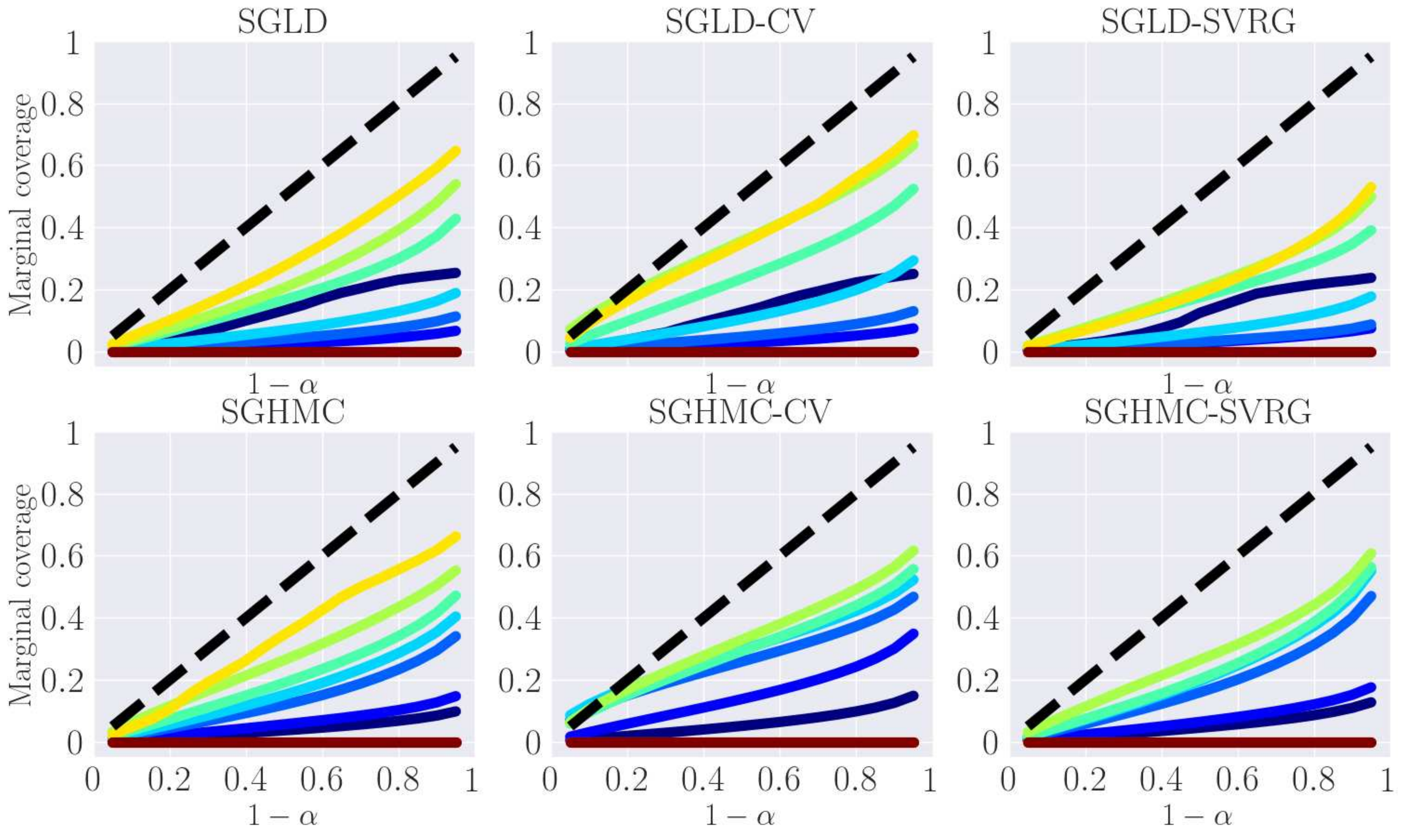}
\includegraphics[width=0.49\textwidth]{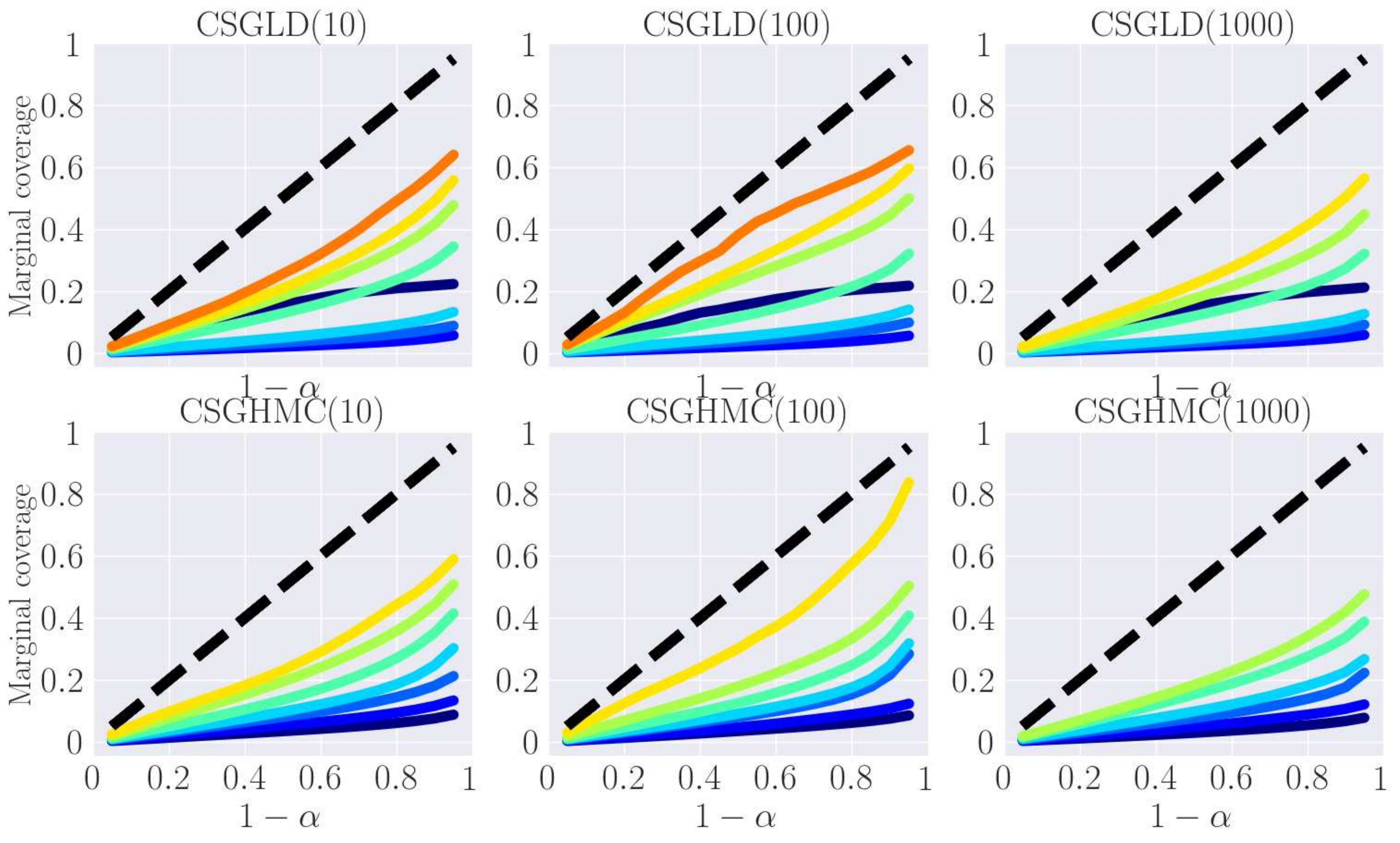}
\caption{Problem AF\#4. Graphs of the marginal coverage probability with respect to the target level for \textbf{SGMCMC} methods. The curves are colored by the value of the underlying step size: dark blue corresponds to the lowest, while dark red corresponds to the highest step size.}
\label{fig:mcp_graphs_sgmcmc_af4}
\end{figure}

\begin{figure}[h]
\centering
\includegraphics[width=\textwidth]{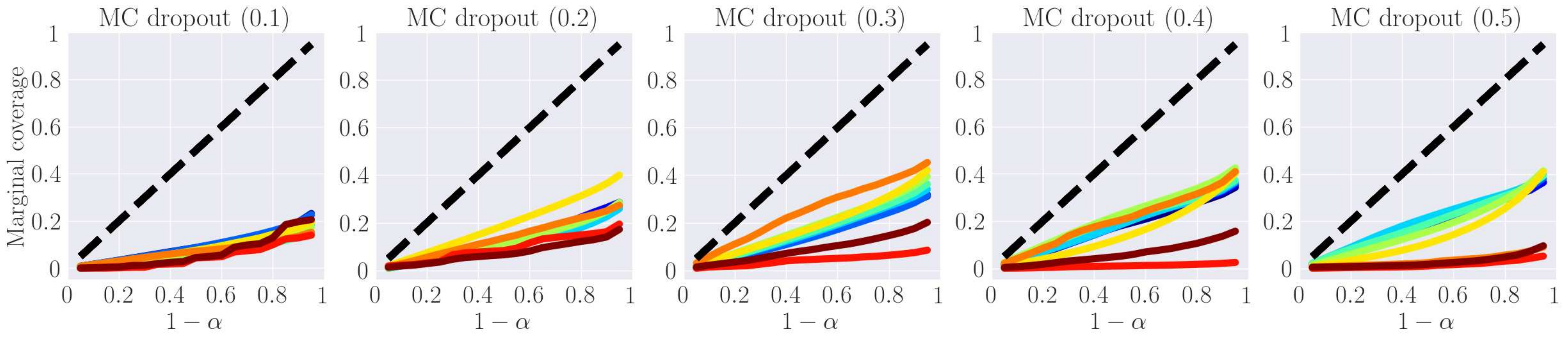}

\includegraphics[width=0.62\textwidth]{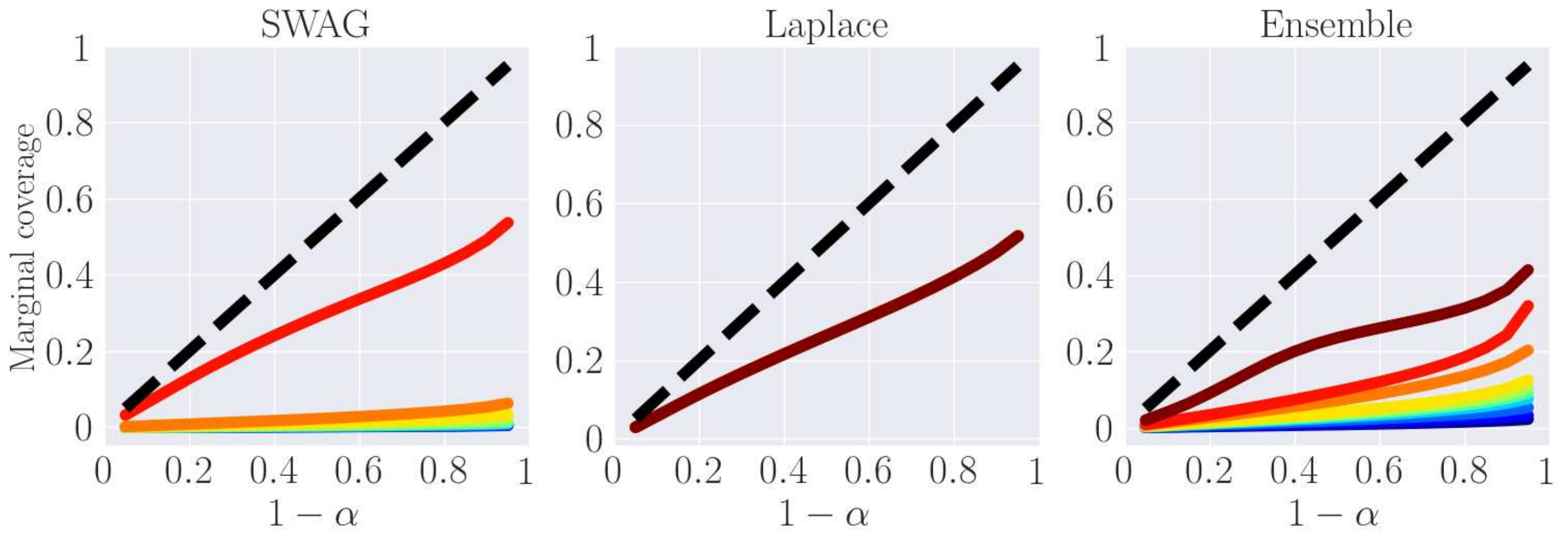}
\includegraphics[width=0.21\textwidth]{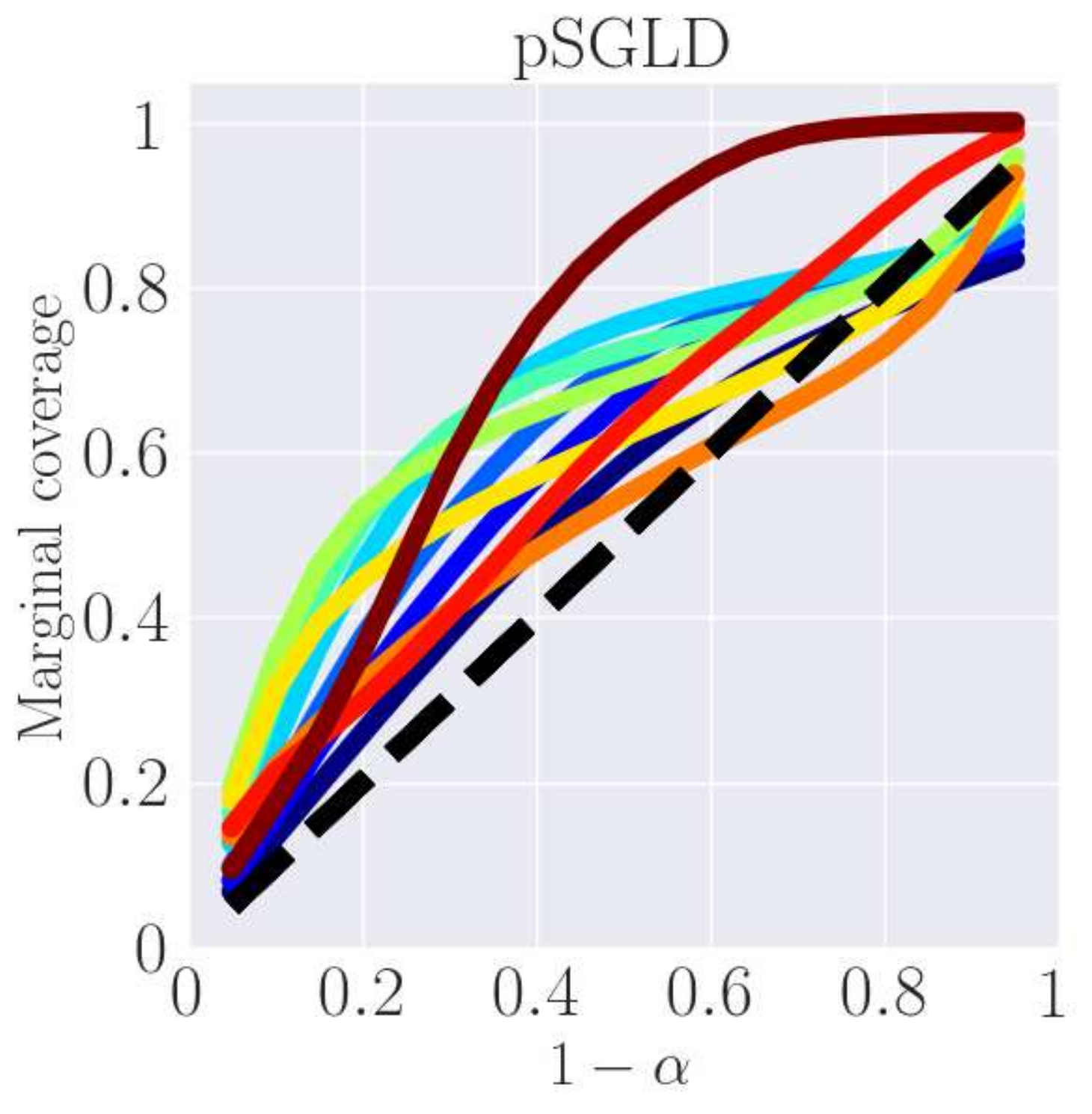}
\caption{Problem AF\#4. Graphs of the marginal coverage probability with respect to the target level for \textbf{MC-Dropout}, \textbf{SWAG}, \textbf{LA-KFAC}, \textbf{deep ensembles}, and \textbf{pSGLD}.}
\label{fig:mcp_graphs_mcdropout_af4}
\end{figure}

All the results obtained for the third and fourth regression problems are shown in Figures \ref{fig:q2_coverage_af3_sgmcmc.pdf}-\ref{fig:mcp_graphs_mcdropout_af3}, and Figures \ref{fig:q2_coverage_af4_sgmcmc.pdf}-\ref{fig:mcp_graphs_mcdropout_af4}. The distributions of the test data differ significantly from the training data in problem AF\#3, and even more in problem AF\#4. It can be seen in the associated results that these problems are challenging for all the methods, which are not able to reach the target coverage probability, nor obtain good regression coeffcients.

\clearpage

\end{document}